\documentclass[final,3p,times,twocolumn]{elsarticle}

\usepackage[english]{babel}
\usepackage[T1]{fontenc}
\usepackage[font=footnotesize]{subcaption}
\usepackage{float}
\usepackage{color}
\usepackage{graphics}
\usepackage{graphicx}
\usepackage{amsmath}
\usepackage{amssymb}
\usepackage{amsfonts}
\usepackage{microtype}
\usepackage{comment}
\usepackage{multirow}
\usepackage{booktabs}

\usepackage{tikz}
\usepackage{url}
\usepackage[noend]{algpseudocode}
\usepackage{algorithm}
\usepackage{algorithmicx}
\usepackage[multiple]{footmisc}
\usepackage[autostyle]{csquotes}
\usepackage{textcomp}
\usepackage{array}

\newcommand{\mainobj}{\textbf{main object}}
\newcommand{\refobj}{\textbf{reference object}}
\newcommand{\scrapbook}{\textit{Scrapbook}}
\newcommand{\cocoscrapbook}{\textit{COCO} \scrapbook}
\newcommand{\cocoplainbgscrapbook}{\textit{COCO Plain BG} \scrapbook}
\newcommand{\uniqueallscrapbook}{\textit{Unique All}\ \scrapbook}

\newenvironment{myenumerate}{
\begin{enumerate}
  \setlength{\itemsep}{5pt}
  \setlength{\parskip}{0pt}
  \setlength{\parsep}{5pt}
}{\end{enumerate}}

\newenvironment{myitemize}{
\begin{itemize}
  \setlength{\itemsep}{5pt}
  \setlength{\parskip}{0pt}
  \setlength{\parsep}{5pt}
}{\end{itemize}}

\journal{JOURNAL NAME}

\begin{document}

\sloppy
\tolerance = 999


\begin{frontmatter}

\title{A Framework for Generating Artificial Datasets \\ to Validate Absolute and Relative Position Concepts}

\author{George Corr\^ea de Ara\'ujo}
\ead{george.araujo@ic.unicamp.br}
\author{Helena de Almeida Maia}
\ead{helena.mais@ic.unicamp.br}
\author{Helio Pedrini}
\ead{helio@ic.unicamp.br}
\address{Institute of Computing, University of Campinas, Campinas, SP, 13083-852, Brazil}

\begin{abstract}
In this paper, we present the \scrapbook\ framework, a novel methodology designed to generate extensive datasets for probing the learned concepts of artificial intelligence (AI) models. The framework focuses on fundamental concepts such as object recognition, absolute and relative positions, and attribute identification. By generating datasets with a large number of questions about individual concepts and a wide linguistic variation, the \scrapbook\ framework aims to validate the model's understanding of these basic elements before tackling more complex tasks. Our experimental findings reveal that, while contemporary models demonstrate proficiency in recognizing and enumerating objects, they encounter challenges in comprehending positional information and addressing inquiries with additional constraints. Specifically, the MobileVLM-V2 model showed significant answer disagreements and plausible wrong answers, while other models exhibited a bias toward affirmative answers and struggled with questions involving geometric shapes and positional information, indicating areas for improvement in understanding and consistency. The proposed framework offers a valuable instrument for generating diverse and comprehensive datasets, which can be utilized to systematically assess and enhance the performance of AI models.
\end{abstract}


\end{frontmatter}


\section{Introduction}
\label{sec:introduction}

The use of deep neural networks has enabled unprecedented progress in the fields of computer vision (CV)~\cite{3439723,9716741,3505244,CHAI2021100134,esteva2021deep,9313289,valipoor2022recent,liu2021plant} and natural language processing (NLP)~\cite{548916,neco.1997.9.8.1735,9640649,NIPS2017_3f5ee243,8949185,hedderich-etal-2021-survey,NEURIPS2020_1457c0d6}, generating significant advances in many challenging tasks. The obvious next step was to combine CV and NLP solutions, giving rise to much more intelligent and comprehensive multimodal models that can take advantage of all the research already done in each separate area.

These models, with the emergence of new multimodal datasets, have enabled the creation of solutions to more complex problems such as: image captioning~\cite{mscoco}, visual question answering in images~\cite{Antol_2015_ICCV,s10462-020-09832-7} and videos~\cite{Tapaswi_2016_CVPR}, visual relationship detection~\cite{lu2016visual}, visual dialogue~\cite{Das_2017_CVPR}, visual reasoning~\cite{suhr-etal-2017-corpus,suhr-etal-2019-corpus}, visual entailment~\cite{arxiv.1901.06706}, among others~\cite{UPPAL2022149,bayoudh2021survey,jair.1.11688,arxiv.2202.10936}. Despite their success, most models are designed for specific tasks, which leads to low transferability. Pre-training huge models (with a lot of parameters) on large-scale general-purpose datasets and then finetuning them for specific tasks is currently the most used technique to increase transferability~\cite{arxiv.2202.10936}.

Pre-training was first successfully used in the area of computer vision~\cite{Simonyan15}, becoming the basis for several solutions, mainly in the area of image segmentation~\cite{NIPS2015_14bfa6bb,Redmon_2016_CVPR,Li_2018_ECCV}. After the emergence of the transformers~\cite{NIPS2017_3f5ee243} and subsequently the BERT model~\cite{devlin-etal-2019-bert}, the pre-training and finetuning paradigm became prevalent in the field of NLP~\cite{51115,hedderich-etal-2021-survey,arxiv.2112.11446,arxiv.2201.11990,arxiv.2203.15556,arxiv.2204.02311,arxiv.2205.01068,J1WhitePaper,20-074,ijcai2021-612}. Due to the success of pre-trained models in the areas of CV and NLP alone, many authors have chosen to pre-train large-scale models using both modalities, called visual-linguistic pre-trained models (VL-PTMs)~\cite{CHAI2021100134,arxiv.2204.07356}.

One problem arising from training models with a large amount of unfiltered data is that it absorbs data biases. These biases can take many different forms~\cite{ai_goofs}, from favoring brands~\cite{chuang-yang-2022-buy}, gender~\cite{amazon_ai_gender_bias,sun-etal-2019-mitigating,50755,matthews-etal-2021-gender}, sexual orientation~\cite{matthews-etal-2021-gender}, race~\cite{sciadv.aao5580,pmlr-v81-buolamwini18a,field-etal-2021-survey}, and more, and can be passed on to various applications~\cite{3531146.3533138}. Studies have been done to mitigate and understand these factors~\cite{arxiv.2112.04359,3531146.3533088}, including research best practices~\cite{blodgett-etal-2020-language,partnership_on_ai,google_responsible_innovation}, analyses on making these models available~\cite{arxiv.1908.09203} and the creation of new workshops\footnote{Bias, Ethics and Fairness in Artificial Intelligence: Representation and Reasoning - BEFAIR$^2$ -- \url{https://befair2.org/}}\footnote{Workshop on BIAS and Fairness in AI -- \url{https://sites.google.com/view/bias2021/}}\footnote{ACM SIGKDD Workshop on Ethical Artificial Intelligence: Methods and Applications -- \url{https://charliezhaoyinpeng.github.io/EAI-KDD22/}}\footnote{ACM Conference on Fairness, Accountability, and Transparency (ACM FAccT) -- \url{https://facctconference.org/}}. These biases can even be something simpler such as an artificial intelligence (AI) learning that a dog is always on the ground, while a bird is usually off it.

\begin{figure*}[!htb]
    \centering
    \includegraphics[width=13.0cm]{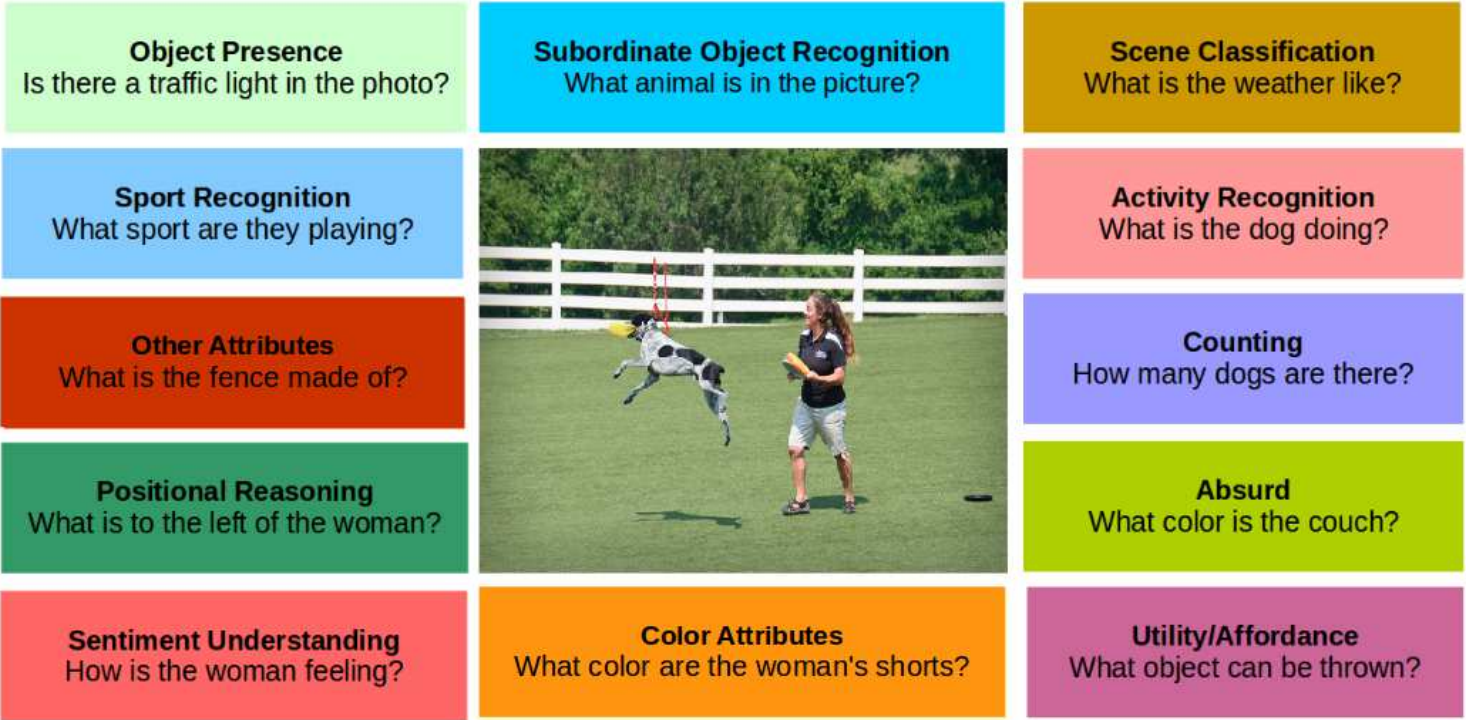}
    \caption{Examples of question types contained in the Task Driven Image Understanding Challenge (TDIUC) dataset~\cite{Kafle_2017_ICCV}.}
    \label{fig:example_tdiuc}
\end{figure*}

One way to minimize incorrect behavior and biased learning from AI is to validate the concepts learned. During model training, metrics are used to evaluate model performance, but they do not exactly capture this kind of detail. Visual question answering (VQA) datasets have a wide variety of question types, ranging from scene classification, object detection, action recognition, sentiment analysis, color attributes, and more~\cite{Kafle_2017_ICCV}, as shown in Figure~\ref{fig:example_tdiuc}. However, this variety does not prove the learning of concepts by the model, which was exposed to a large range of them in small individual amounts.

In this work, we opted to take a different approach: to thoroughly probe the model's learned concepts by extensive questioning it about individual concepts. We created a framework for generating datasets with a large number of questions about a single concept, called \scrapbook. This framework allows us to generate datasets with different characteristics, such as the number of objects in the image, the type of objects, the type of questions, and the questions themselves.

The datasets generated using this framework are designed to validate the model's understanding of basic concepts, such as absolute and relative positions, object recognition, and attribute identification. By focusing on these fundamental concepts, we aim to ensure that the models can reliably understand and process the essential elements of visual information before tackling more complex tasks. By creating different levels of complexity in questions, it is possible to evaluate the model's understanding of the concepts in a more incremental way.

We conducted experiments with three different models, MobileVLM-V2~\cite{chu2024mobilevlm}, TinyGPT-V~\cite{yuan2024tinygptv}, and MiniGPT-4~\cite{zhu2024minigpt}, to evaluate their performance on the generated datasets. The results revealed that while the models showed proficiency in recognizing the presence of objects and counting them, they struggled with understanding positional information and handling questions with additional conditions. Specifically, the MobileVLM-V2 model exhibited significant answer disagreements and plausible wrong answers. The TinyGPT-V model demonstrated a bias toward affirmative responses and struggled with questions involving geometric shapes and positional information. The MiniGPT-4 model exhibited the best performance; nevertheless, it encountered issues with plausible wrong answers and answer disagreements, particularly when absurd questions were introduced.

\section{Related Concepts}
\label{sec:related_concepts}

This section presents the concepts related to the work presented in this paper. It begins with an overview of visual question answering (VQA), interpretability, and explainability, which are the main motivations of this work. It then presents the idea of concept learning, which forms the basis for the methodology used to validate the datasets generated by the \scrapbook\ framework.

\subsection{Visual Question Answering}

Visual question answering (VQA) is a multidisciplinary research problem that has piqued the interest of researchers in the fields of computer vision and natural language processing. The objective in this domain is to develop systems capable of generating natural language responses to visual representations, such as images or videos, when a natural language question is posed. Achieving accurate answers often involves applying common sense reasoning and leveraging world knowledge about the visual content of the scene~\cite{s10462-020-09832-7}.

The potential for significant real-world applications has been a primary factor in attracting researchers to this field. A particularly socially relevant and direct application involves assisting individuals with visual impairments in communicating through a natural language interface with images. The use of text-to-audio and audio-to-text systems facilitates a highly natural interaction through visual question answering~\cite{j.imavis.2021.104327}. Additionally, these systems can interface with, organize, or navigate substantial volumes of typically unstructured visual data, thereby enhancing image retrieval capabilities. An extension of VQA known as visual dialogue has been demonstrated to support natural language instructions for robots and assist with navigation challenges~\cite{yusuf2022analysis}.

In comparison to other visual-linguistic tasks, such as the creation of captions for images, VQA is regarded as a ``complete AI task'' due to its requirement of multimodal knowledge that transcends a single domain. The complexity of VQA stems from the potential inclusion of diverse sub-tasks within the domain of computer vision, such as identifying the most salient object in an image irrespective of its position (i.e., object recognition), detecting the number of instances of a particular object class, and determining the spatial relationship between two objects~\cite{j.imavis.2021.104327}. A selection of questions and their corresponding sub-tasks is illustrated in Figure~\ref{fig:example_tdiuc}.

In addition to visual content, images or videos can contain a lot of other semantic information that can be used for visual question answering tasks. It is quite common to find texts in real scenes, such as in traffic signs, license plates, commercial stores' names, and products names on receipts. Therefore, some early approaches considered VQA on images as a visual Turing test, where the expectation was to incorporate human-level abilities to semantically access visual information in order to answer diverse questions~\cite{pnas.1422953112}.

Most initial researches treated the task as a classification, in which the model must select a correct answer from among two possibilities (binary classification, such as ``yes'' or ``no'') or from a set of more common answers (multiple choice). Although the VQAv2~\cite{Goyal_2017_CVPR} remains the reference dataset for VQA\footnote{This can be seen on the Papers With Code website -- \url{https://paperswithcode.com/dataset/visual-question-answering-v2-0}}, new datasets have emerged, sometimes focusing on specific sub-tasks, such as 360-degree images~\cite{Chou_2020_WACV}, remote sensing images~\cite{9088993}, answers involving optical character recognition (OCR)~\cite{Marino_2019_CVPR,Methani_2020_WACV,Singh_2021_CVPR,Mathew_2021_WACV,Mathew_2022_WACV}, questions in multiple languages~\cite{3404835.3463257}, artistic or abstract representations~\cite{978-3-030-66096-3_8,2021_d3d94468}, pathological images~\cite{he-etal-2021-towards}, visual dialog~\cite{Das_2017_CVPR,kottur-etal-2019-clevr,agarwal-etal-2020-history,dong-etal-2021-visually}, external knowledge~\cite{Shah_Mishra_Yadati_Talukdar_2019}, first-person videos~\cite{Fan_2019_ICCV}, among several others.

Recent advancements in various language generation tasks have prompted a shift in the focus of research in VQA towards the generation of less constrained responses. Some work models VQA as a text generation task~\cite{pmlr-v139-cho21a}, where the answers are of an arbitrary (free-form) nature, depending on the question. This approach removes the constraint imposed by the fixed answer vocabulary and allows it to better generalize to real-world scenarios. The system's objective is to ascertain the most accurate response, whether in a concise phrase or a series of brief sentences~\cite{s10462-020-09832-7,j.imavis.2021.104327}.

The main tasks within the domain of visual question answering (VQA) involve diverse forms of media and/or aggregate knowledge. More specifically:

\begin{myitemize}

\item images, often simply called VQA or Image Question Answering (IQA), is the most prevalent task found in the literature~\cite{Goyal_2017_CVPR,s10462-020-09832-7,j.imavis.2021.104327,Chou_2020_WACV,9088993,Marino_2019_CVPR,Methani_2020_WACV,Singh_2021_CVPR,Mathew_2021_WACV,Mathew_2022_WACV}. There is also the possibility of using image sets~\cite{8603827};

\item videos, known as Video Question Answering (also found as VideoQA or ViQA))~\cite{lei-etal-2018-tvqa,Fan_2019_ICCV,castro-etal-2020-lifeqa,garcia2020knowit,arxiv.2110.13395,Gupta_2022_WACV};

\item dialogues (Visual Dialogue, or VD for short), in which the main goal is to automate conversations about images (or videos) between machines and humans~\cite{Das_2017_CVPR,kottur-etal-2019-clevr,agarwal-etal-2020-history,dong-etal-2021-visually};

\item knowledge bases (Knowledge Base Visual Question Answering, or KB-VQA), where external knowledge, beyond the information available in the image, is used to deeply understand the content of the scene, allowing answering a broader and deeper set of real-world questions~\cite{Shah_Mishra_Yadati_Talukdar_2019};

\item metadata (or extra information) such as an image caption~\cite{Zhou_Palangi_Zhang_Hu_Corso_Gao_2020} or a descriptive paragraph~\cite{kim-bansal-2019-improving}.
\end{myitemize}


\subsection{Interpretability and Explainability}

Fairness in machine learning has recently raised many questions about the transparency of algorithms commonly employed as closed boxes. Examples such as the Correctional Offender Management Profiling for Alternative Sanctions (COMPAS)~\cite{sciadv.aao5580} case show the need to review AI results. In COMPAS, racial disparities in parole decision deliberations were created even though the data used did not include the race of the individuals. This brought to light several biased assumptions made by the models due to biases present in the data, which previously went unnoticed~\cite{UPPAL2022149}.

The successful adoption of artificial intelligence depends on how much decision makers can understand and trust it~\cite{978-3-030-82196-8_2}. A survey conducted by~\citet{dietvorst2015algorithm} highlighted a phenomenon that the authors called algorithm aversion. In their study, they showed that people are especially reluctant to predictive algorithms after seeing them work, even when they outperform results given by humans. This is because people lose trust in algorithmic predictors more quickly than humans after seeing them make the same mistake.

In several applications, AI makes decisions that directly affect human welfare, such as in autonomous cars, credit applications, or the criminal justice system~\cite{sciadv.aao5580,978-3-030-82196-8_2}. For people to trust artificial intelligence, we need models that can summarize the motivations of network behavior and provide insights into the causes of its decisions. Explainable Artificial Intelligence (X-AI) proposes to create more explainable, high performing models. Current discussions in X-AI research suggest that explainability will lead to more realistic expectations, increased reliability, and fairer decisions~\cite{978-3-030-82196-8_2}.

Interpretability and explainability are two terms from this area that are used synonymously most of the time, addressing the explainability of the system. Although the term ``interpretability'' is widely used in the literature, it lacks a common definition. The same goes for the term ``explanation'', in particular what a ``valid explanation'' is and how to measure its quality~\cite{978-3-030-82196-8_2,8631448}. This arises from the difference between causality and correlation. In the COMPAS model example, although it is not clear whether race is used as a decision criterion, it is still strongly correlated with the decision~\cite{sciadv.aao5580,8631448}. While these terms have been used interchangeably, \citet{8631448} argues that there are important reasons for distinguishing between them. Explainable models are interpretable by default, but the reverse is not always true.


\subsubsection{Interpretability}

Although loosely defined as understanding what a model did (or could have done), \citet{8631448} define interpretability as ``the ability to explain or present in terms understandable to a human being.'' This requires understanding the cause-and-effect relationship of certain learned factors and responding to the ``what'' inherent to the model~\cite{UPPAL2022149}. Model interpretability can be understood as how easy it is for a person to predict the output of the model for new inputs based on previous outputs. This concept is called simulatability~\cite{jair.1.13200}.

The goal of interpretability is to describe the internal components of a system in a way that is understandable to humans. The success of this goal is linked to the cognition, knowledge, and biases of the user: for a system to be interpretable, it must produce descriptions that are simple enough, using a vocabulary understandable for a person to comprehend~\cite{8631448}.

\citet{3375627.3375810} argue that interpretability is not an intrinsic property of a model, but a prospective property for a specific audience: engineers, users, or impacted people, each with different goals and requirements regarding interpretability. The authors also state that interpretability needs to be measured against these goals, but currently there is no clear definition or evaluation criteria for this.


\subsubsection{Explainability}

While interpretability is a considerable first step, AIs must have the ability to defend their actions, provide relevant answers to questions, and be auditable~\cite{8631448}. Explainability focuses on explicitly describing facts concerning the ``why'' and ``how''~\cite{UPPAL2022149}.

Explanations of how neural networks work have focused primarily on explaining the processing of the data or the internal representation of the data. For data scientists, the quality of the explanation depends on the degree to which the actual decision criteria are shown. A high quality explanation for wrong decisions would immediately show the scientist that the solution is of low quality, preferably providing an indication of systematic error (e.g., due to biases in the input data)~\cite{978-3-030-82196-8_2}.

For end users, the quality of the explanation depends on the degree to which an AI decision can be evaluated based on it. An explanation of how the model works is secondary compared to information that serves as the basis of a decision for a human validation~\cite{978-3-030-82196-8_2}. For example, a model is presented in the work of~\citet{2939672.2939778} that differentiated between photos of huskies (breed of dog) and wolves by the presence of snow in the background, since wild animals are usually photographed in the nature. To a human, this may be a poor explanation, although it correctly reveals the inner workings of the model. Therefore, it is necessary to distinguish the quality of the explanations according to the different intended purposes.

\citet{arxiv.1711.07414} warns that reliance on human evaluations can lead researchers to create persuasive systems rather than transparent ones. \citet{8631448} believe that it is fundamentally unethical to present a simplified description of a complex system to increase reliability if the limitations of this description cannot be understood by users and, worse, if the explanation is optimized to hide undesirable attributes of the system. Such explanations are misleading and can result in the user justifiably reaching dangerous or unfounded conclusions.


\subsubsection{Discussion and Direction}

A summary of the differences between interpretability and explainability is presented in Table~\ref{tab:interpretability_explainability}. The objective of our work is to achieve \textit{interpretability}, as we aim to provide simulatability by probing the model's learned concepts in an incremental manner. This objective will be achieved through a meticulous validation process, commencing with fundamental concepts such as object recognition and attribute identification. Subsequently, the model will be advanced to more intricate concepts, including absolute and relative positioning, and the combination of multiple concepts. An extensive dataset comprising a wide linguistic variation will be employed to assess the model's performance and ascertain its limitations.

\begin{table*}[!htb]
    \setlength{\tabcolsep}{5mm}
    \centering
    \caption{Key differences between Interpretability and Explainability.}
    \label{tab:interpretability_explainability}
    \small
	\begin{tabular}{lll}
        \toprule
        \textbf{Aspect} & \textbf{Interpretability} & \textbf{Explainability} \\
        \midrule
        \multirow{2}{*}{Focus} & Understanding the what & Understanding the why \\
        & of the model's behavior & and how of decisions \\
        \hline
        \multirow{2}{*}{Depth} & Focuses on simplicity & Aims to provide detailed \\
        & and simulatability & reasoning for decisions \\
        \hline
        \multirow{2}{*}{Goal}  & Making the model's internal & Supporting validation, \\
        & components understandable & auditing, and debugging \\
        \hline
        \multirow{2}{*}{Audience Needs}  & Adaptable to audience & Tailored to specific evaluation \\
        & knowledge and biases & and validation goals \\
        \hline
        \multirow{2}{*}{Challenges}  & Lack of clear & Risk of creating persuasive \\
        & evaluation criteria & but misleading systems \\
        \bottomrule
    \end{tabular}
\end{table*}


\subsection{Learning Concepts}

\begin{figure*}[!htb]
    \centering
    \subfloat[][Question: what time of the day is this? \\ Answer: it is daytime.]{\includegraphics[width=0.45\textwidth]{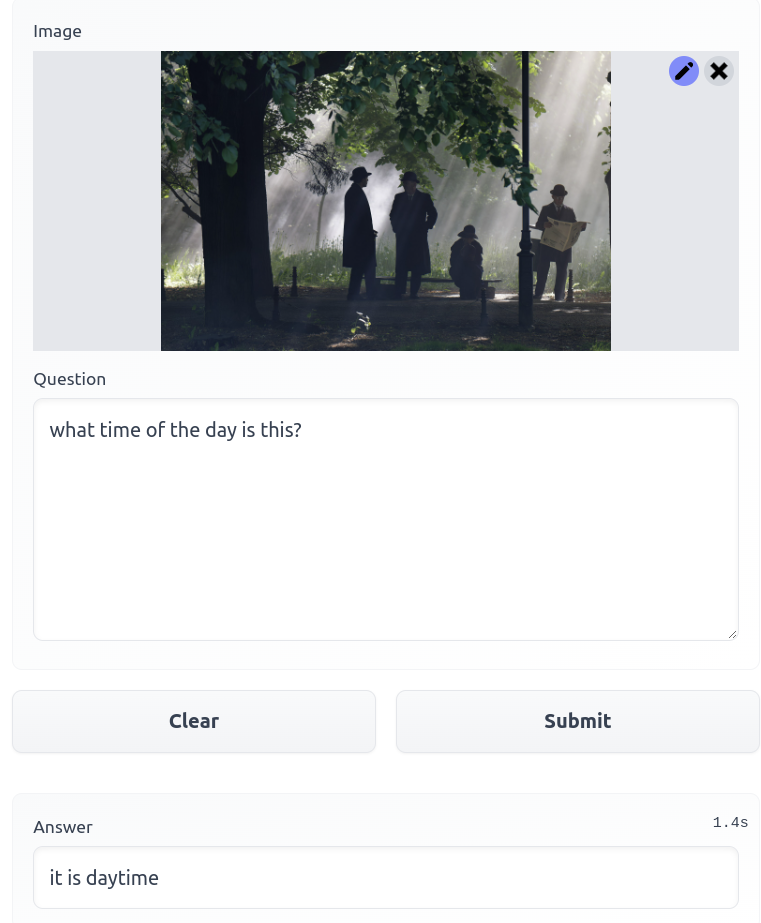} \label{fig:ofa_day}}
    \hfill
    \subfloat[][Question: how many people are standing? \\ Answer: 4.]{\includegraphics[width=0.45\textwidth]{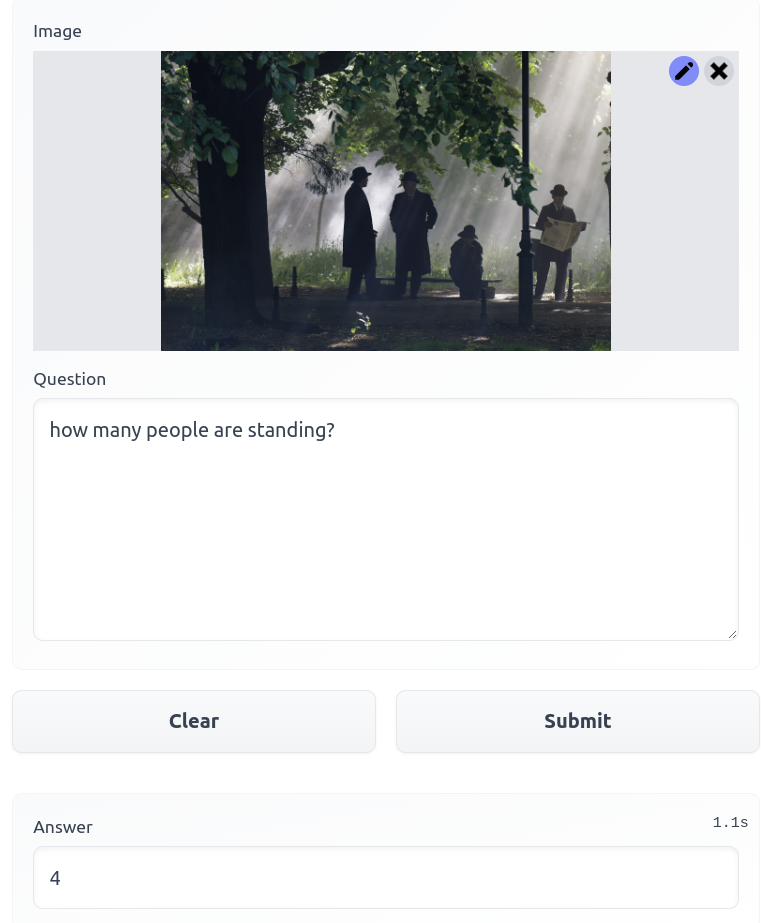} \label{fig:ofa_standing}}
    \caption{Corresponding questions and answers from the same image provided by the OFA model~\cite{pmlr-v162-wang22al}.}
    \label{fig:ofa_answers_people}
\end{figure*}

Despite significant advancements in natural language processing, models continue to demonstrate failures in responses to what are commonly regarded as simple questions, as illustrated in Figure~\ref{fig:ofa_answers_people}. This suggests the possibility that the model has acquired an ``unanalyzed'' language representation, akin to the manner in which a child acquires language during its developmental stages. \citet{bowerman1982starting} conceptualizes the progression of language development as a transition from ``unanalyzed'' to ``analyzed'' representations of language:

\blockquote{It is the hypothesis that the child is capable of acquiring bits and pieces of language and using them completely correctly without being aware of how they are related to each other... forms that are analyzable within the adult system as complex -- that is, consisting of subunits with independent combinatorial potential -- can be used correctly by a child even though he is unaware of their internal structure.}

In essence, AI has acquired the ability to utilize language, yet it remains incapable of comprehending it. To respond to a given query correctly, it is imperative that the AI system develop an understanding of the question's content and engage in deliberative reasoning processes. This capacity to deliberate, or metalinguistic awareness, can be conceptualized as a form of metacognition, defined as an individual's capacity to direct attention toward, reflect upon, and evaluate language as an entity in its own right~\cite{SCHONPFLUG20011171}. The following metalinguistic skills are considered essential: word awareness, syntactic awareness, and pragmatic awareness, which refer to knowledge analysis and control of cognitive operations involving language processing. According to \citet{bialystok1986children}:

\blockquote{Metalinguistic ability, then, is not a discontinuous linguistic achievement, but an emerging ability that reflects gradual progress with these underlying skills. Problems demanding more advanced levels of these underlying skills will be solved later than those that are less demanding.}

Therefore, the model must first demonstrate an understanding of more fundamental concepts before we can infer its comprehension of more complex ones.

The prevailing approach in the literature involves trying to infer the extent of a word by examining its similarities across a set of diverse sentences. A common assumption underlying this approach is that the range of contexts observed during training exhausts the scope of the word. This assumption serves as the foundation for training large language models with billions of parameters, developed by major companies and implemented with extensive datasets extracted from the Internet, like Jurassic-1~\cite{J1WhitePaper} from AI21 Labs; ELMo~\cite{peters-etal-2018-deep} and UNIFIEDQA~\cite{khashabi-etal-2020-unifiedqa} from Allen Institute for AI; Chinchilla~\cite{arxiv.2203.15556} and Gopher~\cite{arxiv.2112.11446} from DeepMind; BERT~\cite{devlin-etal-2019-bert}, GLaM~\cite{pmlr-v162-du22c}, LaMDA~\cite{51115}, PaLM~\cite{arxiv.2204.02311} and T5~\cite{20-074} from Google; OPT~\cite{arxiv.2205.01068} from Meta; DIALOGPT~\cite{zhang-etal-2020-dialogpt}, GODEL~\cite{peng2022godel}, MT-NLG~\cite{arxiv.2201.11990} and UniLMv2~\cite{unilmv2} from Microsoft; Megatron-LM and NeMo-Megatron~\cite{arxiv.2205.05198} from NVIDIA; GPT~\cite{radford2018improving}, GPT-2~\cite{radford2019language}, GPT-3~\cite{NEURIPS2020_1457c0d6}, WebGPT~\cite{arxiv.2112.09332} and InstructGPT~\cite{arxiv.2203.02155} by OpenAI; and Bloom~\cite{scao2022what} by a collaboration between HuggingFace and other companies. However, it should be noted that several words in our vocabulary are associated with tied knowledge. For instance, the word ``gallop'' signifies a specific movement of a particular animal, the horse. Other words encode conventional events involving sets of objects, acts, and places, and could also include regional or cultural customs. For example, ``birthday party'' encodes an event that includes songs, cakes with candles, pointy hats, group games, among others~\cite{HUTTENLOCHER198763}.

Going in the opposite direction, this work aims to validate the understanding of simpler concepts by an artificial intelligence model, initially focusing on the recognition of objects, their absolute and relative positions. This will be done using a large dataset with a wide linguistic variation, which can be easily extended.

\begin{figure*}[!htb]
    \centering
    \subfloat[][Question: where are the cats? \\ Answer: on a piece of wood.]{\includegraphics[width=0.45\textwidth]{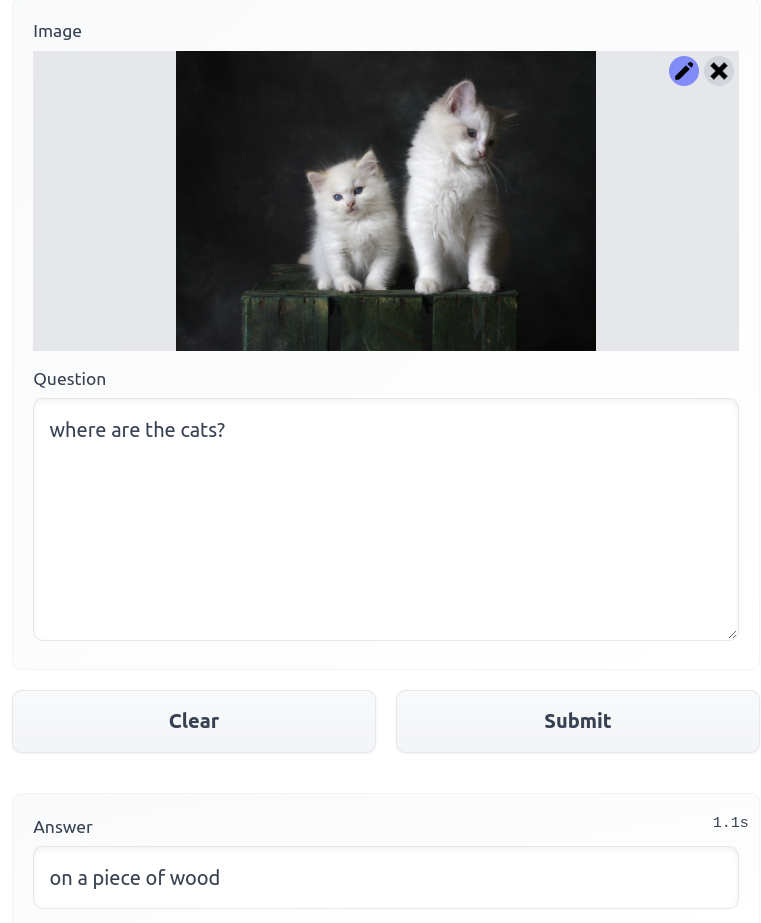} \label{fig:ofa_cats}}
    \hfill
    \subfloat[][Question: where are the rabbits? \\ Answer: next to each other.]{\includegraphics[width=0.45\textwidth]{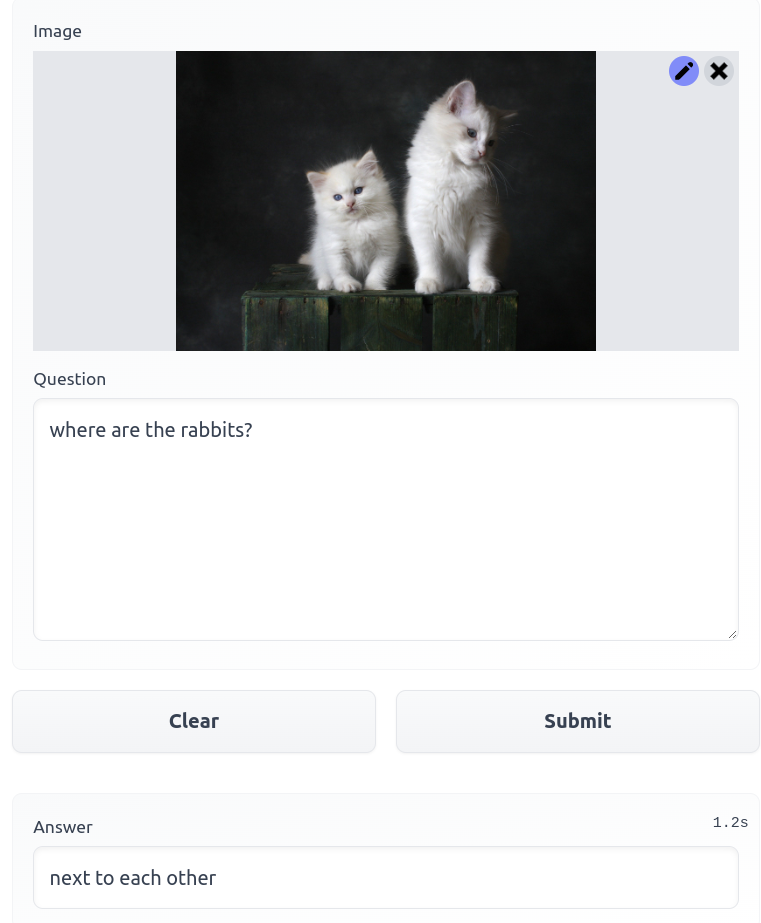} \label{fig:ofa_rabbits}}
    \caption{Corresponding questions and answers from the same image provided by the OFA model~\cite{pmlr-v162-wang22al}.}
    \label{fig:ofa_answers_cats}
\end{figure*}

Like a child, an AI model assimilates the biases embedded in the training set, which often results in the formation of incorrect assumptions. During the experiment conducted with children by~\citet{HUTTENLOCHER198763}, one child responded with the word ``ball'' when he saw his mother's tennis racket. While it is not plausible to argue that the racket is part of the meaning of ``ball'', for this child, as for all the children in the study, balls appeared frequently in a wide variety of contexts. In some cases, the mother's tennis racket was present, and a ball was often observed in its vicinity. A similar phenomenon is likely to have happened during the training of the OFA~\cite{pmlr-v162-wang22al} model, where images of rabbits were presented to the model along with questions about them, and some responses might involve them being nearby, which would explain the answer given to the question in Figure~\ref{fig:ofa_rabbits}.

\begin{figure}[!htb]
    \captionsetup[subfigure]{labelformat=empty}
    \centering
    \subfloat[][Source: COCO dataset~\cite{mscoco}]{\includegraphics[width=0.225\textwidth]{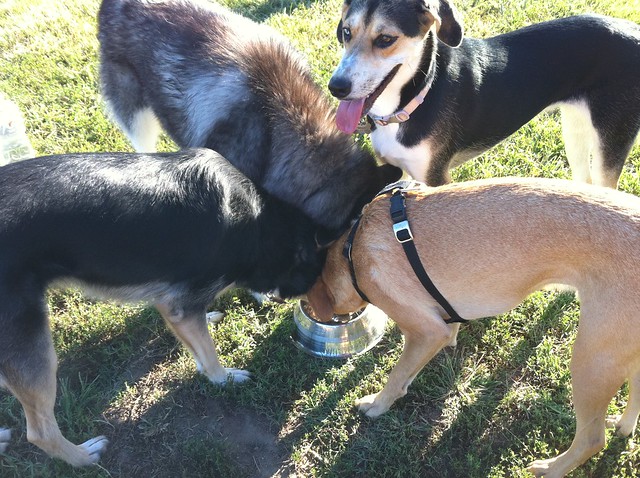}}
    \hspace*{0.1cm}
    \subfloat[][Source: COCO dataset~\cite{mscoco}]{\includegraphics[width=0.225\textwidth]{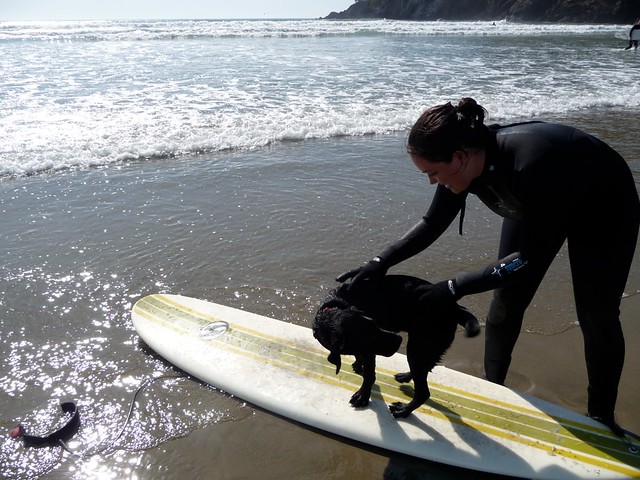}} \\ 

    \subfloat[][Source: Know Your Meme~\cite{knowyourmeme_dog}]{\includegraphics[width=0.225\textwidth]{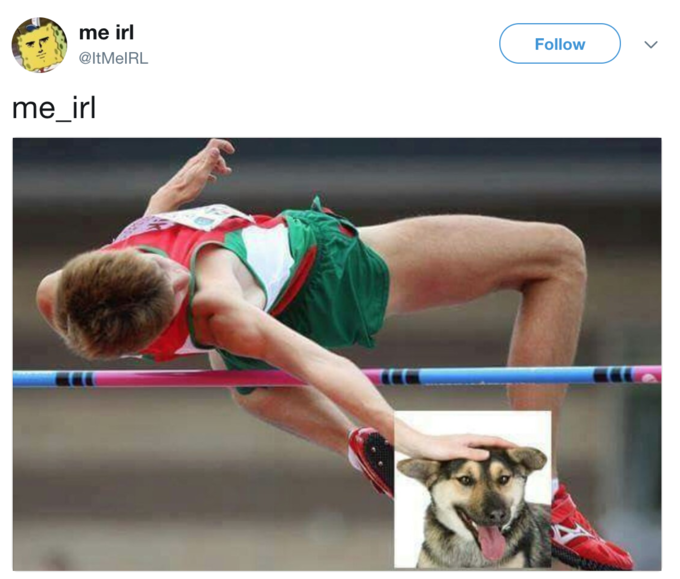}}
    \hspace*{0.1cm}
    \subfloat[][Source: Dicionário Popular~\cite{meme_hipocrisia}]{\includegraphics[width=0.225\textwidth]{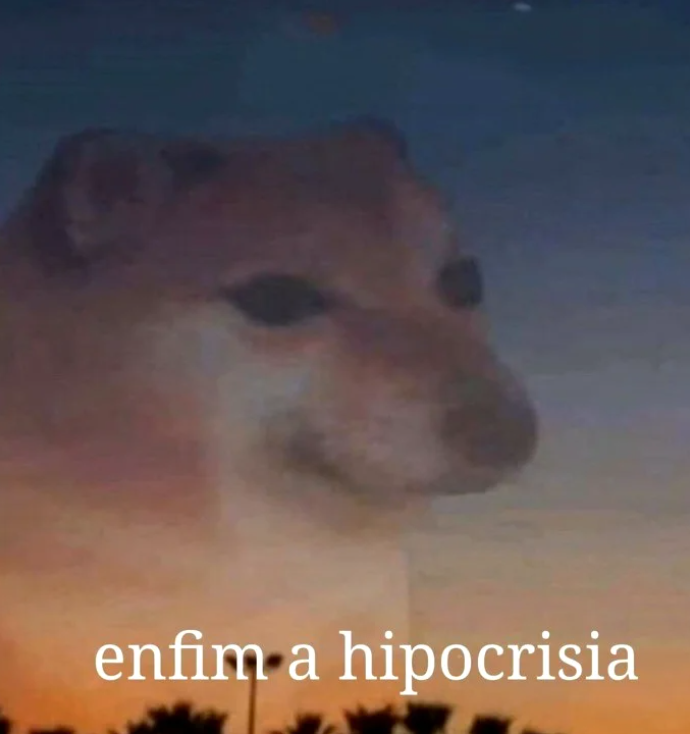}}
    \caption{Examples of images in which a dog can still be identified by a human.}
    \label{fig:dogs}
\end{figure}

A wide range of objects can be recognized regardless of context. While certain objects are defined in terms of their context (for example, a window on a roof is considered a skylight), others, such as a dog, are identifiable even when situated in a grassy background, on a beach, or in the sky, as illustrated in Figure~\ref{fig:dogs}. In light of these observations, we have developed an artificial dataset comprising objects removed from their context and overlaid on diverse background images. This approach enables greater control over the positioning of the objects, thereby ensuring a diverse array of configurations and combinations that effectively convey the desired concepts.

\section{Related Work}

Many works have been published aiming at the explainability of artificial intelligence models for VQA. While some approaches chose to create datasets to validate or support the decisions made, other works preferred to follow the path of trying to open the closed box that are the models, using interpretability/explainability methods. Some authors have decided that the best way to explain the result is to train the solution from the beginning thinking about explaining how it works. This chapter discusses these different groups of solutions.


\subsection{VQA Datasets that Validate Concepts}

\subsubsection{SHAPES}

To test the ability of their approach to handle more complex questions, \citet{Andreas_2016_CVPR} introduced a new dataset of synthetic images consisting of abstract shapes and complex questions involving spatial relationships, reasoning about sets, and recognition of shapes and attributes. The SHAPES dataset consists of questions with a high level of difficulty about arrangements of colored shapes. The questions contain two to four attributes, object types or relationships. It contains 244 unique questions, matching each question with 64 different images (for a total of 15,616 unique question/image pairs, with 14,592 on the training set and 1,024 on the test set).

To eliminate guesswork on the part of the network, all questions have a ``yes'' or ``no'' answer, but good performance requires the system to learn to recognize shapes and colors and to understand spatial and logical relationships between sets of objects. The concepts used in this work are: find, transform, combine, describe, and measure. Almost all compositional operations take place in the attention space, which means that all modules introduced by the authors receive at least one attention map as input, and half of them generate another attention map. Since the concepts that are used to select modules are only processed internally, but are not produced in the final output, the dataset uses accuracy as a metric.

\subsubsection{R-VQA}

Visual attention-only models have limited performance when further understanding of relations is required. In order to use information about them in the VQA task, \citet{3219819.3220036} proposed to learn evidence about relations by selecting them through semantic attention. To accomplish this, the Relation-VQA (R-VQA) dataset was created from the Visual Genome set by adding facts that prove the answers. These facts are categorized into three types: entity concept, entity attribute, and entity relation.

Entity concept involves verifying the existence of an object, in the form \textit{there is <object>}. Entity attribute requires a perception of object characteristics, such as color. It takes the form \textit{<subject> is <attribute>}. Entity relationship means identifying the relationship between two objects, such as positioning. It is presented as \textit{<subject> <relation> <object>}.

Only images from the Visual Genome that contain QA pairs and semantic knowledge are used, and this knowledge is treated as eligible to become facts. After calculating the relevance between each QA pair and its fact candidates, and removing candidates with matching scores below a certain threshold, the fact with the highest score is chosen as true (ground truth). To ensure the quality of the selected facts, the authors recruited people to validate that the generated facts are correctly related to the provided QA pair.

To balance the quality of the facts and the amount of data, the authors chose 0.3 as the scoring threshold, resulting in a dataset of 198,889 samples with an average combined accuracy of 90.8\% for all question-answer-fact triples. That is, for each image in the dataset, a question and a correct answer corresponding to the content of the image are provided, as well as a relation fact that provides a good explanation for the question and answer. There are 5,833 relations, 2,608 subjects, and 6,914 unique objects, covering a wide range of topics.

A relation fact describes semantic information, which gives a VQA model a better understanding of the image. For this reason, \citet{3219819.3220036} developed a relation fact detector to obtain a fact referring to both the question and the semantic content of the image. The authors provide recall@1, recall@5, and recall@10 as evaluation metrics for the detector, where recall@$k$ is defined as the fraction of times the correct relation fact is among the top $k$ predicted facts. Accuracy is used to validate the answer.

\subsubsection{Considerations} 

SHAPES is an artificial dataset that emphasizes spatial relationships, logic about sets, and recognition of shapes and colors. It contains a limited variety of objects and a reduced size, limited to abstract shapes. Although it includes the notion of certain concepts, the variety of questions is limited and each question is formed by at least two attributes. The R-VQA dataset, on the other hand, is built from real images taken from the Visual Genome set. The use of authentic images introduces limitations on the variety of objects and their applications, which can introduce biases into the model, such as objects that invariably appear within a specific context. Both datasets feature complex questions, thereby emphasizing the collective treatment of concepts rather than their individual examination. The \scrapbook\ datasets developed in this study were designed to emphasize elementary concepts with a substantial linguistic variety prior to progressing to more intricate concepts that depend on prior validated concepts.


\subsection{VQA Datasets with Visual Grounding}

\subsubsection{Visual7W}

While previous work has established a weak association between questions, answers, and images, \citet{Zhu_2016_CVPR} created a new dataset that establishes a semantic connection between textual descriptions and image regions according to the visual grounding of objects. The QA pairs in their Visual7W dataset are a subset of the Visual Genome, more specifically the COCO dataset. It also includes additional annotations such as visual grounding of objects, multiple answer choices, and human testing, making it a complete set for evaluation and analysis.

Inspired by the Five W's technique, an acronym representing the essential questions a journalist asks when writing an article, the authors categorized questions with textual answers as questions that reveal (what, where, when, who, why, and how) and with visual answers as questions that point (which). The dataset consists of 47,300 images, 561,459 object bounding boxes (an average of 12 per image), and 36,579 categories. The visual grounding of objects is primarily used to analyze the attention behavior of the model, and is not used as a metric.

\subsubsection{VQA-HAT}

\citet{das-etal-2016-human} designed a game-inspired interface to collect attention maps of where people look to answer questions. They presented participants with a blurred image and a question about it, and asked them to remove as little blur from the image as possible in order to answer the question correctly. Using images and questions from the VQA dataset, the authors created the VQA-HAT (Human ATtention) dataset, with human attention maps collected for 58,475 and 1,374 question-image pairs for training and validation, respectively. Since the VQA-HAT annotations are collected by having humans ``unblur'' the images, they can introduce noise if irrelevant regions are unblurred.

By using the rank correlation metric, i.e., reducing the attention maps generated by the model and humans to 14$\times$14, ranking the pixels according to their spatial attention, and computing the correlation between these two ranked lists, the authors showed that VQA attention models do not appear to ``look'' at the same regions as humans when producing a response. This visual correlation increases as attention-based VQA models become more accurate.

\subsubsection{VQS}

By linking the instance segmentations provided by the COCO dataset to the QA pairs in the VQA dataset, \citet{Gan_2017_ICCV} unified human supervision between previously separate tasks (semantic segmentation and VQA), creating the Visual Questions and Segmentation Answers (VQS) dataset. The authors retain only images that contain at least three instance segmentations, and ask study participants to draw bounding boxes when the visual answer is only part of a segment. The resulting data make up the VQS dataset, with a total of 37,868 images, 96,508 questions, 108,537 instance segmentations, and 43,725 bounding boxes.

The authors show that by making the model perceive different regions of the image using the links between QA and segmentation, they can significantly improve the performance of a simple MultiLayer Perceptron (MLP) model in the multiple-choice VQA task. They also propose the Question-Focused Semantic Segmentation (QFSS) task, which aims to generate pixel-by-pixel segmentations to visually answer questions about images, to reduce possible shortcuts in model learning. The resulting segmentation mask is validated using the IoU metric.

\subsubsection{Visual Genome}

To understand images in more detail, \citet{s11263-016-0981-7} believe that three elements are essential: a linguistic foundation of visual concepts, a more complete set of descriptions and QAs for each image based on multiple regions of it, and a formal representation of the components of an image. To capture this diverse information from the visual world, the authors presented the Visual Genome dataset, which contains over 100,000 images from the intersection of YFCC100M and COCO. Each image has over 42 descriptions for different regions of the image, with an average of 21 objects, 18 pairwise relationships between them, 18 attributes, and 17 question-answer pairs based on the descriptions.

The Visual Genome dataset consists of seven main components: region descriptions, objects, attributes, relations, region graphs, scene graphs, and QA pairs. Each image is represented by a scene graph, which formally expresses the relationship between all objects in the image, and has been shown to improve results in semantic image retrieval tasks. All objects, attributes, and relationships are converted to their corresponding WordNet identifiers (called synset IDs). Despite acknowledging the absence of effective methods for evaluating the extent to which models comprehend images, the authors employ accuracy as their metric.

\subsubsection{VQA-X}

\citet{Park_2018_CVPR} proposed VQA and activity recognition as case studies for explanations, as they are important and challenging visual tasks with interesting properties for explainability. To this end, they present two new datasets with multimodal human-annotated explanations, one for activity recognition (ACT-X) and one for visual question answering (VQA-X), and a ``pointing and justifying'' (PJX) model that justifies the given answer by pointing to parts of the image and generating a textual explanation.

The VQA-X dataset consists of 28,180 images, 32,886 QA pairs (13,921 unique questions and 1,236 unique answers), and 41,817 explanations. The explanations are provided by humans and collected via Amazon Mechanical Turk. Annotators are given an image and an answer, and are asked to segment objects and/or regions that best justify the answer.

The metrics BLEU-4, METEOR, ROUGE, CIDEr, and SPICE, which measure the degree of similarity between generated and expected sentences, are used to evaluate the textual justifications. Human evaluation was also included since the calculated metrics do not always reflect human preference. For the visual pointing task, the authors used the Earth Mover's Distance (EMD) metric, which measures the distance between two probability distributions in a region, and Rank Correlation. Their results showed the limitations of training explanation systems with descriptions, justifying the need for specially created datasets with explanations in mind.

\subsubsection{GQA}

In creating the GQA dataset, \citet{Hudson_2019_CVPR} were inspired by the CLEVR dataset, which consists of questions about the composition of synthetic images. By selecting real images and a semantically diverse set of questions, the dataset serves as a clean benchmark to evaluate models in a more controlled and comprehensive manner, even though the questions are not as natural as in other VQA datasets.

Each image is annotated with a dense scene graph, where objects are connected by relation edges representing actions (verbs), spatial relations (prepositions), and comparisons. Each question is associated with a functional program that lists the sequence of reasoning steps required to reach the answer. Each answer is accompanied by textual and visual justifications that highlight the relevant region within the image.

The GQA dataset consists of 22,669,678 questions on 113,018 images. It has a vocabulary size of 3,097 words and 1,878 possible answers. Although smaller than natural language datasets, it covers more than 85\% and 70\% of the VQA questions and answers, respectively, demonstrating its great diversity. To gain more insight into visual reasoning methods and to point out missing features, five new metrics have been introduced: consistency, validity, plausibility, distribution, and grounding. Some of these are explained in Section~\ref{sec:metrics}.

\subsubsection{AiR}

To facilitate the analysis of human attention in VQA tasks, \citet{978-3-030-58452-8_6} constructed the first eye-tracking dataset collected from humans performing VQA tasks. Based on GQA's balanced validation set, they performed strict quality control on the data used: only images with at least 320$\times$320 pixels, 16 relationships in the scene graph, and where the total area of objects related to the questions does not exceed 4\% of the image were kept. Subsequently, one of the authors manually selected 987 images and 1,422 questions to ensure that the correct answers are accurate and unique. The remaining questions are non-ambiguous and non-trivial, requiring close attention to the scene and an active search for the answer. Eye-tracking data are collected from 20 paid participants, 16 male and 4 female, ranging in age from 18 to 38.

The authors also propose a new quantitative metric, AiR-E, which measures progressive attention and reasoning based on a set of predefined atomic reasoning operations. These operations are drawn from GQA's 127 types of operations, which fully represent all questions. The regions of interest (RoIs) for each operation are jointly determined by the operation type and the scene information. Given the sequence of operations and the annotated scene information, one can follow the reasoning process by starting with all objects in the scene and sequentially applying the operations to obtain the RoIs at each step.

Several interesting findings emerged from this work. The authors observed that object-based attention is much more accurate than spatial attention, that soft attention is better overall, and that transformer attention is better for logical operations (and, or). \textit{query} is the most challenging operation for models, probably due to their inferior recognition ability compared to humans, while \textit{compare} is the most intensive operation for humans, as it requires a lot of attention to multiple objects and thus takes more processing time. The attention performance of humans consistently outperformed that of the models.

Another result of the experiment is the observation that human attention accuracy and task performance are correlated for most operations, suggesting that human performance is more consistent when quality of attention is present. This was also reflected in the AiR-E metric, with humans who responded correctly having significantly higher scores than those who responded incorrectly.

Finally, the study reveals a significant spatio-temporal discrepancy between human and model attention. The spatio-temporal correlations suggest that following the correct order of reasoning operations is important for humans to focus on regions and respond correctly. In contrast, models tend to focus directly on the final RoIs rather than shifting their attention progressively, leading the authors to propose supervising the model's attention throughout the reasoning process to jointly optimize attention, reasoning operations, and task performance.



\subsubsection{TextVQA-X}

\citet{nagaraj-rao-etal-2021-first} pointed out that existing explainability approaches do not include textual image data in the explanations. Therefore, they developed a new dataset, TextVQA-X, consisting of visual rationales and textual explanations with multiple references that are comparable to human interpretations. This helps to justify the decision of the models and provides useful information for diagnosing an incorrect answer.

The authors note that it is arguable whether attention mechanisms actually generate explanations, and that these should be part of the initial design of the interpretable model, not added later. They argue that intrinsic explanations, which may not be true to the model decision but are consistent with human explanations, are beneficial to end users.

The TextVQA-X dataset consists of 11,681 images, 18,096 questions (of which 15,374 are unique), and 67,055 visual and textual explanations. Each question in the TextVQA dataset has 10 human-generated answers, and the accuracy of these answers is measured by voting. The authors evaluate the textual explanations using the standard metrics: BLEU-4, ROUGE, METEOR, and CIDEr. All textual metrics handle multiple references by averaging the individual scores. Visual explanations are evaluated using the Intersection over Union (IoU) metric with a threshold of 0.5.

\subsubsection{CLEVR-Answers}


To enable evaluation of the visual grounding in the CLEVR dataset, \citet{Urooj_2021_CVPR} generated new question-answer pairs with the bounding box labels for all objects on which the answer is based (if any), using the same structure as the original CLEVR. They used the same training and validation scenes and generated 10 new QA pairs for each image. This process resulted in 901,000 bounding boxes (for about 700,000 QA pairs) for the training set and 193,000 boxes (for about 150,000 QA pairs) for the validation set.

To evaluate the location of the correct response (visual grounding), the metrics accuracy, revocation, and F1-score based on overlap and IoU were used, calculated over the bounding boxes for the detected objects, which are considered true positives and real, provided they have a minimum of 50\% spatial overlap. For the textual part, the accuracy metric is used.

\subsubsection{VizWiz-VQA-Grounding}

Based on the VizWiz-VQA dataset, \citet{Chen_2022_CVPR_a} presented the first visual grounding dataset derived from a real use case. Focusing on visual questions asked by blind people who took pictures and asked questions about them to overcome real visual difficulties, this dataset presents different challenges from planned settings, such as: lower quality of pictures, more conversational questions, and the need for different visual skills to arrive at answers.

By eliminating all questions that could not be answered, consisted of multiple sub-questions, involved multiple regions of an image due to ambiguity, could not be visually grounded, or for which most analysts could not agree on a single answer, a total of 9,998 VQAs remained. For these questions, answer grounding segmentation masks were collected from expert participants recruited via Amazon Mechanical Turk, who focused on grounding only the most popular answer for each visual question. This process resulted in a total of 9,998 visual groundings for the 9,998 triples (image, question, and answer).

The authors used IoU and mAP@IoU to measure the similarity of each binary segmentation mask to the actual segmentation. They observed that state-of-the-art (SOTA) models in VQA and VQA-Grounding have difficulty when: the visual evidence occupies a small fraction of the image, the images are of higher quality, and for visual questions that require text recognition skills. They also observed that state-of-the-art VQA algorithms significantly outperform SOTA models of visual VQA grounding pre-trained on the same datasets, achieving improvements of up to 2.5$\ times$. This finding suggests that much of the success of SOTA VQA models is still due to learned biases unrelated to visual evidence.

\subsubsection{Considerations} 

Most of the visually grounded datasets presented are based on real images, which can introduce contextual biases into the models when trained on them. For example, the AI may learn that birds are usually on a blue background (sky), while dogs are not. With the exception of the VQS and VizWiz-VQA-Grounding datasets, which generate masks to segment responses, the datasets provide only bounding boxes, which are not very accurate. Our dataset removes some of the context bias by adding real objects along with their masks to different background images. The fact that we incrementally add the objects to the images helps to verify that the model understands what each object is individually.


\subsection{Explanability / Interpretability Methods for VQA Models}

\subsubsection{Visual Explanation}

The explanation of an answer is as important or more than the answer itself, because it makes the process of answering the question more understandable and comprehensible. \citet{chandrasekaran-etal-2018-explanations} argued that for human-AI interactions to be more effective, humans must also understand the AI's beliefs, knowledge, and peculiarities. In their paper, they evaluate the role of machine-generated explanations in making a VQA model predictable to a human.

The authors considered two tasks that demonstrate the degree to which a human understands an AI: Failure Prediction (FP) and Knowledge Prediction (KP). In FP, participants on Amazon Mechanical Turk were asked to predict whether the model would correctly answer a given question about an image. In KP, they try to predict the exact answer of the artificial intelligence.

The humans were helped to form a mental model of the AI by familiarizing them with its behavior in a ``training'' phase and exposing them to the model's internal states through various forms of explanation. Specifically, the image is accompanied by the spatial attention map (generated using the Grad-CAM technique) and the words of the question, indicating the regions that the model is looking at and reading, respectively. The main results were:

\begin{myitemize}
\item people are actually able to predict successes, failures, and outcomes of the VQA model better than chance;
\item explicitly training humans to become familiar with the model improves its performance, and;
\item existing explanation modalities do not improve human performance.
\end{myitemize}

Existing methods focus on using visual information without considering whether it is actually useful or not, or aim to enrich the information extracted from both the image and the question, even if this information becomes noisy. Aiming to effectively use the visual content of the image and the semantic clues of the question to reason and predict the correct answer, \citet{Nguyen_2022_CVPR} presented a new framework that reason over different levels of granularity.

First, features and predicates (keywords about objects, relations, or attributes) are extracted from the image and the question. Features are extracted from regions of interest in the image by an object detector inspired by the Faster R-CNN model. Visual predicates, on the other hand, are extracted by classifying attributes and relations based on RoIs converted into a word representation generated by the GloVe model. Each predicate follows one of three forms: single predicate, attribute-based, or relation-based. To extract features and learn the dependencies between all the words in the query, a representation of the query is generated from another GloVe model, followed by a recurrent GRU module.

Next, a coarse-to-fine reasoning (CFR) module is used to jointly learn the features and predicates. This module uses three steps: information filtering, multimodal learning, and semantic reasoning. Since the question and image features and predicates are extracted by pre-trained models, they may contain noise or incorrect information. Information filtering aims to remove unnecessary visual information from the image based on the predicates, and to help understand the importance of each RoI for each question.

The multimodal learning module learns the semantic mapping between the two modalities (visual and textual) and identifies which instances in the image are useful for answering the question. This is done at both coarse and fine-grained levels. While the coarse-grained level works out the interaction between the two extracted features, the fine-grained level learns the interaction between the filtered information obtained in the information filtering step. Finally, the semantic reasoning module combines the results of the multimodal learning step using trained adaptive weights to generate the response.

The authors' main observation is that features of the image and the question can be successively extracted at different levels of granularity, allowing stronger connections to be made by mapping them at each level. By visualizing the explicit contribution of regions of interest and predicates in both the input image and the question, the coarse-to-fine reasoning framework can not only increase the accuracy in the VQA task, but also provide a way to understand the prediction results.


\subsubsection{Textual Explanation}

A common way to explain the predicted answers is to refer to the attention maps, which indicate ``where to look''. However, this explanation is implicit and does not fully reveal what the model captures from the attended regions to answer the questions, which occurs in cases where the model attends to the right regions but generates incorrect answers. Due to this lack of transparency, several authors have proposed to explain the predicted answers by generating a textual explanation.

The textual explanation can be a hint to justify the answer, a complementary definition about the context of the question and answer, or even a detailed specification of abstract concepts~\cite{Li_2018_ECCV_2}. Such textual explanations are important as they make communication more accessible to the visually impaired, who are the potential users of VQA techniques, by providing information that allows the conversation to be extended, elaborated, and enriched with more relevant information.

\citet{Li_2018_ECCV_2} proposed a new VQA-E (VQA with Explanation) task, where models have to generate a textual explanation in addition to the answer to the question. For this, they built a new dataset with textual explanations for the answers, which were automatically derived from the VQAv2 dataset by intelligently exploiting the available captions. They also proposed a new multitasking learning architecture for the VQA-E task.

The authors quantitatively demonstrated that additional supervision of explanations not only produces explanatory text sentences to justify answers, but also improves model performance for all question types. They argue that this additional supervision can reduce language bias because, to generate a good explanation, the model must explore the entire image content, attending to important regions and interpreting them explicitly in the context of the questions.

\citet{li-etal-2018-tell} proposed to divide end-to-end training into two steps: explanation and reasoning. The first extracts attributes and generates descriptions as explanations for an image, while the second uses these explanations instead of the image to infer a response. First, two levels of explanations for an image are generated using attribute detectors and a pre-trained image caption model. The word-level (attribute) explanations indicate individual objects and attributes that the system learns from the image, while the sentence-level (caption) explanations represent the relationship between objects and attributes. The generated explanations and questions are then presented to a reasoning module, which generates an answer.

This method has two main advantages: first, the attributes and captions can reflect what the system extracts from the image, as well as what information is lost during the image comprehension step, and; second, decoupling explanation from reasoning allows locating which step of the VQA process -- image comprehension or response inference -- is responsible for the error if the generated response is incorrect.


\subsubsection{Multimodal Explanation}

An integral part of human communication, understanding, and learning is explaining decisions. Typically, we provide explanations by pointing to evidence or by speaking/writing. To build deep learning models that are capable of explaining decisions with similar fluency in visual and textual modalities, \citet{Park_2018_CVPR} proposed a multimodal approach to explanation, arguing that the two modalities are complementary.

Their new model, Pointing and Justification Explanation (PJ-X), includes an explanatory attention step that allows pointing to visual evidence and justifying a decision with text. First, given an image and a question, the model generates an answer. Next, an attention map (visual pointing) is generated from the answer, the question, and the image. This visual pointer is used to create attention features that guide the generation of the textual justification.

The authors show quantitatively that learning to point helps produce high-quality textual explanations, and conversely, using reference textual explanations to train helps produce better visual pointing. They also show cases where visual explanation is more useful than textual explanation and vice versa, supporting the idea that multimodal explanation models offer significant benefits over unimodal approaches. Furthermore, the model proves to be easier for performance diagnosis by humans than models trained only on descriptions and also models trained only to generate textual explanations.


While it is important to provide interpretability to better understand the models, it is essential to ensure that this interpretability is robust and consistent. \citet{Patro_2020_WACV} showed that VQA methods generate responses and explanations that may diverge for an image corrupted by noise-based perturbations. To address this problem, they proposed systems for joint generation of explanations and responses, trained with a new correlated collaborative network, which ensures that responses and explanations are coherent and correct. This helps to generate robust explanations that are tolerant to such perturbations, even if it has not been trained with them.

The collaborative correlated network learns a unified representation by training on the textual explanation and a representation of the answers in a joint adversarial mechanism. It consists of three modules: an encoder, a generator, and a correlation module. The encoder module encodes images and questions and combines them via an attention mechanism to obtain attention features. The generator uses these features to generate answers and explanations. The correlated module defends the model against perturbations in the input.

The authors found that the best approach is to jointly check that the answers and explanations are correct and coherent. Thus, the correlated module can be understood as a robust method for checking whether the answer and the explanations are consistent. It not only helps to generate better textual justifications, but also to better locate the visual explanation. Through a detailed empirical analysis of variants of their model and state-of-the-art solutions, the authors demonstrate that their model produces a textual and visual justification that is coherent and supports the decision, while consistently generating better answers and being tolerant to adversarial attacks.


\section{Dataset Generation Framework}

The present study sought to perform concept validation by creating a new benchmark comprising images and questions pertaining to one or more concepts. Each question is accompanied by three supplementary pieces of information that can be provided to the model alongside the question, thus helping it answer correctly. Additionally, the benchmark includes the object mask (when applicable), as well as a list of the underlying concepts related to the question. To facilitate the generation of this dataset, a framework was developed, comprising four distinct stages: initial object selection, background image selection, image generation, and question generation.The subsequent sections will provide a detailed account of the dataset generation process.


The framework enables the generation of datasets with varying characteristics, including the number of objects in the image, the type of objects, the type of questions, and the specific content of the questions. This approach allows the generation of datasets that vary in their level of difficulty, enabling the validation of specific concepts.

The decision to generate a synthetic dataset primarily using real objects and images was made due to the interesting mix of visual information that this configuration allows. For instance, the use of real objects in unconventional settings allows for the validation of a model's comprehension of intended concepts without the influence of environmental bias, as observed in the model presented by~\citet{2939672.2939778}, which identified wolves by the nature background. Concurrently, this approach enables more precise control over their positioning, which is a crucial aspect for the intended position concepts.

Models that have been trained to make associations between specific visual features and particular concepts may be affected by changes in the background image or the location of objects. For illustration, a model that has been taught to associate birds with blue backgrounds may be influenced by the addition or removal of blue elements in the background. Similarly, a model that has been trained to identify shoes may be influenced by changes in the position of shoes in the image, with shoes typically appearing in the lower region of the image.


It is noteworthy that the objects added to the image are not subjected to any form of enhancement to make the collage more natural, such as adjustments to lighting, styling, or edges. While these factors are worthy of further investigation, the experiments conducted demonstrate that models trained on natural images occasionally fail to identify objects and concepts under these circumstances, a phenomenon not observed in humans. For this reason, we have chosen to name our framework and dataset as \scrapbook.

Following a series of experiments, we opted to also generate a dataset comprising colored geometric shapes. This will enable us to validate the model's comprehension of the concepts in a more controlled setting, where the objects lack any visual information that might influence the model's response. It is imperative to acknowledge that while success on this dataset is \textit{not} a definitive indicator of a VQA model's robustness, we believe that it is a crucial component.

\subsection{Selecting the Objects}


We start by defining an $S$ amount of objects sets to be used, being each set $s$ composed of $N$ objects. The objects are selected from a list of known classes, which can be either geometric shapes: \textit{circle}, \textit{square}, \textit{triangle}, \textit{pentagon}, \textit{hexagon}, or \textit{heptagon}; or real objects, from a list of 59 possibilities extracted from the COCO 2017 dataset~\cite{mscoco}. When using geometric shapes, we also select the colors of the objects, which can be: \textit{black}, \textit{blue}, \textit{green}, \textit{orange}, \textit{red}, \textit{white}, or \textit{yellow}; and the size of the objects, which start from a bounding box size of 70$\times$70 with increments of 40 to each dimension for each new object considered. When using real objects, they are not modified in any way, maintaining their original colors and sizes.

For the selection of the individual characteristics for the composition, like objects classes and their respective colors and sizes in the case of geometric shapes, we have 3 options: \textit{same}, \textit{unique}, and \textit{random}. When using the \textit{same} option, all selections have the same value; the \textit{unique} option ensures that each selection has a different value; whereas the \textit{random} option allows the values to be chosen randomly, with the possibility of repetition.

Note that it is possible to use different options for the objects classes, colors, and sizes. For example, it can be randomly selected a set of geometric shapes, while all having the same color but unique sizes. These options were added to allow the creation of questions that might have different levels of difficulty, depending on the characteristics of the objects in the image. For example, all objects having the same color might make it easier for the model to answer questions about the colors of the objects, while having different colors might make it harder.

Currently, there are two main possibilities when selecting the objects for the sets: \textit{deterministic} and \textit{random}. In the \textit{deterministic} case, the objects, their colors, and sizes are \textit{uniquely} selected by a sliding window, which moves 1 position for every new set. For example, considering the possible shapes referred before, the first set of 3 objects would include a \textit{circle}, a \textit{square}, and a \textit{triangle}, while the second set would be composed by a \textit{square}, a \textit{triangle}, and a \textit{pentagon}. This ensures that every iteration of the dataset generation will have one same subset of objects, colors, and sizes, while still adding new elements. The same applies to the real objects, with the difference being that only the objects classes are selected. In the \textit{random} case, the objects, colors, and sizes are \textit{randomly} selected for each set, with the possibility of repetition if allowed.

When using geometric shapes, the objects and their respective masks are created using the Pillow python library~\cite{andrew_murray_2023_7791017}. For the version with real objects, both the objects and their masks are extracted from the COCO 2017~\cite{mscoco} dataset, more specifically the validation set, excluding objects that are small (with any of its dimensions smaller than 48 pixels) or that are at the edge of the image (possibly incomplete). We chose the COCO 2017 dataset as our object source because most of the models used in the VQA literature are trained on it, so it would be expected that they can identify these objects. After this initial filter, we manually selected the objects that are visually identifiable in an isolated manner, not excluding incomplete objects, resulting in 59 objects classes. Some examples can be visualized in Figure~\ref{fig:mscoco_objects} and the complete list of objects classes can be found in Appendix~\ref{app:mscoco_objects}.

\begin{figure}[!htb]
    \captionsetup[subfigure]{labelformat=empty}
    \centering
    \subfloat[Backpack]{\includegraphics[width=0.2\textwidth]{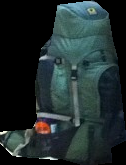}}
    \hspace*{0.2cm}
    \subfloat[Oven]{\includegraphics[width=0.2\textwidth]{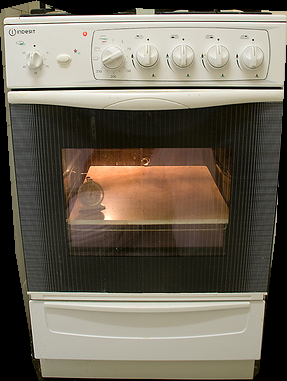}} \\
    \subfloat[Bench]{\includegraphics[width=0.2\textwidth]{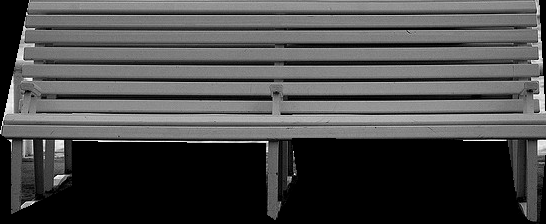}}
    \hspace*{0.2cm}
    \subfloat[Remote]{\includegraphics[width=0.2\textwidth]{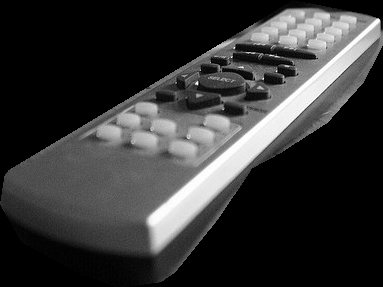}}
    \caption{Examples of objects extracted from the MSCOCO validation set.}
    \label{fig:mscoco_objects}
\end{figure}



\subsection{Selecting the Background Images}

\begin{figure}[!htb]
    \captionsetup[subfigure]{labelformat=empty}
    \centering
    \includegraphics[height=2.1cm,width=3.3cm]{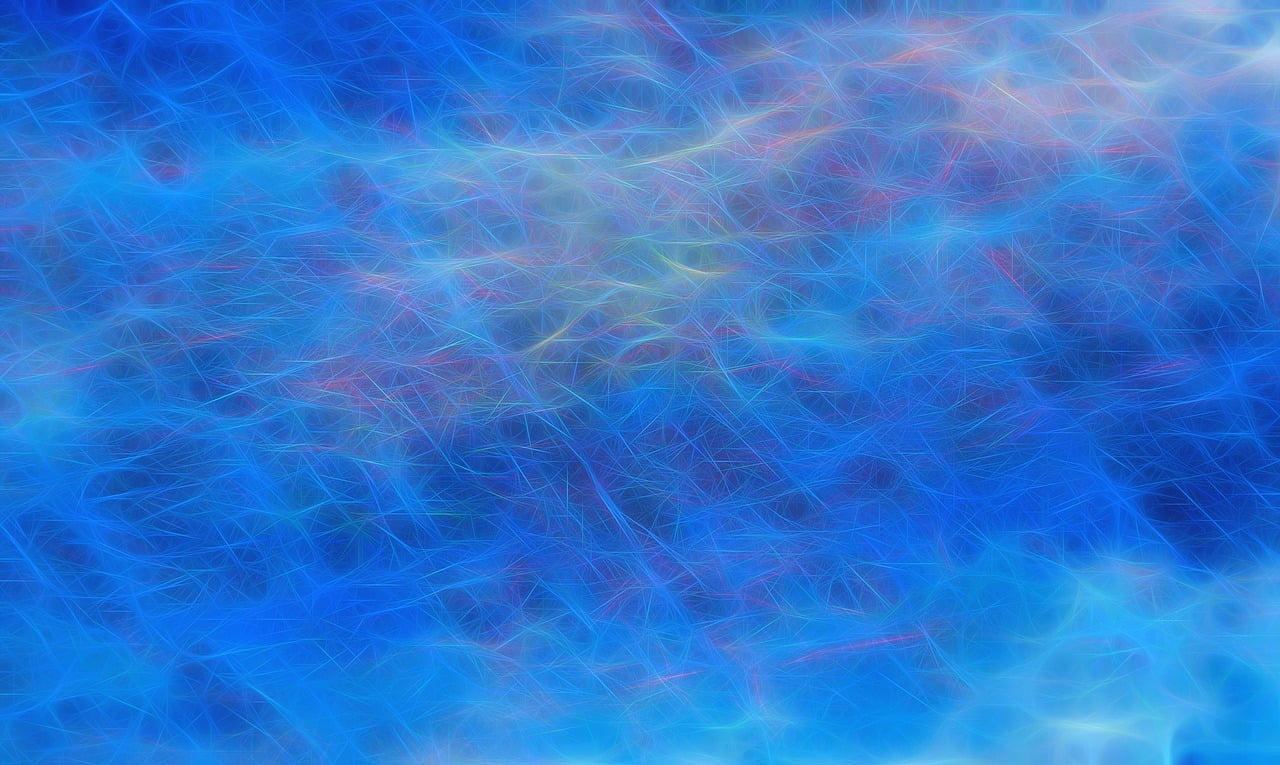}
    \hspace*{0.2cm}
    \includegraphics[height=2.1cm,width=3.3cm]{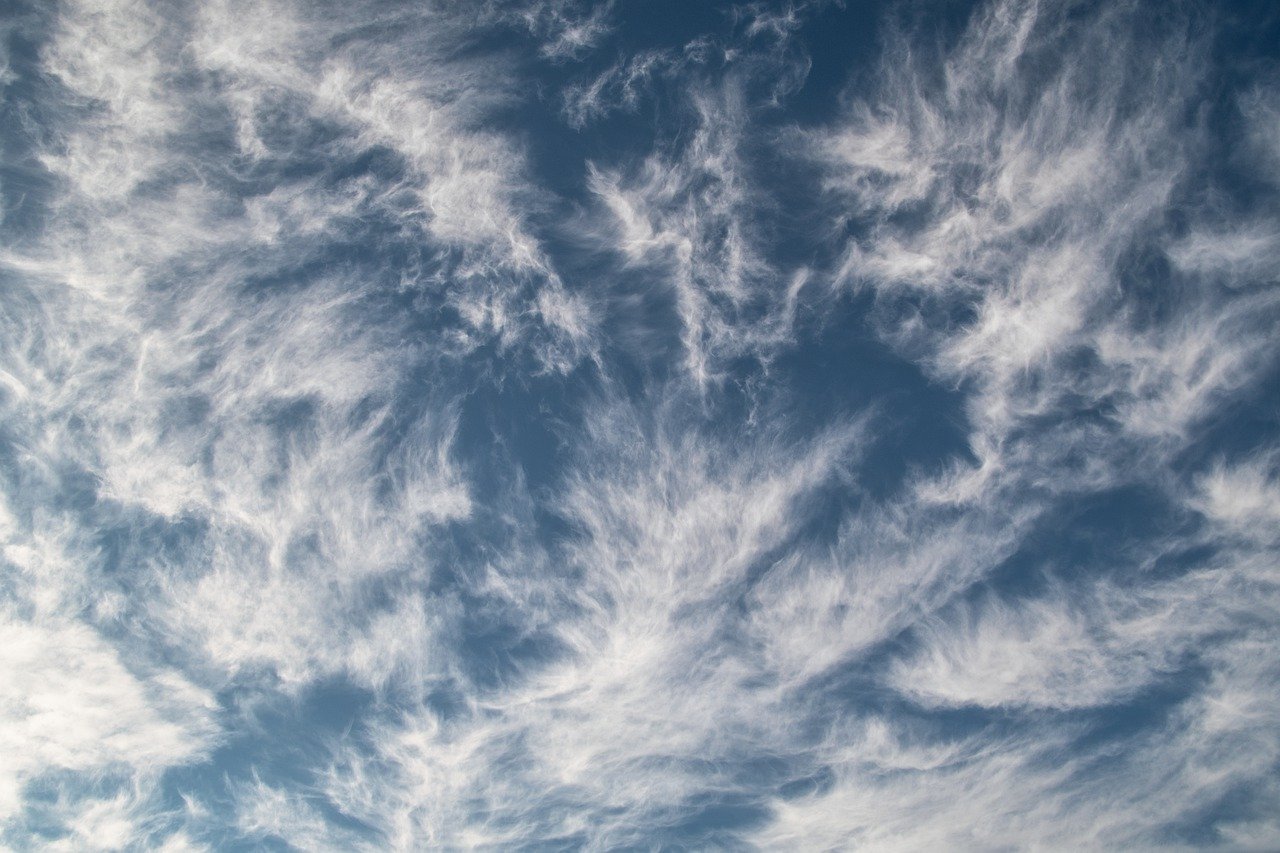} \\[0.2cm]
    \includegraphics[height=2.1cm,width=3.3cm]{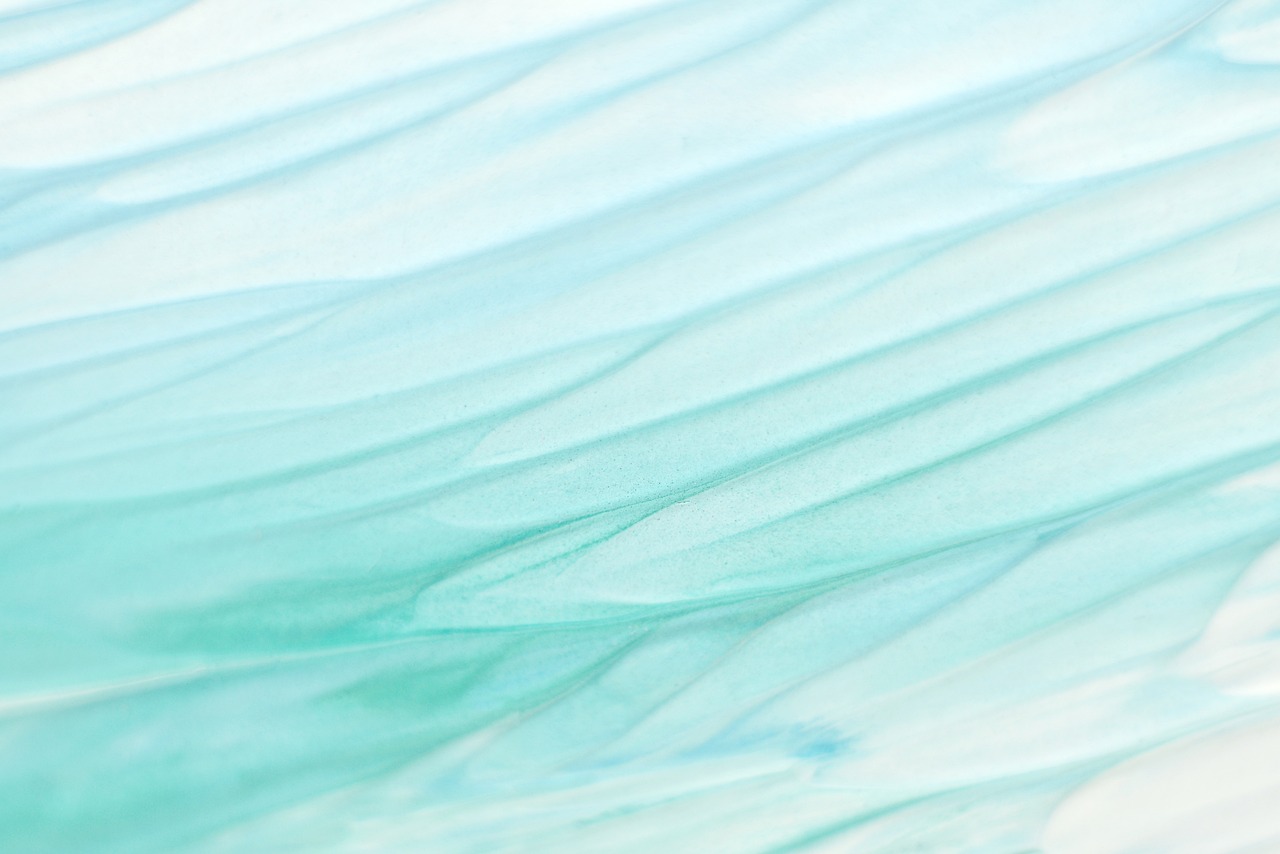}
    \hspace*{0.2cm}
    \includegraphics[height=2.1cm,width=3.3cm]{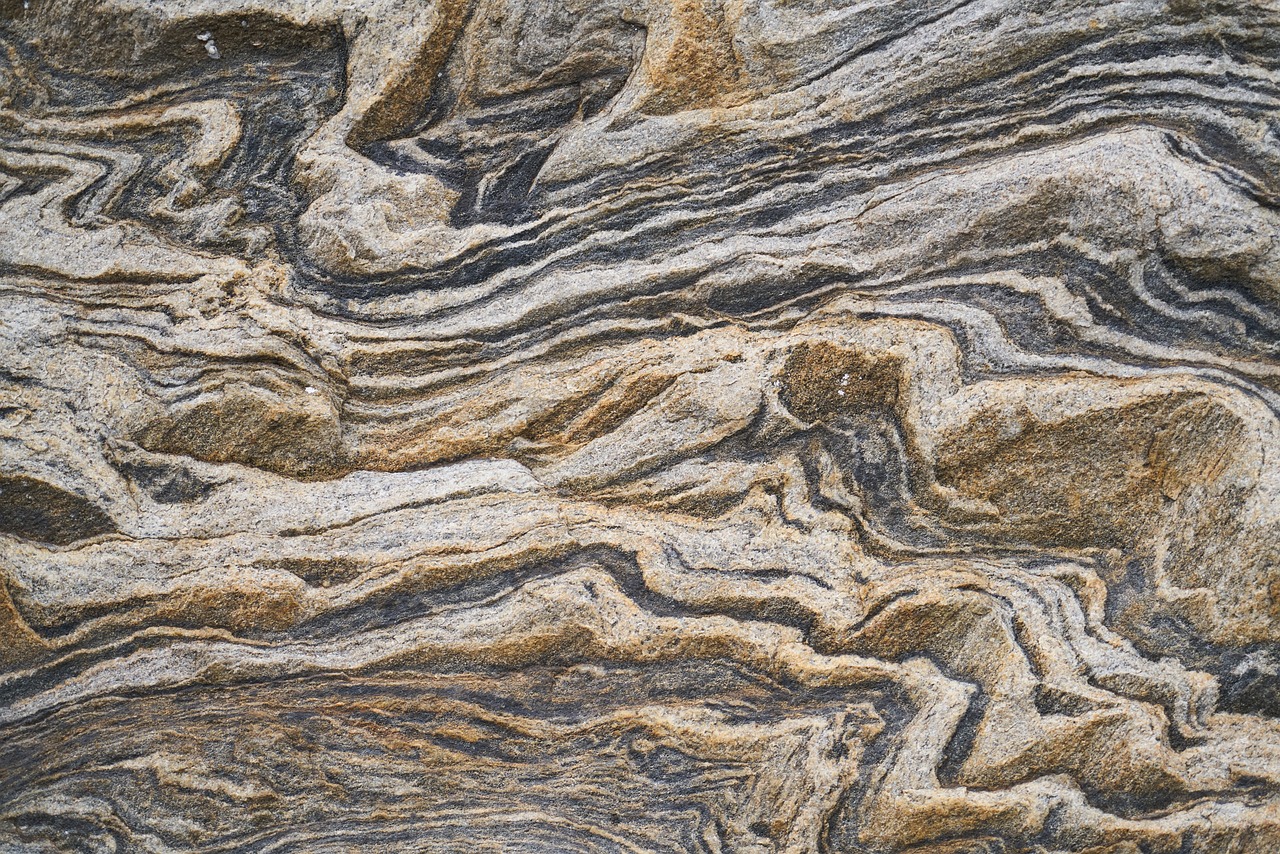}
    \caption{Examples of background images selected from Pixabay.}
    \label{fig:scrapbook_backgrounds}
\end{figure}

Background images are either solid colors not used in any of the objects from the current set (when using colored shapes), which are used to fill an image of dimensions 1280$\times$768 pixels, or permissively licensed images taken from the Internet that have a minimum size of 500$\times$500 pixels. These downloaded images were chosen to be as diverse as possible, so that the objects can be positioned in different ways, and without much identifiable visual information, to avoid the bias of the environment in which the objects can be found. This is done to avoid open-ended models to answer questions considering other objects in the image, like ``to the left of the tree'', for example. A few examples are shown in Figure~\ref{fig:scrapbook_backgrounds}.

\subsection{Generating the Images}


The images consist of one to at most $N$ objects from a set $s$, positioned in background images. To create the images, first $S$ sets of $N$ objects are selected. For each set $s$, $B$ background images are also chosen from the available ones. Then, for every set $s$ of objects, considering the background image $b$, $A$ arrangements of these objects are selected. Each arrangement $a$ is formed by a pair of objects from the set $s$, the first object being called \mainobj\ and the second \refobj, and the remaining objects from the set. The arrangements are created so that the pair \mainobj\ and \refobj\ is always different, so the maximum number of possible arrangements is $A \leq N!/(N! - 2!)$.

From an arrangement $a$, the first $x$ objects will be added to an image, with $x \leq N$. The objects from $a$ that are not in the current image are used to create questions about the absence of objects, as well as to create questions about objects with incorrect characteristics.

\begin{figure}[!htb]
    \centering
    \includegraphics[width=0.8\linewidth]{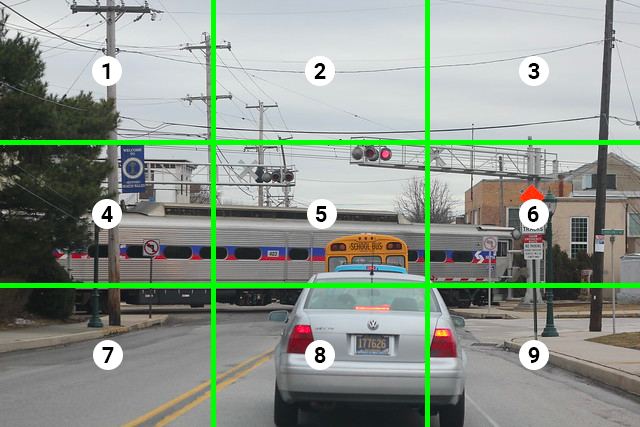}
    \caption{Possible absolute positions in an image. 1) Top left; 2) Top center; 3) Top right; 4) Center left; 5) Center; 6) Center right; 7) Bottom left; 8) Bottom center; 9) Bottom right. Image taken from COCO dataset~\cite{mscoco}.}
    \label{fig:abs_pos}
\end{figure}
\begin{figure}[!htb]
    \centering
    \includegraphics[width=0.8\linewidth]{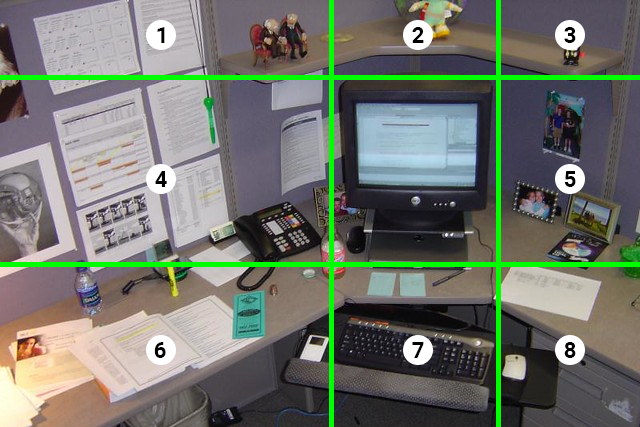}
    \caption{Possible relative positions from an object in an image, in this case the monitor. 1) to the upper left; 2) above; 3) to the upper right; 4) to the left; 5) to the right; 6) to the lower left; 7) below; 8) to the lower right. Image taken from COCO dataset~\cite{mscoco}.}
    \label{fig:rel_pos}
\end{figure}

In a next step, $P \leq 81$ pairs of regions of the same image (called absolute positions) are chosen, consisting of two options from those shown in Figure~\ref{fig:abs_pos}. For each pair of absolute positions ($abs\_pos$), all 8 possible relative positions ($rel\_pos$) among the options shown in Figure~\ref{fig:rel_pos} are considered. If it is possible to add the \mainobj\ in the first absolute position and the \refobj\ in the second absolute position, so that the \mainobj\ is $rel\_pos$ of the \refobj, images and questions will be created for these settings.

\begin{figure}[!htb]
    \centering
    \subfloat[]{\includegraphics[width=0.8\linewidth]{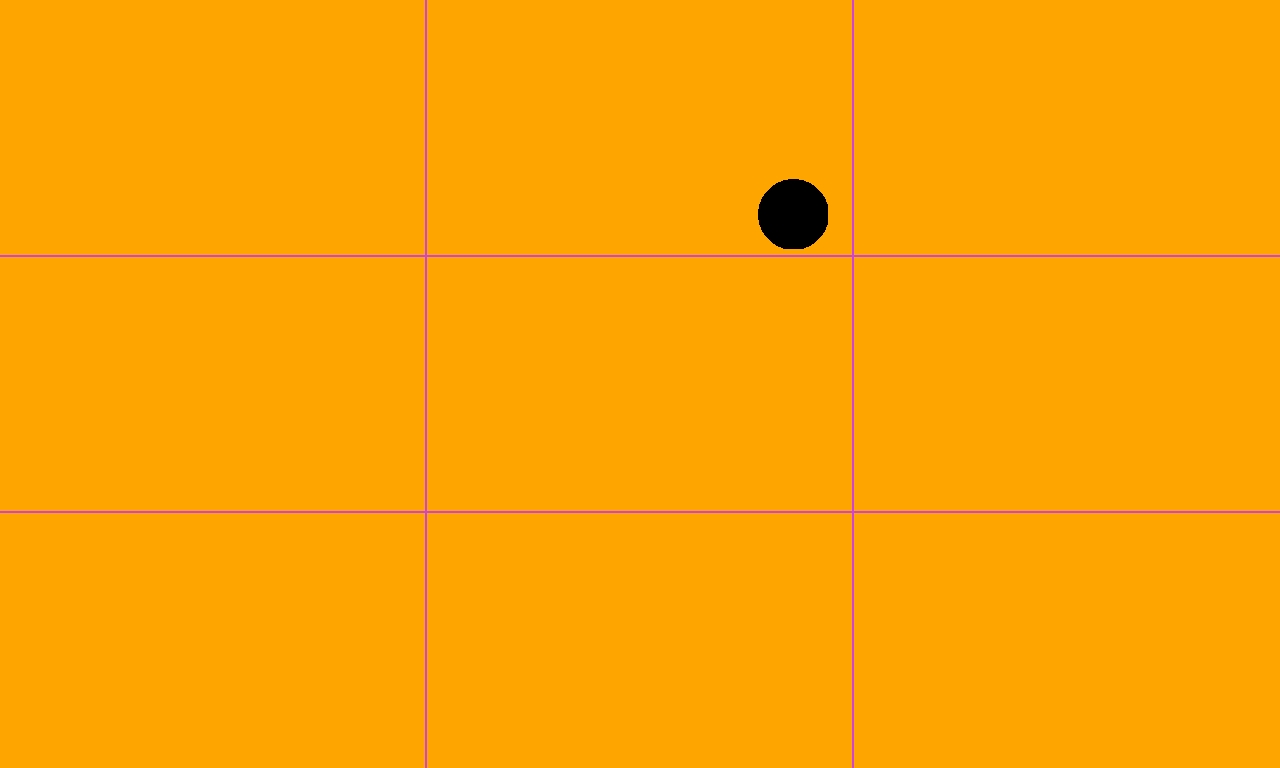}} \\
    \subfloat[]{\includegraphics[width=0.8\linewidth]{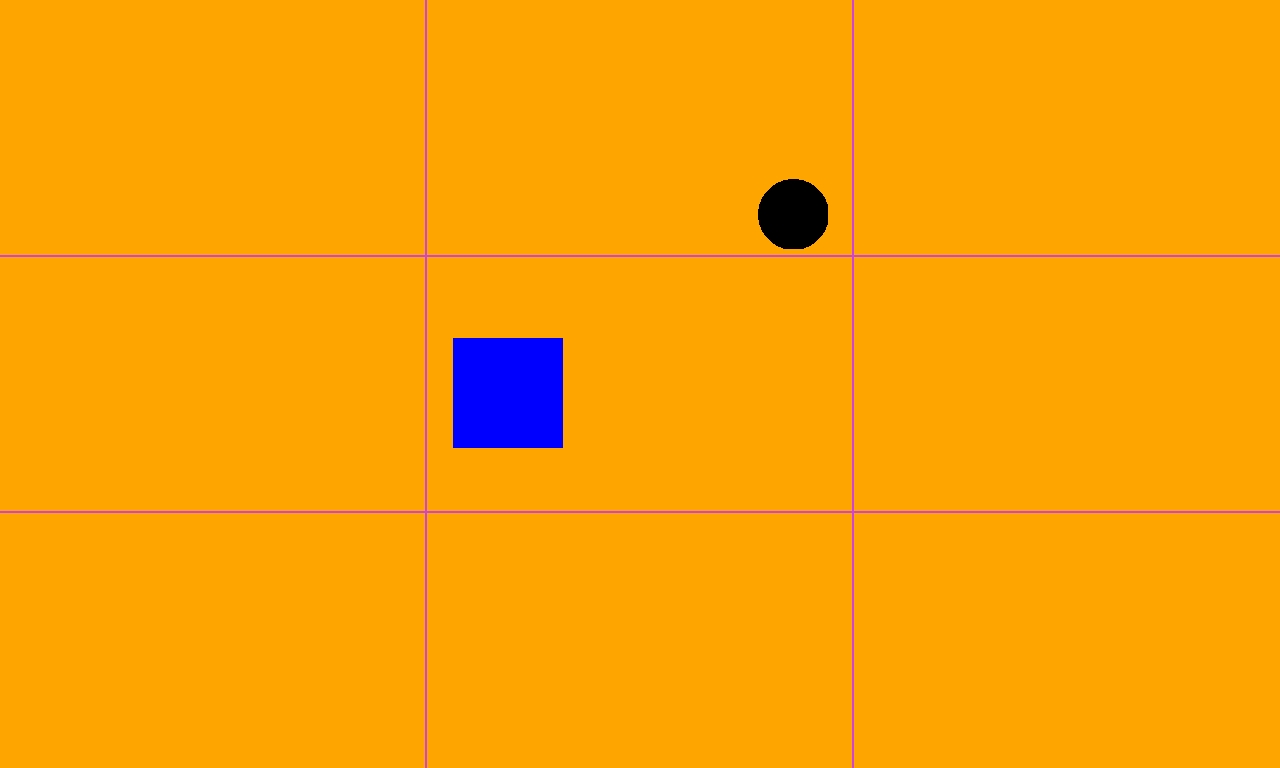}} \\
    \subfloat[]{\includegraphics[width=0.8\linewidth]{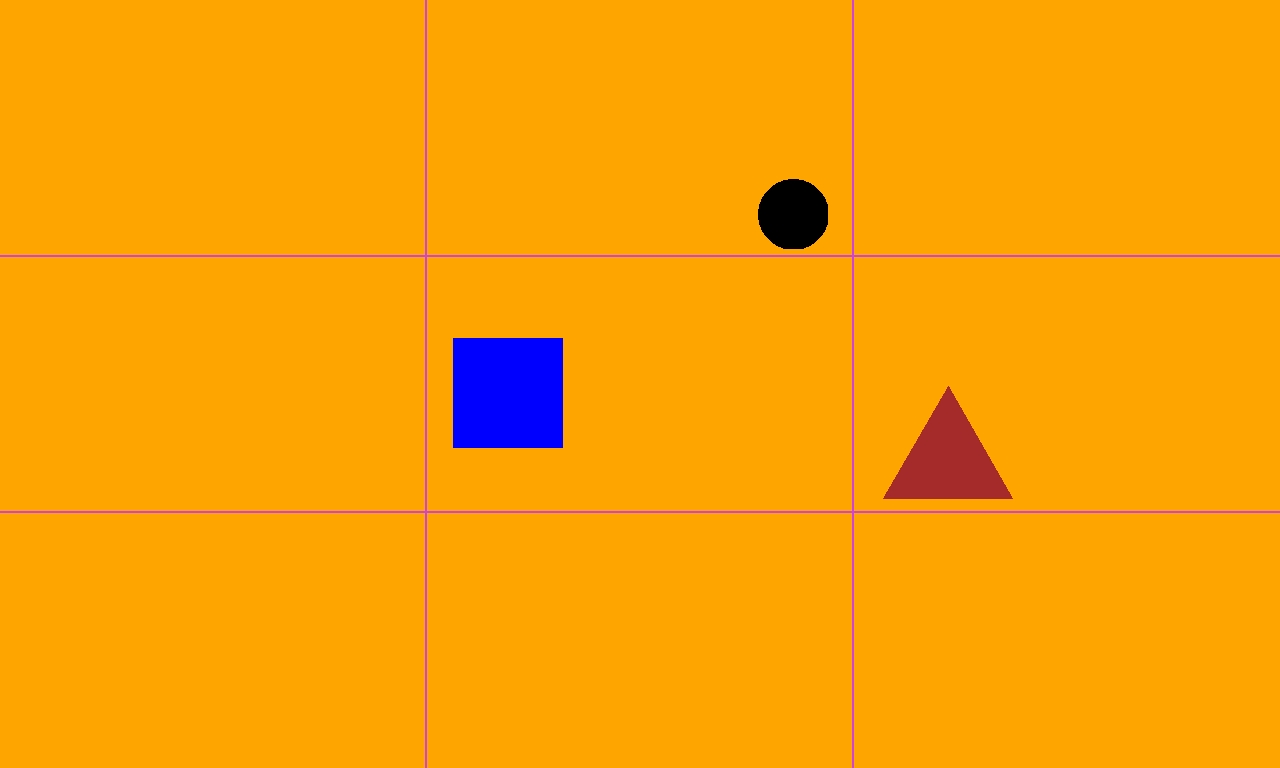}}
    \caption{Example of images generated with the \textit{unique} option for the objects classes, colors, and sizes. (a) \mainobj\ (black circle) added in the top center of the image; (b) \refobj\ (blue square) added to the center of the image, respecting the ``top right'' relative position restriction; (c) third object (brown triangle) added to the right of the image, still ensuring that only the \mainobj\ is to the ``top right'' of the \refobj. Grid lines were added to help visualize the absolute positions.}
    \label{fig:example_image}
\end{figure}

During image creation, the \mainobj\ is arbitrarily positioned within the first absolute position, such as that at least 3 quarters of it are completely inside that region and the whole object is contained in the background, generating one image. Then, the \refobj\ is randomly added within the second absolute position, respecting the same constraints and the relative position constraint $rel\_pos$, generating a second image. From this point on, the remaining objects up to $x$ will be added incrementally into the image, but ensuring that only the \mainobj\ is in $rel\_pos$ of the \refobj, creating new images with each new object added. The new object will only be added if a valid position is found for it, so that it is completely contained in the image and there are no overlaps between objects. An example with 3 objects is shown in Figure~\ref{fig:example_image}.

\subsection{Generating the Questions}


For each image we also create questions about it. $Q$ questions were generated for each type, where the types are about the objects, their positions, and their characteristics (if the objects are colored shapes). The questions are generated from predefined templates, with the values of the variables being replaced according to the characteristics of the objects in question, their absolute or relative positions if necessary, and by values contained in an expression dictionary. In this way, we can guarantee a wide linguistic variety from a few templates.


For example, starting from the template ``are you aware of \{art\} \{obj\}\{where\_affirmation\} \{abs\_pos\}?'', considering the object \textit{dog} and the absolute position \textit{left}, it is possible to create variations such as:

\begin{myitemize}
    \item are you aware of a dog on the left?
    \item are you aware of a dog on the left part?
    \item are you aware of a dog on the left of the scene?
    \item are you aware of a dog somewhere on the left in the image?
    \item are you aware of a dog somewhere on the left half?
    \item are you aware of a dog somewhere on the left half of the picture?
    \item are you aware of a dog on the left half in the pic?
\end{myitemize}

We first create questions about the objects depicted in the image, followed by the subsequent generation of questions concerning the objects from arrangement $a$ that are not present in the image. In instances where there are two or more objects in the image, questions are also generated to address the relative position of the objects. The generated questions are classified into four distinct categories: presence, counting, confirmation, and recognition.

Presence questions concern the presence or absence of an object, as exemplified by the query, ``Is there a dog in the picture?''. Counting questions refer to the number of objects in the picture, for instance, ``How many dogs are there?''. Confirmation questions are related to confirming the characteristics of an object or the presence of an object in a particular position, for example, ``Is the square yellow?'' or ``Is there a dog on the left?''. Recognition questions are related to recognizing an object or its characteristics, or identifying the position based on some characteristics. Examples include ``What is the color of the triangle?'' or ``Where is the blue circle?''.

Both presence and confirmation questions have three possible answers: \textit{yes}, \textit{no}, or \textit{$<$unk$>$}. Count questions yield a numeric answer, which corresponds to the number of objects in the image, or else the answer is designated as \textit{$<$unk$>$}. Recognition questions entail a textual answer, which is the property of the object in question, or the position of the object based on some property, or else the answer is \textit{$<$unk$>$}. The use of the \textit{$<$unk$>$} indicates an instance in which the question is deemed to be incoherent or invalid. For instance, the question ``Is the circle blue?'' is incoherent when there is no circle visible in the given image. Similarly, the question ``Are there any dogs to the left of the cat?'' is deemed invalid when no cat is observed.


We divided the questions types further into groups and subgroups, with the exception of recognition questions, given their assumption of the existence of the object in question. Recognition questions only have groups about the lack of positional information (e.g., ``What is the color of the circle?''), that have absolute position (e.g., ``What is the color of the circle on the top left?''), or have relative position (e.g., ``What is the shape of the red object to the left of the square?'') information, including when the answer is the position itself (e.g., ``Where is the red circle?''). Since recognition only have one subgroup, its questions files always end with a suffix of \_1, meaning that the answer is expected to be the correct value.

The remaining question types are categorized into the same three distinct groups, based on the presence of position information: no position information, absolute position information, or relative position information. The group without position information is further subdivided into two subgroups. The first subgroup corresponds to cases where the object focused on is present in the image (question files ending with \_1). In this case, the expected answer is either \textit{yes} or the correct number in the case of counting questions. The second subgroup corresponds to cases where the object is not present in the image (question files ending with \_2). In this case, the expected answer is either \textit{no} or zero for counting.

The group with absolute position is divided into three subgroups: one where the object focused on is in the image and in the correct position (question files ending with \_1), another where the object focused on is in the image but in another position (ends with \_2), and the last one where the object focused on is not in the image (suffixed with \_3). The expected response for the first subgroup is \textit{yes} or the correct number, while for the other two, it is \textit{no} or zero. or zero.

The remaining group (with relative position) is divided into four subgroups: one where the \mainobj\ and \refobj\ are in the image and in the correct relative position (suffix \_1); another where the \mainobj\ and \refobj\ are in the image but in another relative position (\_2); the third fourth involves the absence of the \mainobj\ in the image, while the \refobj\ is present (case \_3); and the fourth subgroup is characterized by the absence of the \refobj, regardless of whether the \mainobj\ is (case \_4). For the first subgroup, the required answer is \textit{yes} or the correct amount, while for both subgroups, \_2 and \_3, it is \textit{no} or zero. For the last subgroup, the expected answer is \textit{$<$unk$>$} since it is an incoherent question.

Each question has four different forms: the original question, with an added condition, a question with a direction, and enumerated answers. The original question is the question as it was generated, with no additional information. The question with an added condition is the original question with an addendum, such as ``answer with the object shape, when applicable''. The version with a direction adds a slight constraint about how to answer it, like ``based on the scene, answer this question with a short sentence''. When using enumerated answers, all the possibilities are listed, such as ``choose one of the alternatives: \textit{yes}, \textit{no}, or \textit{not applicable}''.

It is noteworthy that while a question might require the comprehension of multiple concepts to be answered correctly, the response is constrained to a single concept. To illustrate, the question ``Is there a dog on the left of the cat?'' assesses the model's understanding of the concept of a dog, the concept of a cat, the containment of both within the figure, and the relative position ``on the left'' with respect to the cat. However, the answer is exclusively concerned with the presence of the dog, as it is the main object in this instance. To mitigate challenges in validating the model's comprehension of various concepts programmatically, we avoided questions that demand the application of multiple concepts in the response, such as ``what is the color and shape of the object on the left?''.

Given the capacity of the templates to generate a substantial number of paraphrased questions, a new parameter, designated as $n\_questions\_per\_type$, has been introduced to regulate the number of questions generated for each type. This parameter functions as a limit on the number of questions generated for a given set of question parameters. For instance, if $n\_questions\_per\_type$ is set to five, only five questions will be generated about the presence of an object given a specific position. Note that this phenomenon is specific to a given set of parameters; when the position is altered, five questions will be generated for this new set. Consequently, an imbalance in the number of questions generated for each type is inevitable.

The recognition type, which possesses only one subgroup, invariably generates fewer questions. Typically, there is an increase in the number of questions concerning objects absent from the image or with erroneous characteristics or positions, depending on the parameters used. To illustrate this, consider the scenario where the parameter $N$, denoting the number of objects per set, is set to five and the parameter $x$, representing the number of objects to be added to an image, is set to three. In this case, when the initial image is created, the first object is added, while the remaining objects (four in this case) are utilized to formulate questions concerning objects that are not present in the image. Additionally, questions will be created about the placement of the initial object in both correct and incorrect positions, with the latter corresponding to the locations where the other objects from the set will be added (if not the same as the correct one). This process will be repeated for characteristics such as color and shape.

In summary, the following divisions were established:
\begin{myitemize}
    \item 4 types: presence, counting, confirmation, and recognition;
    \item 3 groups: no position information, absolute position information, and relative position information;
    \item 1 subgroup for recognition questions (\_1), and 2 or more for the remaining types, depending on the presence of the objects in the image and their positions;
    \item 4 forms: original question, question with a condition, question with a direction, and enumerated answers.
\end{myitemize}

\section{Validating Generated Datasets}

Since the \scrapbook\ framework can generate massive datasets with a large number of questions, it is important to validate the results to ensure that the generated questions are indeed testing the concepts of interest. In this section, we present a methodology to validate the results of the framework. We first describe the methodology used to validate the results, and then we present the results of the validation process.



\subsection{Methodology}
\label{sec:methodology}

The primary objective of the \scrapbook\ framework is to generate a dataset that can be utilized to validate the concepts of absolute and relative positions. However, both of these concepts require reference points to be validated. Therefore, we have established what we refer to as ``levels'' of validation. Within each level, multiple concepts can be addressed. The lower the level, the fewer concepts are required to answer the questions. The levels are defined as follows:

Level 1 -- This level requires only a single concept to answer each question. Specifically, presence and counting questions devoid of positional information are the only types of questions that meet this criterion. Illustrative examples include ``Is there a red object in the image?'' or ``How many red objects are in the image?''. These questions need only the concept of color, red, to be correctly answered.

Level 2 -- It involves questions that depend on two concepts, such as presence and counting questions with two concepts (where one can be absolute position) and confirmation and recognition questions \textbf{without} position information, since these require at least two concepts: one for the reference and one for the answer. Examples of such questions include ``What is the color of the triangle?'', ``Can you see a red object on the left?'', and ``Is it possible to see a green square?''.

Level 3 -- It comprises questions that include presence and counting questions with three concepts, one of which \textbf{is} absolute or relative position. Other possibilities are confirmation and recognition questions with three concepts, one of which \textbf{can be} absolute or relative position. Examples of questions that require three concepts include ``How many red circles are on the left side of the image?'', ``Is there a green object to the right side of the red object?'', and ``What is the color of the pentagon that is on the lower right?''.

Level 4 -- This level is the most complex, as it requires 4 concepts to be answered. This includes presence and counting questions with 3 concepts and relative position, also confirmation and recognition with 3 concepts and an absolute or relative position. Examples of questions that require 4 concepts are ``how many red triangles are to the left side of the square?'', ``is the circle on the right side of the blue object green?'', and ``what is the shape of the blue object that is to the right side of the hexagon?''.

Level 4 -- It is the most complex level, as it requires four concepts to be answered. This includes presence and counting questions that need three concepts and relative position. Additionally, confirmation and recognition questions, which demand three concepts and an absolute or relative position, are included. Examples in this level include the following: ``How many red triangles are to the left side of the square?'', s and relative position, also confirmation and recognition with 3 concepts and an absolute or relative position. Examples of questions that require 4 concepts are ``how many red triangles are to the left side of the square?'', ``Is the circle on the right side of the blue object green?'', and ``What is the shape of the blue object that is to the right side of the hexagon?''.

In order to consider a concept valid for that level in the dataset, we need to have questions considering the presence and the absence of it in the picture. Starting on the first level, we would need presence questions where the answer is ``yes'' and ``no'' for the color red, also counting questions with answer 0 and with answers greater than 0. Only if these requirements are met, we can consider the concept of color red as valid for that level, and it can be used as a reference for other questions. This way we can ensure that the concepts can be incrementally used to answer more complex questions, and that there will always be ``easier'' questions to validate the concepts before ``harder'' questions can be asked.

In order to consider a concept valid for a given level in the dataset, the presence and absence of the concept in the picture must be considered. At the initial level, the presence of the color ``red'', for example, should be assessed through questions that include ``yes'' and ``no'' responses, also counting questions with answer 0 and with answers greater than 0. The validation of the concept of color ``red'' at this level is dependent on the fulfillment of these criteria. Once validated, the concept can serve as a reference for subsequent levels. This approach ensures that the concepts can be progressively employed to address increasingly complex questions, with easier ones serving to validate the concepts prior to progressing to more challenging questions.


\subsection{\scrapbook\ Datasets}

\begin{figure}[!htb]
    \centering
    \subfloat[]{\includegraphics[width=0.8\linewidth]{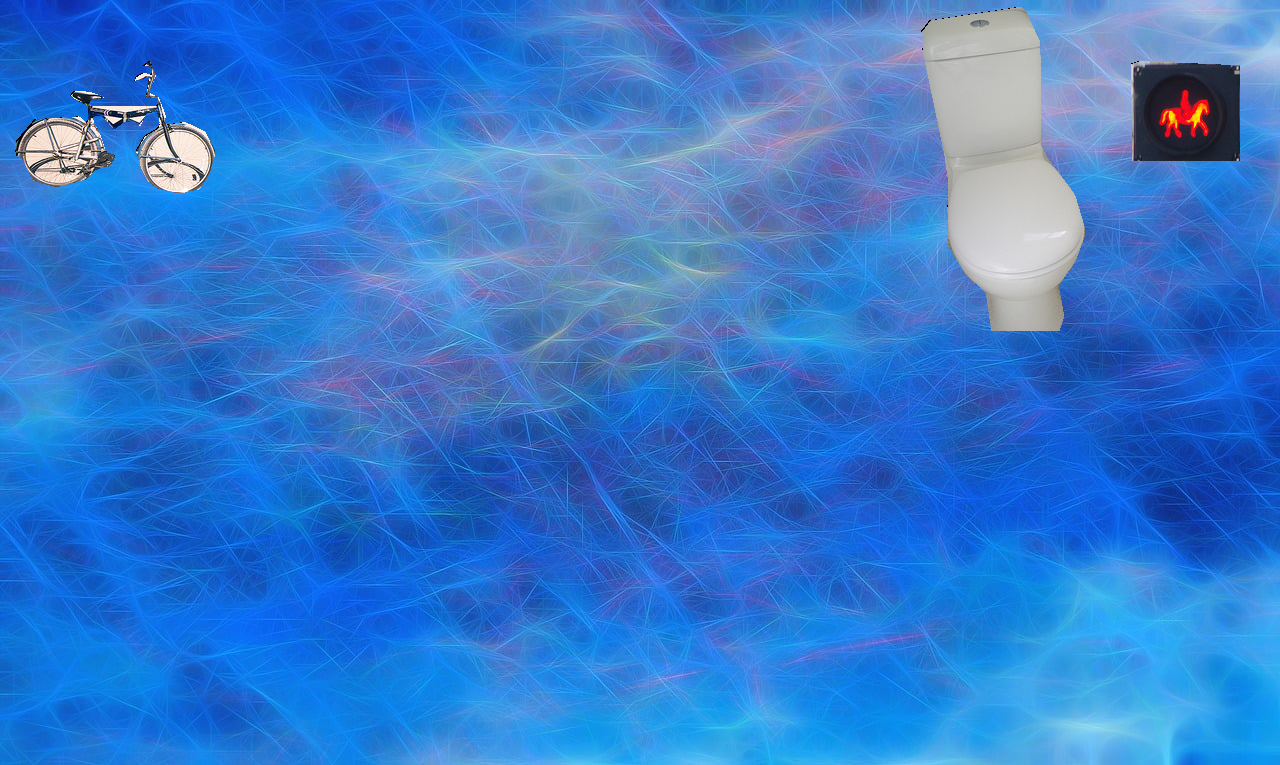}} \\
    \subfloat[]{\includegraphics[width=0.8\linewidth]{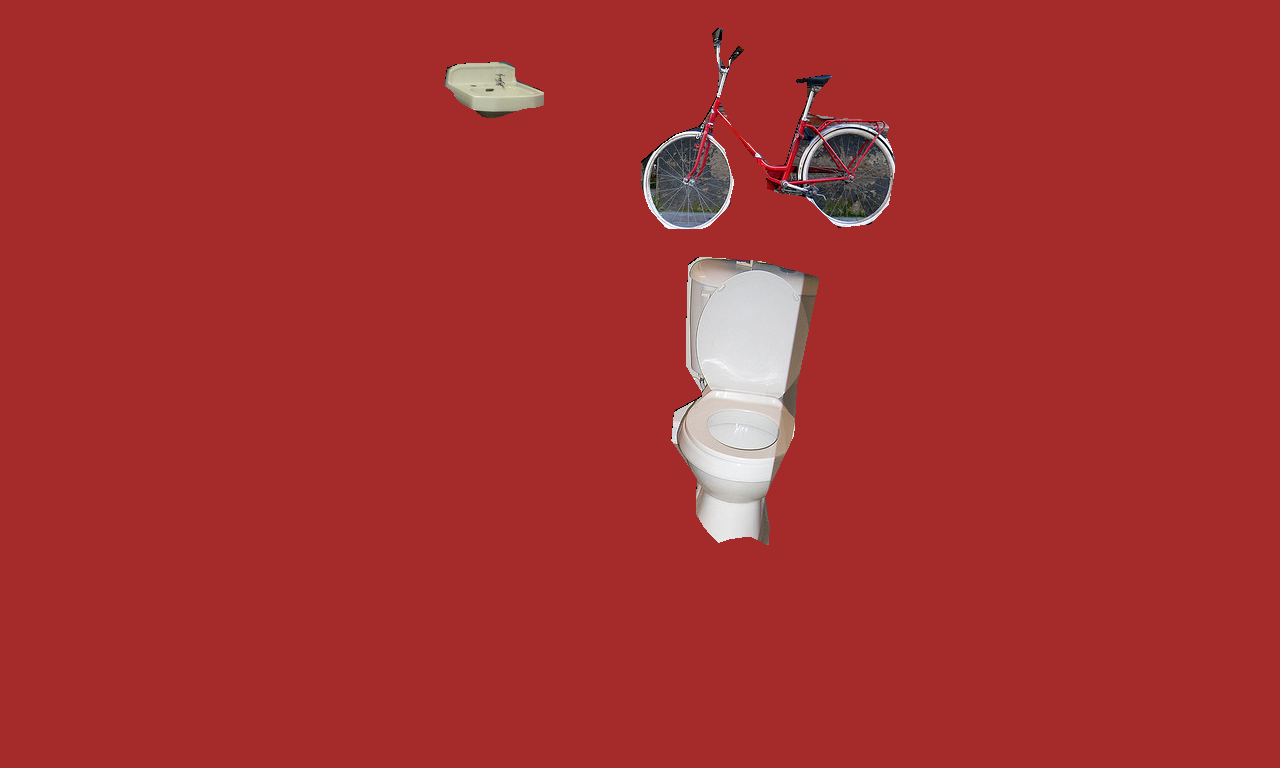}} \\
    \subfloat[]{\includegraphics[width=0.8\linewidth]{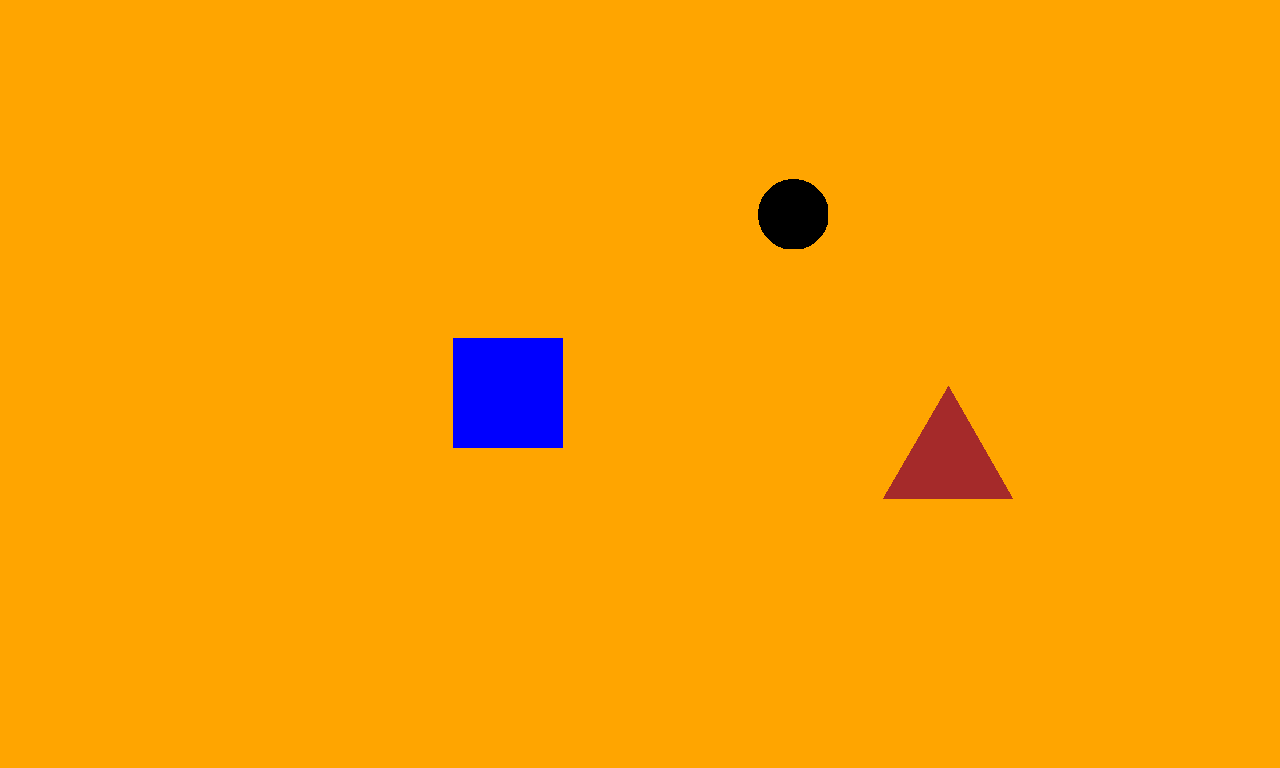}}
    \caption{Example of images from each version of the \scrapbook\ Dataset. (a) \cocoscrapbook; (b) \cocoplainbgscrapbook; (c) \uniqueallscrapbook.}
    \label{fig:datasets_samples}
\end{figure}

Three distinct versions of the \scrapbook\ dataset have been produced. The first version, entitled \cocoscrapbook, was created through the integration of objects from the COCO dataset and publicly available images sourced from the internet. The second version, \cocoplainbgscrapbook, was developed by utilizing the same objects as \cocoscrapbook, but utilizing plain colors as the background. The third version, designated as \uniqueallscrapbook, involves the incorporation of colored geometric shapes into a plain background. Figure~\ref{fig:datasets_samples} presents images from each version, while the set of parameters employed for generating all versions of the dataset is detailed in Table~\ref{tab:scrapbook_params}.

\begin{table*}[!htb]
    \setlength{\tabcolsep}{5mm}
    \centering
    \caption{Parameters used for the creation of the \scrapbook\ datasets.}
    \label{tab:scrapbook_params}
    \small
    \begin{tabular}{lc}
        \toprule
        \textbf{Parameter} & \textbf{Value} \\
        \midrule
        Object selection & deterministic \\ 
        Selection of the individual characteristics & unique \\ 
        Objects sets ($S$) & 3 \\ 
        Objects per set ($N$) & 5 \\ 
        Background images ($B$) & 3 \\ 
        Arrangements of objects ($A$) & 2 \\ 
        Objects to be added to an image ($x$) & 3 \\ 
        Pairs of regions ($P$) & 3 \\ 
        Questions per parameter set ($n\_questions\_per\_type$) & 3 \\ 
        \bottomrule
    \end{tabular}
\end{table*}


\subsubsection{\cocoscrapbook}

\begin{figure}[!htb]
    \centering
    \includegraphics[width=0.8\linewidth]{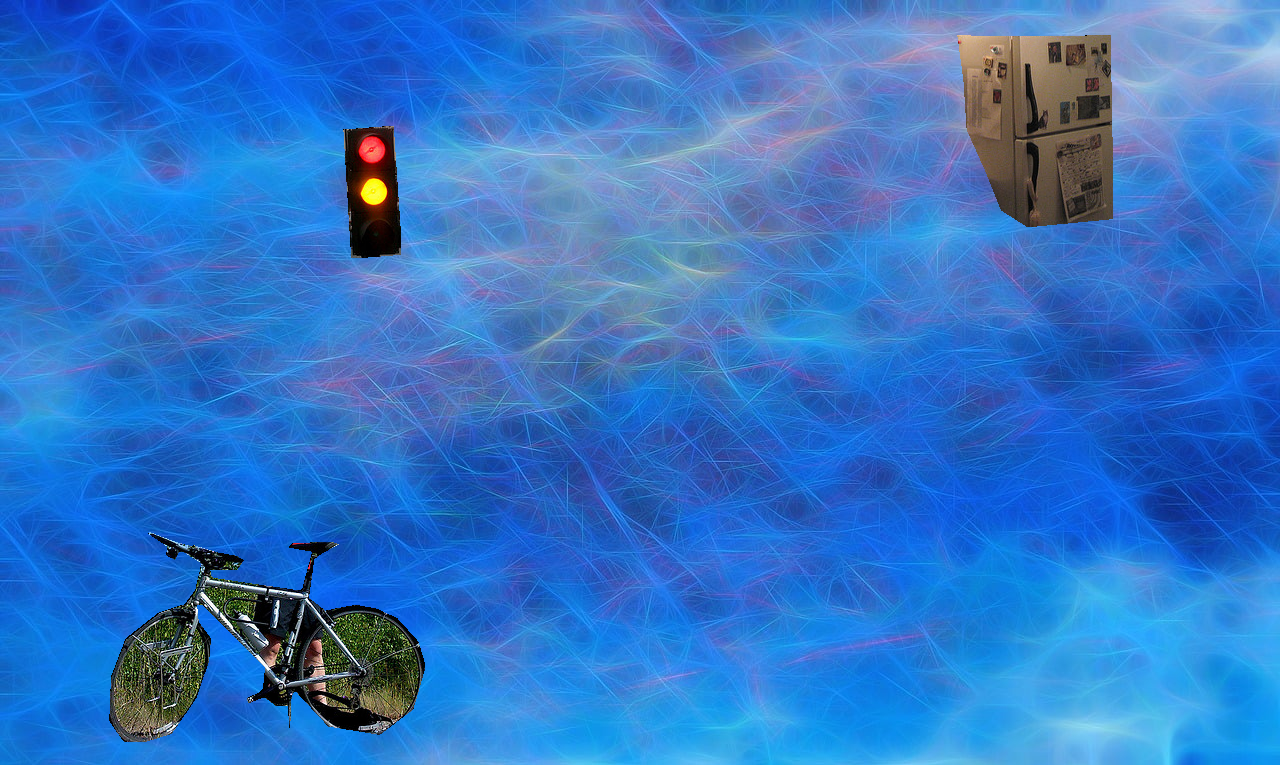}
    \caption{An image from the \cocoscrapbook\ Dataset with presence questions about the concept of ``bicycle''.\\- here, bicycles, are there any?\\- about bicycles... can we find any?\\- here, a bicycle, is there one present?}
    \label{fig:coco_scrapbook_37}
\end{figure}

This preliminary version of the dataset was created to validate the robustness of the baseline models. It comprises 142 images and 126,162 questions, which are divided into 1,627 subgroups. These questions were generated using objects from the COCO dataset and public domain backgrounds obtained from the internet. The dataset contains an average of 888 questions per image, with at least three questions per subgroup, up to 675. The dataset encompasses a total of 37,290 presence questions, 37,290 counting questions, 49,224 confirmation questions, and 2,358 recognition questions. Although a total of 126,162 questions were generated, only a limited number of concepts are enabled for validating the initial level of the four possible levels described in Section~\ref{sec:methodology}. These concepts are ``bicycle'', ``oven'', ``refrigerator'', ``toilet'', and ``traffic light'' as illustrated in Table~\ref{tab:coco_scrapbook_level_1}. An example of presence questions for the concept of ``bicycle'' is displayed in Figure~\ref{fig:coco_scrapbook_37}.


\begin{table}[!htb]
    \setlength{\tabcolsep}{5mm}
    \centering
    \caption{Level 1 presence and counting concepts enabled for the \cocoscrapbook\ dataset.}
    \label{tab:coco_scrapbook_level_1}
    \small
    \begin{tabular}{lcc}
        \toprule
        \textbf{Concept Value} & \textbf{Type} & \textbf{\# of Questions} \\
        \midrule
        \multirow{2}{*}{bicycle} & 1 & 216 \\
        & 2 & 210 \\
        \multirow{2}{*}{oven} & 1 & 69 \\
        & 2 & 33 \\
        \multirow{2}{*}{refrigerator} & 1 & 189 \\
        & 2 & 72 \\
        \multirow{2}{*}{toilet} & 1 & 81 \\
        & 2 & 345 \\
        \multirow{2}{*}{traffic\_light} & 1 & 291 \\
        & 2 & 132 \\
        \bottomrule
    \end{tabular}
\end{table}

\subsubsection{\cocoplainbgscrapbook}

\begin{figure}[!htb]
    \centering
    \includegraphics[width=0.8\linewidth]{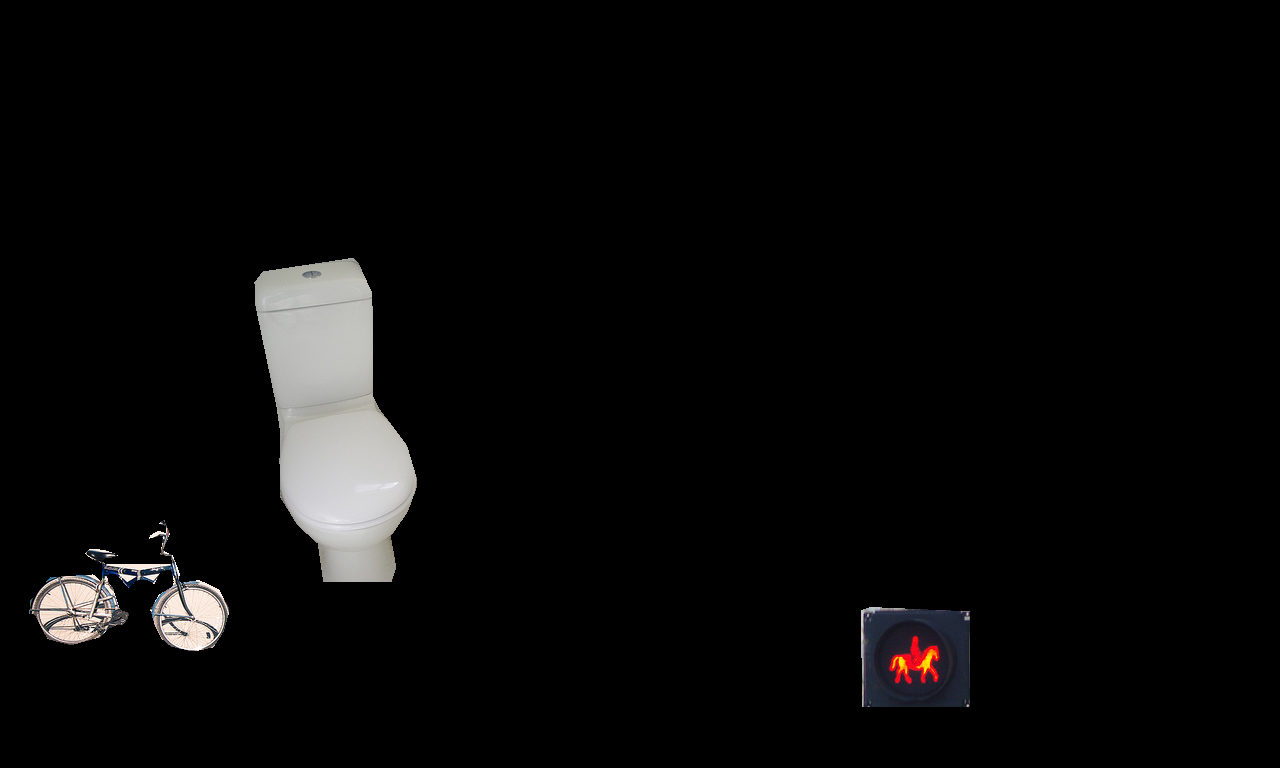}
    \caption{An image from the \cocoplainbgscrapbook\ Dataset with counting questions about the concept of ``toilet''.\\- how many items that are being displayed are toilets?\\- what is the number of toilets that you can find?\\- how many items are toilets here?}
    \label{fig:coco_plain_bg_scrapbook_59}
\end{figure}

This version is a variation of the \cocoscrapbook\ dataset, but with plain colors as the background. It is composed of 161 images and 144,093 questions, which are divided into 1,671 subgroups. Each subgroup contains at least three questions, with a maximum of 780 questions. The set contains 42,606 presence questions, 42,606 counting questions, 56,193 confirmation questions, and 2,688 recognition questions. The concepts enabled for the first level of the \cocoscrapbook\ dataset are also enabled for this one, with the exception of ``oven'', which was replaced by ``sink'', as illustrated in Table~\ref{tab:coco_plain_bg_scrapbook_level_1}. Examples of counting questions for the concept of ``toilet'' are displayed in Figure~\ref{fig:coco_plain_bg_scrapbook_59}.

\begin{table}[!htb]
    \setlength{\tabcolsep}{5mm}
    \centering
    \caption{Level 1 presence and counting concepts enabled for the \cocoplainbgscrapbook\ dataset.}
    \label{tab:coco_plain_bg_scrapbook_level_1}
    \small
    \begin{tabular}{lcc}
        \toprule
        \textbf{Concept Value} & \textbf{Type} & \textbf{\# of Questions} \\
        \midrule
        \multirow{2}{*}{bicycle} & 1 & 321 \\
        & 2 & 162 \\
        \multirow{2}{*}{refrigerator} & 1 & 108 \\
        & 2 & 54 \\
        \multirow{2}{*}{sink} & 1 & 81 \\
        & 2 & 402 \\
        \multirow{2}{*}{toilet} & 1 & 213 \\
        & 2 & 270 \\
        \multirow{2}{*}{traffic\_light} & 1 & 240 \\
        & 2 & 81 \\
        \bottomrule
    \end{tabular}
\end{table}

\subsubsection{\uniqueallscrapbook}

Following a thorough analysis of the findings from the preceding two versions of the dataset, a decision was made to generate an additional version that incorporates colored geometric shapes within a plain background, designated as \uniqueallscrapbook. The dataset encompasses a total of 162 images and 451,347 questions, categorized into 11,641 distinct subgroups. This dataset is considerably larger than the previous ones, having 119,868 presence questions, 119,868 counting questions, 181,983 confirmation questions, and 29,628 recognition questions. With an average of 2,786 questions per image and at least three questions per subgroup up to 618, this version has two levels of concepts enabled. The disparity in size can be attributed to the utilization of colored geometric shapes as objects, a choice that enables the formulation of additional questions concerning individual characteristics such as color and shape.

\begin{figure}[!htb]
    \centering
    \includegraphics[width=0.8\linewidth]{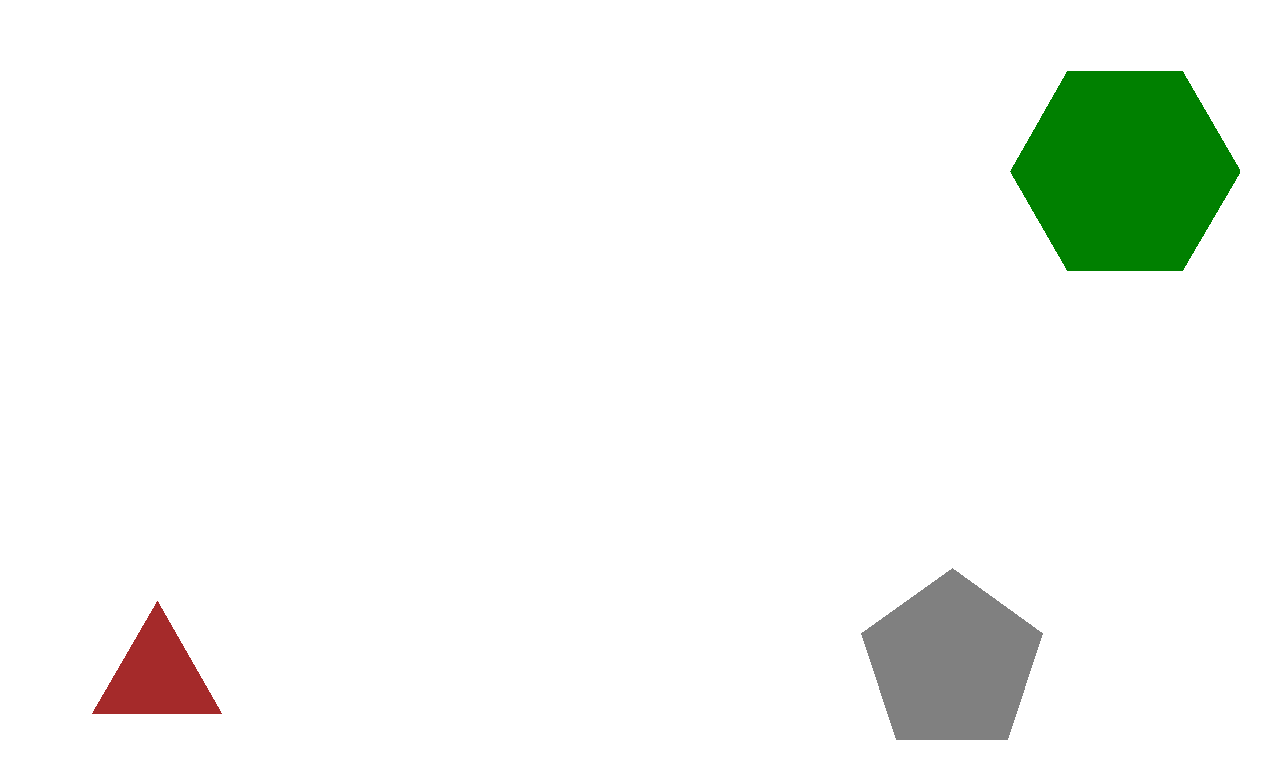}
    \caption{An image from the \uniqueallscrapbook\ Dataset with confirmation questions about the concept of ``pentagon'' including the absolute position ``bottom right''.\\- is the object on the bottom right a pentagon?\\- is it correct that the object on the bottom right is a pentagon?\\- is the stuff on the bottom right a pentagon?}
    \label{fig:unique_all_scrapbook_147}
\end{figure}

The initial level of concepts enabled for the uniqueallscrapbook dataset encompasses the colors ``black'', ``gray'', and ``green'', in addition to the objects ``circle'', ``pentagon'', and ``triangle'', as illustrated in Table~\ref{tab:unique_all_scrapbook_level_1}. The subsequent level of concepts enabled encompasses the absolute positions ``absolute right'', ``down left'', ``down right'', and ``up left''; the colors ``black'' and ``gray''; the object ``pentagon''; and the colored object combination `gray pentagon''. The amounts of some of these questions can be observed in Table~\ref{tab:unique_all_scrapbook_level_2}. An example of counting questions for the concept of ``circle'' is displayed in Figure~\ref{fig:unique_all_scrapbook_147}.

\begin{table}[!htb]
    \setlength{\tabcolsep}{1.8mm}
    \centering
    \caption{Level 1 presence and counting concepts enabled for the \uniqueallscrapbook\ dataset.}
    \label{tab:unique_all_scrapbook_level_1}
    \small
    \begin{tabular}{llcc}
        \toprule
        \textbf{Concept Value} & \textbf{Type} & \textbf{\# of Questions} \\
        \midrule
        \multirow{6}{*}{color} &\multirow{2}{*}{black} & 1 & 108 \\
        & & 2 & 27 \\
        & \multirow{2}{*}{gray} & 1 & 216 \\
        & & 2 & 33 \\
        & \multirow{2}{*}{green} & 1 & 81 \\
        & & 2 & 27 \\
        \hline \multirow{6}{*}{object} &\multirow{2}{*}{circle} & 1 & 108 \\
        & & 2 & 81 \\
        & \multirow{2}{*}{pentagon} & 1 & 216 \\
        & & 2 & 162 \\
        & \multirow{2}{*}{triangle} & 1 & 324 \\
        & & 2 & 162 \\
        \bottomrule
    \end{tabular}
\end{table}

\begin{table}[!htb]
    \setlength{\tabcolsep}{1.8mm}
    \centering
    \caption{Level 2 presence and counting concepts enabled for the \uniqueallscrapbook\ dataset.}
    \label{tab:unique_all_scrapbook_level_2}
    \small
    \begin{tabular}{llcc}
        \toprule
        \textbf{Concept Value} & \textbf{Type} & \textbf{\# of Questions} \\
        \midrule
        \multirow{6}{*}{color} &\multirow{3}{*}{black} & 1 & 108 \\
        & & 2 & 324 \\
        & & 3 & 102 \\
        & \multirow{3}{*}{gray} & 1 & 216 \\
        & & 2 & 648 \\
        & & 3 & 207 \\
        \hline \multirow{3}{*}{object} &\multirow{3}{*}{pentagon} & 1 & 216 \\
        & & 2 & 648 \\
        & & 3 & 537 \\
        \hline \multirow{2}{*}{color + object} &\multirow{2}{*}{gray pentagon} & 1 & 216 \\
        & & 2 & 27 \\
        \bottomrule
    \end{tabular}
\end{table}

\section{Baseline Models}
\label{sec:baselines}

In recent years, the field of artificial intelligence has witnessed considerable advancements, largely due to the development of multimodal large language models (MLLMs), such as Flamingo~\cite{alayrac2022flamingo} and GPT-4V~\cite{openai2024gpt4technicalreport}. Notable developments include the Gemini family models~\cite{team2023gemini,team2024gemini}, as well as the Claude 3 family models~\cite{claude3}, which have demonstrated exceptional performance in image captioning (IC), visual question answering (VQA), and other tasks. Despite their impressive capabilities, their closed-source nature constrains their widespread application and adaptability. In contrast, the open-source landscape for MLLMs is undergoing rapid evolution, with the emergence of models such as LLaVA~\cite{liu2023visual}, LLaVA-1.5~\cite{Liu_2024_CVPR}, VisualGPT~\cite{Chen_2022_CVPR_b}, and InstructBLIP~\cite{dai2023instructblip}, which demonstrate proficiency across a range of vision-language tasks, often attaining results comparable to those of closed-source counterparts.

While some open-source MLLMs exhibit impressive vision-language capabilities, they often require substantial computational resources for training and inference. To illustrate, the entire training and finetuning process of the LLaVA model~\cite{liu2023visual} requires approximately 14 hours on eight A100 GPUs with 80GB of memory each. The training of LLaVA-1.5~\cite{Liu_2024_CVPR} is approximately twice as lengthy as that of LLaVA, with a total of approximately six hours of pretraining and approximately 20 hours of visual instruction tuning, using the same hardware. In view of these considerations, some researchers have put forth the proposition of developing more efficient models by making use of frozen pre-trained models. For example, the Frozen~\cite{NEURIPS2021_01b7575c} model employs a frozen NF-ResNet-50 vision encoder and a seven billion parameter transformer language model. BLIP-2~\cite{pmlr-v202-li23q} combines a ViT-g/14 vision transformer model from EVA-CLIP and an instruction-trained FlanT5 model. These models have demonstrated competitive performance in vision-language tasks while requiring significantly fewer computational resources for training and inference.

This work will focus on smaller multimodal models, including MiniGPT-4~\cite{zhu2024minigpt}, TinyGPT-V~\cite{yuan2024tinygptv}, and MobileVLM V2~\cite{chu2024mobilevlm}. Notably, TinyGPT-V~\cite{yuan2024tinygptv} and MobileVLM V2~\cite{chu2024mobilevlm} have exhibited impressive performance in vision-language tasks while necessitating considerably fewer computational resources for training and inference. These models are well-suited for local deployment and inference on mobile devices, thereby making them accessible to a broader audience. For this same reason, it is imperative that they be subjected to more rigorous scrutiny, particularly in regard to their capacity to comprehend the interrelationship between vision and language, as well as their ability to generate accurate and consistent responses.

In this section, we provide an overview of the baseline models that will be used in our experiments. We will describe the main characteristics of each model, including their architecture, training process, and performance on a variety of vision-language tasks.

\subsection{MobileVLM V2}

Building upon the previous work of~\cite{chu2023mobilevlm}, \citet{chu2024mobilevlm} proposed MobileVLM V2, a family of significantly improved vision language models. The authors posit that a sophisticated integration of novel architectural design, an optimized training methodology tailored for mobile VLMs, and a comprehensive, high-quality dataset curation can significantly enhance VLMs' performance, paving a promising path towards advanced AI in resource-limited scenarios.

\begin{figure}[!htb]
    \centering
    \includegraphics[width=0.95\linewidth]{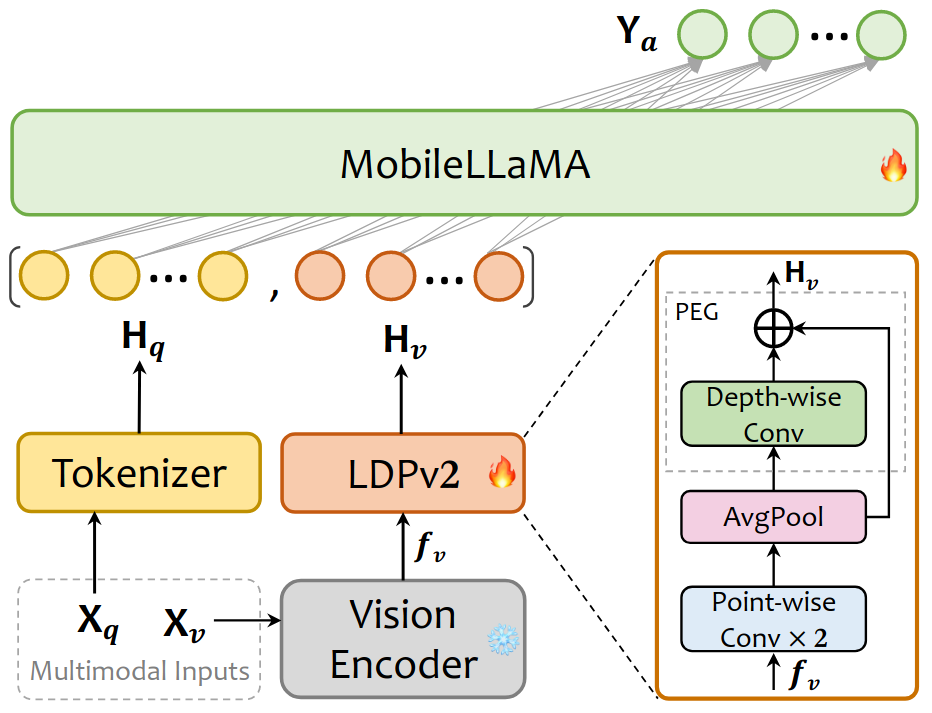}
    \caption{Diagram of MobileVLM-V2 model. Extracted from~\cite{chu2024mobilevlm}.}
    \label{fig:diagram_mobilevlm-v2}
\end{figure}

In a similar fashion to MobileVLM~\cite{chu2023mobilevlm}, the architecture of MobileVLM V2 comprises (as illustrated in Figure~\ref{fig:diagram_mobilevlm-v2}): a pre-trained CLIP ViT-L/14 serving as the vision encoder to extract image features; and a MobileLLa MA-1.4B-Chat and MobileLLaMA-2.7B-Chat are employed as pre-trained large language models to process multi-modal tokens and generate final answers. Additionally, a mobile-friendly lightweight downsample projector (denoted as LDPv2), inspired by the LDP design of MobileVLM, is utilized to perform vision-language feature alignment with fewer parameters. The key improvements of MobileVLM V2 are primarily focused on three key areas: the utilisation of contributive training data on small VLMs, the investigation of effective training strategies, and the development of a high-performance lightweight projector.

\paragraph{Contributive training data on small VLMs}

The training process is divided into two stages: pre-training and multi-task training. During the pre-training stage, the model was trained using the ShareGPT4V-PT dataset, comprising 1.2 million high-quality image-text pairs, with the objective of effectively aligning vision-language features. In this stage, LDPv2 is randomly initialized, and both the vision encoder and the language model are initialized with the pre-trained weights from CLIP ViT-L/14 and MobileLLaMA, respectively. This initialization serves as a robust foundation for the subsequent unified training process. The pretraining phase involves a global batch size of 256 across 8 NVIDIA A100 GPUs for approximately 5 hours.

In the multi-task training phase, the authors utilized a multitude of datasets encompassing a diverse range of tasks to further enhance the model's versatility and proficiency. These datasets were meticulously selected with the objective of enhancing conversational abilities through the usage of the Visual Dialog dataset [19], improving OCR skills through the TextVQA dataset [60], developing scene understanding capabilities through the utilisation of the COCO Caption [11] and SBU [54] datasets, and advancing location understanding through the VSR dataset [43], among others.

It should be noted that the SBU dataset used is a cleaned and refined version, and therefore the data volume does not correspond exactly to the officially released version. In total, the aggregated data for this phase consists of 2.4 million samples, thereby ensuring a comprehensive learning experience across different modalities and tasks. The weights of MobileVLM V2 are initialized from the results of the first stage. The training process requires the use of eight NVIDIA A100 GPUs for approximately nine hours with a global batch size of 128.

\paragraph{Exploring effective training strategies}

With regard to the training paradigm, the authors conducted comprehensive training of all parameters of the projector and language model during both the pre-training and instruction tuning stages. This approach proved advantageous in harnessing the full potential of superior-quality data. In contrast to the training paradigm typically employed for most vision-language models (VLMs), MobileVLM V2 enables the full training of both the projector and the LLM in both stages while fixing the vision encoder. It should be noted that freezing ViT also has the effect of reducing the training cost. The model's training objective is then concentrated on the prediction of the next token, utilizing an autoregressive loss function. By focusing on this specific task, the model is better equipped to learn the intricacies of language generation in the context of visual information, which in turn leads to improved performance on multimodal tasks.

\paragraph{Novel high-performance lightweight projector}

A new projector was developed based on the LDP design of MobileVLM, with the objective of achieving superior vision-language feature alignment with a reduced number of parameters. The system comprises three principal components: feature transformation, token reduction, and positional information enhancement. Initially, the authors implement two point-wise convolution layers on image tokens with the objective of aligning the feature dimension of LLM. Subsequently, an average pooling layer is introduced to significantly reduce the number of image tokens. In conclusion, a straightforward yet highly effective module, PEG [16], with a skip connection, is employed to enhance positional information. In comparison to LDP, this positional component is more efficient, exhibiting a notable reduction in the number of parameters (by 99.8\%) and a slight improvement in processing speed.

In a similar manner to MobileVLM, the authors adopt a series of benchmarks, including the image question answering series GQA [29], SQA [49], TextVQA [60], the comprehensive benchmarks MME [25], MMBench [47], and the object hallucination benchmark POPE [40]. Despite the fact that the MobileVLM V2 models have been designed for deployment in resource-constrained environments, they demonstrate superior performance to the majority of previous models, with a notable margin, while the training cost is comparable to that of the computationally efficient LLaVA-1.5 [45]. MobileVLM V2 3B exhibits a 75\% faster processing speed than the recently proposed MoE-LLaVA-2.7B$\times$4 [41] model, achieving an average score that is 1.4 points higher. By scaling the model to 7B parameters, the authors conduct a comparative analysis with mainstream large-scale VLMs, including LLaVA-1.5 7B [44] and ShareGPT4V 7B, with respect to both accuracy and inference speed. While exhibiting a nearly 20\% increase in speed, the MobileVLM V2 7B model has demonstrated clear superiority over previous SOTA models.

To ensure consistency, the inference latency of the MobileVLM V2 has been measured on the NVIDIA AGX Jetson Orin platform with identical configurations to those used for MobileVLM [15]. The results show that the MobileVLM V2 has a lower inference latency than its counterparts at the same parameter scale.


\subsection{MiniGPT-4}

\begin{figure}[!htb]
    \centering
    \includegraphics[width=0.95\linewidth]{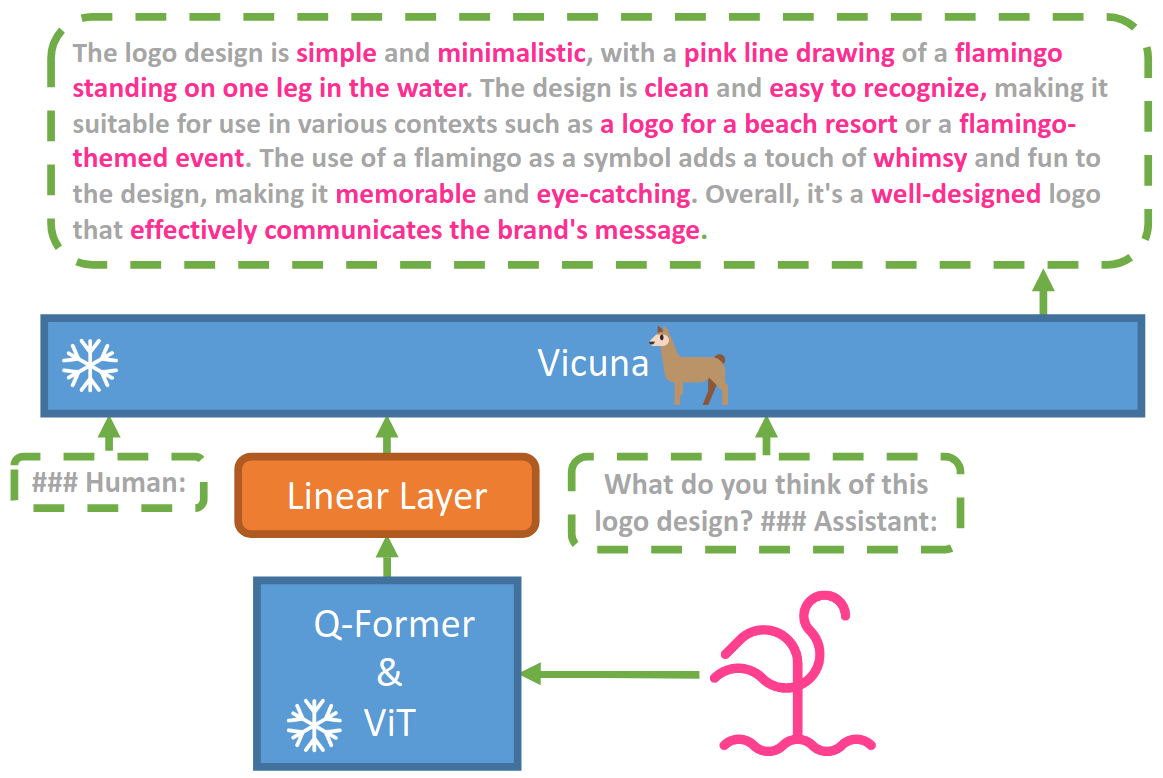}
    \caption{Diagram of MiniGPT-4 model. Extracted from~\cite{zhu2024minigpt}.}
    \label{fig:diagram_minigpt-4}
\end{figure}

In order to align visual information from a pretrained vision encoder with an advanced frozen large language model (LLM), \citet{zhu2024minigpt} proposed the MiniGPT-4 model. The model employs the Vicuna language decoder~\cite{vicuna2023}, which is based on LLaMA and is capable of performing a wide range of complex linguistic tasks~\cite{touvron2023llama}. In regard to visual perception, the same pretrained vision components were utilized as those employed in BLIP-2~\cite{pmlr-v202-li23q}, a ViT-G/14 backbone derived from EVA-CLIP, coupled with the pre-trained Q-Former network. To align the encoded visual features with the Vicuna language model, the authors incorporated a single trainable linear projection layer. A diagram of the model is provided in Figure~\ref{fig:diagram_minigpt-4}.

The model was trained in two stages. In the initial phase, the model was pre-trained on approximately five million pairs of images and short captions for 20,000 steps, which took approximately 10 hours on four A100 GPUs. The objective of this phase was to impart vision-language knowledge to the model. Subsequently, the model was fine-tuned using 3.5k detailed image description pairs with a designed conversational template for 400 steps, which takes approximately seven minutes on a single A100. The objective of this fine-tuning was twofold: to improve the naturalness of the generated language and to ensure both reliability and usability.

During the pretraining phase, the vision and language components are maintained in a frozen state, with only the linear projection layer undergoing training. A combined image captioning dataset comprising images from LAION~\cite{schuhmann2021laion400m}, Conceptual Captions~\cite{changpinyo2021cc12m}, and SBU~\cite{NIPS2011_5dd9db5e} was utilized for this process. The output from the projection layer serves as a soft prompt for the LLM, directing it to generate corresponding ground-truth texts. Following its preliminary training, MiniGPT-4 has demonstrated the capacity to retain a substantial corpus of knowledge and respond in a reasonable manner to human queries. Nevertheless, it persisted in demonstrating constraints in its capacity for robust conversational capabilities. On occasion, the model produces outputs that are incoherent, such as repetitive words or sentences, fragmented phrases, or irrelevant content. This impairs its capacity for fluid visual communication with humans, which resembles that of a chatbot.

To further enhance its ability to produce outputs that are more human-friendly and in line with the concept of instruction fine-tuning datasets, which are largely used in the language realm, the authors curated a smaller, high-quality image-text dataset that is specifically tailored for the purpose of aligning vision and language. The authors employed the model derived from the initial pretraining stage to generate comprehensive descriptions of 5,000 images randomly selected from the Conceptual Caption dataset~\cite{changpinyo2021cc12m}. As the generated image descriptions are affected by issues such as the use of repetitive words or sentences, fragmented sentences, and irrelevant content, the authors utilize ChatGPT with a specific prompt to enhance the descriptions. Subsequently, they manually refine the generated captions by eliminating any redundant words or sentences that ChatGPT is unable to detect. Only approximately 3,500 image-text pairs meet the authors' requirements and are employed for the second-stage alignment process.

The model was evaluated on a dataset comprising 100 images, which were divided into four tasks: meme interpretation, recipe generation, advertisement creation, and poem composition. Each of the aforementioned tasks comprised 25 images. A comparative analysis was conducted by human evaluators between MiniGPT-4 and BLIP-2, as referenced in~\cite{pmlr-v202-li23q}. The former exhibited superior performance, particularly in the domains of recipe, advertisement, and poem generation, effectively addressing 80\% of these tasks. Moreover, the model exhibited proficiency in interpreting humor in memes, accurately identifying the underlying intent in eight out of twenty-five cases, a challenging aspect for BLIP-2. It is worth noting that MiniGPT-4's visual perception remains somewhat limited, with an apparent difficulty in differentiating between spatial localization. For example, the model was unable to identify the location of windows in a given image, which may be due to a lack of aligned image-text data designed for spatial information understanding. The authors propose that training on datasets such as ReferItGame~\cite{kazemzadeh-etal-2014-referitgame} or Visual Genome~\cite{s11263-016-0981-7} may prove an effective solution to this issue.


\subsection{TinyGPT-V}

TinyGPT-V, a novel open-source multimodal large language model (MLLM) designed by~\citeauthor{yuan2024tinygptv}, features a compact yet powerful architecture and is intended to satisfy the growing demand for efficient and high-performance vision language models. The model has been designed for tasks such as image captioning (IC) and visual question answering (VQA), and it offers competitive performance while significantly reducing the computational requirements. By combining a compact Phi-2 language model (2.8 billion parameters) with pre-trained vision encoders and an innovative mapping module, TinyGPT-V achieves noteworthy results on multiple benchmarks with only 24 GB of GPU memory for training and 8 GB for inference. This renders it uniquely suited for resource-constrained devices, including mobile platforms.

\begin{figure}[!htb]
    \centering
    \includegraphics[width=0.95\linewidth]{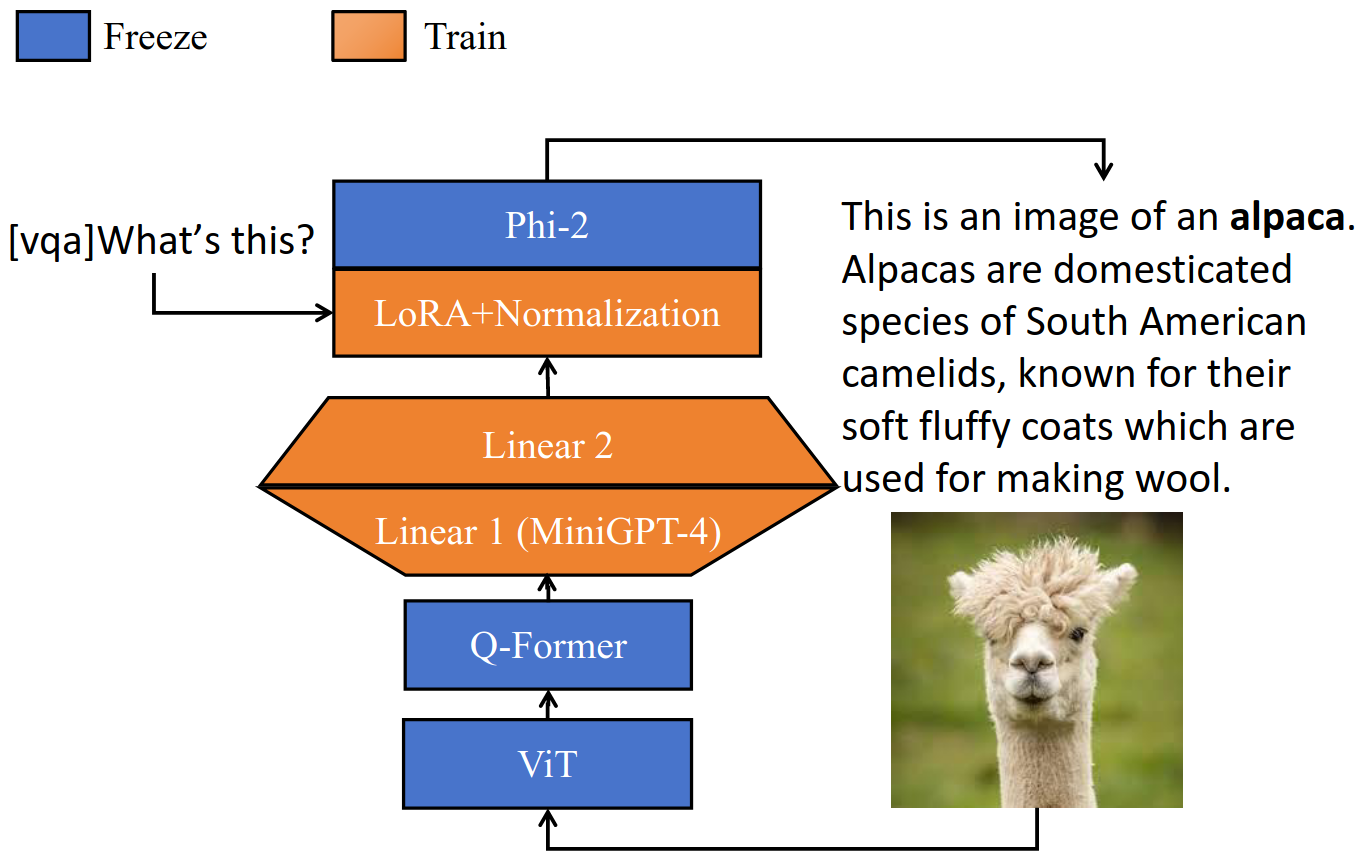}
    \caption{Diagram of TinyGPT-V model. Image from~\cite{yuan2024tinygptv}.}
    \label{fig:diagram_tinygpt-v}
\end{figure}

The model architecture, illustrated in Figure~\ref{fig:diagram_tinygpt-v}, comprises three principal components: a frozen EVA-ViT visual encoder, projection layers based on BLIP-2's Q-Former, and the Phi-2 language model. The projection layers facilitate the integration of visual tokens and language embeddings, while advanced normalization techniques, such as RMS Norm and Query-Key Normalization, promote stability during training and mitigate issues such as gradient vanishing. To enhance efficiency, TinyGPT-V employs LoRA (Low-Rank Adaptation) to fine-tune specific layers while maintaining the cost of inference. The model's innovative 8-bit quantization method enables its effective deployment on lower-end hardware without compromising accuracy.

TinyGPT-V is trained in a structured four-step process, as detailed in Table~\ref{tab:tinygpt-v_datasets}:

\begin{myenumerate}
    \item Warm-up training -- The model is trained to learn vision-language alignment using datasets such as LAION and Conceptual Captions. This establishes the foundation for the model to process image-text input;
    \item Pre-training -- The LoRA module is trained with the objective of improving multimodal data integration and reducing training loss on image-text pairs;
    \item Instruction fine-tuning -- The model is fine-tuned with structured prompts and instructions to enhance its conversational skills and improve response generation;
    \item Multi-Task Learning -- TinyGPT-V is trained on diverse datasets (e.g., Flickr30k, LLaVA dataset, and Unnatural Instruction) to handle complex tasks such as visual spatial reasoning and visual question answering.
\end{myenumerate}

\begin{table*}[!htb]
    \setlength{\tabcolsep}{2mm}
    \centering
    \caption{XXXXX.}
    \label{tab:tinygpt-v_datasets}
    \small
    \begin{tabular}{lp{6cm}cccc}
        \toprule
        \textbf{Data Types} & \textbf{Dataset} & \textbf{Stage 1} & \textbf{Stage 2} & \textbf{Stage 3} & \textbf{Stage 4} \\
        \midrule
        Image-text pair & LAION, CC3M, SBU & $v$ & $v$ & $x$ & $x$ \\
        Instruction tuning & MiniGPT-4 Stage2 for CC \& SBU & $x$ & $x$ & $v$ & $x$ \\
        Caption & Text Captions, COCO Captions & $x$ & $x$ & $x$ & $v$ \\
        REC & RefCOCO, RefCOCO+, RefCOCOg, Visual Genome & $x$ & $x$ & $x$ & $v$ \\
        VQA & GQA, VQAv2, OK-VQA, AOK-VQA, OCR-VQA & $x$ & $x$ & $x$ & $v$ \\
        Multimodal instruction & LLaVA dataset, Flickr30k, Multi-task conversation & $x$ & $x$ & $x$ & $v$ \\
        Language dataset & Unnatural Instructions & $x$ & $x$ & $x$ & $v$ \\
        \bottomrule
    \end{tabular}
\end{table*}

Despite its smaller size, TinyGPT-V exhibits robust performance on a number of benchmarks. For example, it attains a superior zero-shot score of 54.7\% in Visual-Spatial Reasoning (VSR) tasks, is on par with larger models such as InstructBLIP in IconQA and VizWiz, and presents concise, pragmatic responses in real-world scenarios. In comparison to other models, such as LLaVA and MiniGPT-4, which require up to 23.5GB of memory, TinyGPT-V demonstrates a significantly enhanced response speed (0.067 seconds per word) and a reduced memory footprint (5.6GB).

A qualitative analysis indicates that TinyGPT-V is capable of providing accurate and succinct responses, outperforming models such as GPT-4V in generating practical answers to complex visual reasoning tasks. For instance, it demonstrates an ability to avoid incorrect suggestions or excessive verbosity when identifying objects or describing scenes.


\subsection{Discussion}

Due to their multimodal characteristics, optimal performance on a range of tasks, and joint training on diverse datasets, these models are suitable candidates for deployment in a real-world scenario, which renders them well-suited for the scrutiny of the \scrapbook\ Datasets. Prior research has demonstrated the significance of conventional pre-training tasks, such as visual question answering (VQA) and image captioning, in this context. However, tasks such as detection, visual grounding, and grounded captioning were not included in various studies. In their experiments, \citet{pmlr-v162-wang22al} found that tasks which predict regions are crucial for multimodal tasks, with increased performance in image captioning and VQA.

This indicates that visual detection and grounding, in addition to grounded captioning, facilitate models' comprehension of precise alignments between vision and language. The contribution of region information to image generation from text is minimal, as this task requires only a limited alignment of region and text information. Intriguingly, detection has been shown to enhance visual comprehension, suggesting that incorporating region information may be essential for visual comprehension, particularly in images with complex objects. A pivotal next step in this direction is validating positioning information.


\section{Performance Evaluation}
\label{sec:metrics}

A plethora of metrics for VQA problems have been introduced in recent studies~\cite{j.imavis.2021.104327}. Despite the variety of metrics available and the multitude of ways they can be used, we have selected accuracy as the primary form of validation. This is because it provides a general overview and a single value for the model. The accuracy is given by the following equation:

\begin{equation}
\text{Acuraccy} = \frac{\# \text{ correct answers}}{\# \text{ total answers}}
\end{equation}

Furthermore, we endorse the utilization of the consistency metric proposed by~\citet{Hudson_2019_CVPR}. This metric assesses the consistency of responses across a set of questions, determining whether a visual question answering system does not provide contradictory information when presented with a new query. For example, an appropriate response to a new question about an apple that has just been identified as red would be not green. As a result of creating multiple iterations of the same question in our \scrapbook\ dataset, this metric can be employed to assess the consistency of the responses generated by the models, taking into account their language comprehension abilities. In considering the consistency metric, a threshold of 0.8 was introduced, indicating that the model should provide an accurate response at least 80\% of the time for the same question posed in different forms.

Another metric that is employed is a variant of the aforementioned consistency metric, which takes into account the precedence of the image. During the process of generating the \scrapbook\ dataset, it is determined whether the image is an entire new composition or a previously created image with an additional object. This enables verification of the model's capability to provide consistent responses across consecutive images. To illustrate this point, consider the presentation of an image depicting a dog and a subsequent image adding a cat. In this scenario, the model should respond by indicating the presence of a dog in both images, as the presence of an additional object does not occlude the previous object. This ability to disregard superfluous information and focus on the essential elements pertinent to the posed question is crucial for assessing the model's capacity to provide consistent and accurate responses across different consecutive images.

In order to calculate the metrics, five different approaches might be considered with regard to the versions of the same question posed to the model. These versions include the question itself, with additional conditioning (e.g., answer with the object shape, when applicable), with a given direction (e.g., based on the scene, answer this question with a short sentence), or by selecting from a list of possible answers. The calculation of the metrics can be performed on any one of the four versions. Alternatively, an answer can be deemed valid if it was answered correctly for all four versions of the question. This final approach is the most rigorous, and although we believe that this should be the primary methodology, we will employ all five to assess the models.

Given the open-ended nature of all models, which are capable of generating any response, it was necessary to adopt a cautious approach to the evaluation of the generated answers. The evaluation of the generated responses is based on the variation of the question, and the following criteria are applied: if the question does not have additional information or has a given direction, the expected answer must be contained in the generated answer; conversely, when the question involves additional conditions or the selection of answers from a given list, the generated answer must precisely match the expected answer. The model errors are then categorized to facilitate further analysis of its performance. The following categories are employed to structure this analysis:

\begin{myitemize}
    \item wrong answer -- the anticipated response was ``green'', yet the model provided ``yellow'' as answer, indicating a plausible but erroneous response;
    \item more than one answer -- the model provided multiple plausible answers (e.g., ``yes and no'');
    \item unexpected answer -- for example, the correct answer should have been a shape, but the model answered something different;
    \item invalidated by simpler image -- the model provided a correct answer to the question but an erroneous one to the same question asked in a version of the image with fewer objects.
\end{myitemize}

Additionally, error categories were devised for the aggregated answers, which correspond to the responses generated by the models considering all versions of a given question. These categories are the same when all answers are in agreement. However, to accommodate instances where there is discordance, additional categories have been introduced:

\begin{myitemize}
    \item answer disagreement -- the model provided distinct responses to the same question presented in varying formats. However, all responses deemed plausible (e.g., different colors, yes and no);
    \item error disagreement -- the same question may yield different errors when posed in different versions. For instance, an incorrect response may be given in one version, while in another, an unexpected answer may be provided.
\end{myitemize}

As previously stated in Section~\ref{sec:methodology}, the dataset was divided into distinct levels of complexity based on the number of concepts required to accurately answer the question. To assess the models, we will calculate the accuracy for each level of complexity. Once a concept has been correctly validated at the initial level, the model will be evaluated for the subsequent level, incorporating that concept. This process will continue until all levels have been considered. Additionally, we will calculate the error categories for the individual and aggregated answers to facilitate a comprehensive analysis of the models' performance.

Additional potential metrics for evaluating the response have been contemplated, including those that take into account both the textual response and the visual region utilized in the decision-making process for models that generate a visual grounding. One potential metric is the Evidence-based Evaluation (EvE) metric, as proposed by~\citet{Wang_2020_CVPR}. This metric requires that a VQA model provide evidence in the form of bounding boxes as a visual grounding for the generated response. An alternative approach would be to employ metrics that consider not only the visual reasoning and the generated response, but also the rationale employed. In this context, the AiR-E metric, as defined by~\citet{978-3-030-58452-8_6}, represents a particularly promising option. The metric measures the alignment of the attention map with regions of interest relevant for reasoning. When evaluated by humans, attention maps that led to correct answers exhibited higher AiR-E scores, while incorrect attention maps with lower scores failed to focus on the most important object, suggesting a potential correlation between AiR-E and task performance. The investigation of additional metrics is a subject for future research.

\section{Validating Results}

The baseline models introduced in Section~\ref{sec:baselines} were implemented and evaluated on all versions of the \scrapbook\ dataset. Their results were then compared across versions to assess the validity of the approach. An overview of the models's results can be visualized in Table~\ref{tab:models_results}. The MiniGPT-4 model achieved the highest accuracy, with 64.15\% on the \cocoscrapbook\ Dataset, 66.69\% on the \cocoplainbgscrapbook\ Dataset, and 19.41\% on the \uniqueallscrapbook\ Dataset, considering only the part of the datasets that do not include absurd questions. Despite the satisfactory overall performance of the model, it will be demonstrated that the model exhibits deficiencies in comprehending fundamental concepts, such as colors and shapes. This will be evidenced through an analysis of the results according to level and question type.

\begin{table*}[!htb]
    \setlength{\tabcolsep}{2.8mm}
    \centering
    \caption{Models overall results for all \textit{Scrapbook} datasets.}
    \label{tab:models_results}
    \small
    \begin{tabular}{c|c|c|cr|cr|cr}
        \hline
        \multirow{2}{*}{Dataset} & \multirow{2}{*}{Filter} & \multirow{2}{*}{\# Questions} & \multicolumn{2}{c|}{MobileVLM-V2} & \multicolumn{2}{c|}{MiniGPT-4} & \multicolumn{2}{c}{TinyGPT-V} \\ \cline{4-9}
        & & & \multicolumn{1}{c|}{Corrects} & \% & \multicolumn{1}{c|}{Corrects} & \% & \multicolumn{1}{c|}{Corrects} & \% \\
        \hline
        & Non-absurd & \multirow{2}{*}{74,844} & \multicolumn{1}{r|}{7,491} & 10.01 & \multicolumn{1}{r|}{48,016} & 64.15 & \multicolumn{1}{r|}{8,217} & 10.98 \\
        COCO & Non-absurd* & & \multicolumn{1}{r|}{5,149} & 6.88 & \multicolumn{1}{r|}{44,590} & 59.58 & \multicolumn{1}{r|}{7,065} & 9.44 \\
        Scrapbook & Full & \multirow{2}{*}{141,129} & \multicolumn{1}{r|}{7,491} & 5.31 & \multicolumn{1}{r|}{48,016} & 34.02 & \multicolumn{1}{r|}{8,217} & 5.82 \\
        & Full* & & \multicolumn{1}{r|}{5,149} & 3.65 & \multicolumn{1}{r|}{44,590} & 31.60 & \multicolumn{1}{r|}{7,065} & 5.01 \\
        \hline
        & Non-absurd & \multirow{2}{*}{52,833} & \multicolumn{1}{r|}{2,936} & 5.56 & \multicolumn{1}{r|}{35,236} & 66.69 & \multicolumn{1}{r|}{5,123} & 9.70 \\
        COCO Plain BG & Non-absurd* & & \multicolumn{1}{r|}{1,950} & 3.69 & \multicolumn{1}{r|}{32,964} & 62.39 & \multicolumn{1}{r|}{4,824} & 9.13 \\
        Scrapbook & Full & \multirow{2}{*}{100,773} & \multicolumn{1}{r|}{2,936} & 2.91 & \multicolumn{1}{r|}{35,236} & 34.97 & \multicolumn{1}{r|}{5,123} & 5.08 \\
        & Full* & & \multicolumn{1}{r|}{1,950} & 1.94 & \multicolumn{1}{r|}{32,964} & 32.71 & \multicolumn{1}{r|}{4,824} & 4.79 \\
        \hline
        & Non-absurd & \multirow{2}{*}{330,618} & \multicolumn{1}{r|}{21,958} & 6.64 & \multicolumn{1}{r|}{64,177} & 19.41 & \multicolumn{1}{r|}{48,694} & 14.73 \\
        Unique All & Non-absurd* & & \multicolumn{1}{r|}{17,300} & 5.23 & \multicolumn{1}{r|}{57,364} & 17.35 & \multicolumn{1}{r|}{45,261} & 13.69 \\
        Scrapbook & Full & \multirow{2}{*}{593,745} & \multicolumn{1}{r|}{21,958} & 3.70 & \multicolumn{1}{r|}{64,177} & 10.81 & \multicolumn{1}{r|}{48,694} & 8.20 \\
        & Full* & & \multicolumn{1}{r|}{17,300} & 2.91 & \multicolumn{1}{r|}{57,364} & 9.66 & \multicolumn{1}{r|}{45,261} & 7.62 \\
        \hline
        \multicolumn{8}{l}{* Invalidating a correct answer when it contradicts a wrong answer in a simpler image.}
    \end{tabular}
\end{table*}

The results of the models for the first level are presented in the format of stacked bars, with each question group (e.g., ``presence'') comprising five bars. The aggregated result for a given question type is represented by the first bar, and it is used to verify the model's comprehension of the concept. The subsequent bars represent the potential variations of the question addendum, including a condition, a direction, an enumeration, and the absence of an addendum. The color coding of the bars signifies the quantitative percentage of each bar, with green denoting values above the 80\% threshold and red indicating otherwise.


\subsection{MobileVLM-V2}

The MobileVLM-V2 model was the least performant, attaining an accuracy of merely 10\% for the non-absurd questions in the \cocoscrapbook\ dataset, as illustrated in Table~\ref{tab:models_results}. The majority of these errors stemmed from disagreements between answers to different versions of the same question posed to the model. These errors accounted for approximately 56,300 instances in the \cocoscrapbook\ dataset and 40,900 instances in the \cocoplainbgscrapbook\ dataset, as illustrated in Table~\ref{tab:mobilevlm-v2_errors}. This phenomenon occurs when the model provides correct responses to one version of a question, such as without any addendum, yet provides incorrect responses to another, for example, with added direction.The addition of 141,129 absurd questions to the dataset results in a significant decline in model accuracy, with a decrease to 5.3\%. A similar performance decline was observed for the other datasets presented in Table~\ref{tab:models_results}.

\begin{table*}[!htb]
    \setlength{\tabcolsep}{2.8mm}
    \centering
    \caption{MobileVLM-V2 error types for all \textit{Scrapbook} datasets.}
    \label{tab:mobilevlm-v2_errors}
    \small
    \begin{tabular}{c|cr|cr|cr}
        \hline
        \multirow{2}{*}{Errors} & \multicolumn{2}{c|}{\cocoscrapbook} & \multicolumn{2}{c|}{\cocoplainbgscrapbook} & \multicolumn{2}{c}{\uniqueallscrapbook} \\ \cline{2-7}
        & \multicolumn{1}{c|}{Non-absurd} & All & \multicolumn{1}{c|}{Non-absurd} & All & \multicolumn{1}{c|}{Non-absurd} & All \\
        \hline
        Answer disagreement & \multicolumn{1}{c|}{\textcolor{red}{54,290}} & 56,352 & \multicolumn{1}{c|}{\textcolor{red}{40,890}} & 42,142 & \multicolumn{1}{c|}{\textcolor{red}{170,557}} & 177,611 \\
        Error disagreement & \multicolumn{1}{c|}{2,789} & 6,127 & \multicolumn{1}{c|}{1,538} & 2,570 & \multicolumn{1}{c|}{32,453} & 48,386 \\
        Plausible wrong answers & \multicolumn{1}{c|}{10,238} & 71,123 & \multicolumn{1}{c|}{7,429} & 53,085 & \multicolumn{1}{c|}{105,190} & 345,330 \\
        Unexpected answers & \multicolumn{1}{c|}{36} & 36 & \multicolumn{1}{c|}{40} & 40 & \multicolumn{1}{c|}{460} & 460 \\
        Invalidated by simpler image & \multicolumn{1}{c|}{2,342} & 2,342 & \multicolumn{1}{c|}{986} & 986 & \multicolumn{1}{c|}{4,658} & 4,658 \\
        Total & \multicolumn{1}{c|}{69,695} & 135,980 & \multicolumn{1}{c|}{50,883} & 98,823 & \multicolumn{1}{c|}{313,318} & 576,445 \\
        \hline
    \end{tabular}
\end{table*}

\begin{figure}[!htb]
    \centering
    \includegraphics[width=\linewidth]{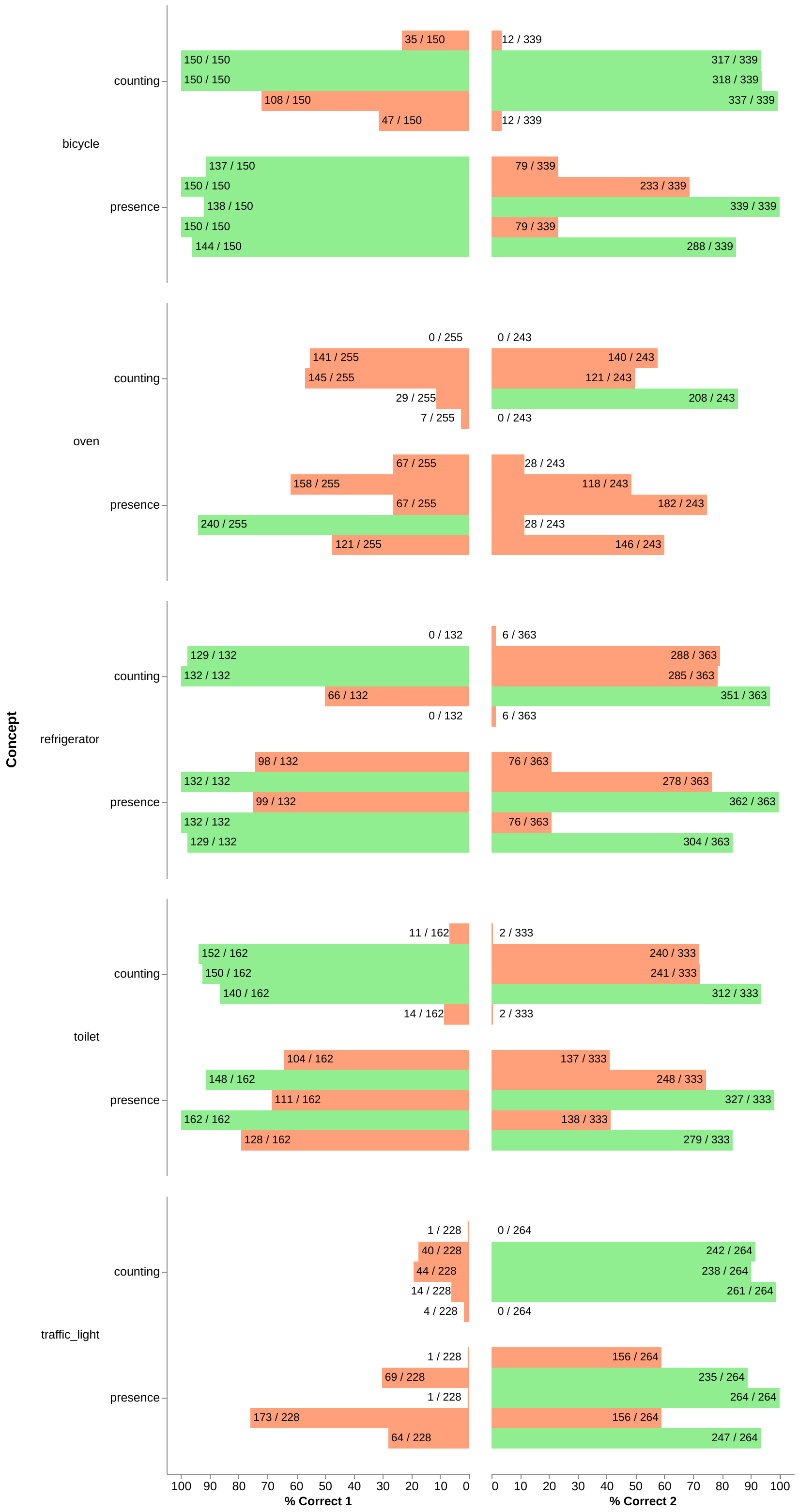}
    \caption{MobileVLM-V2 level 1 results on \cocoscrapbook.}
    \label{fig:mobilevlm-v2_coco_results}
\end{figure}

\begin{figure}[!htb]
    \centering
    \includegraphics[width=\linewidth]{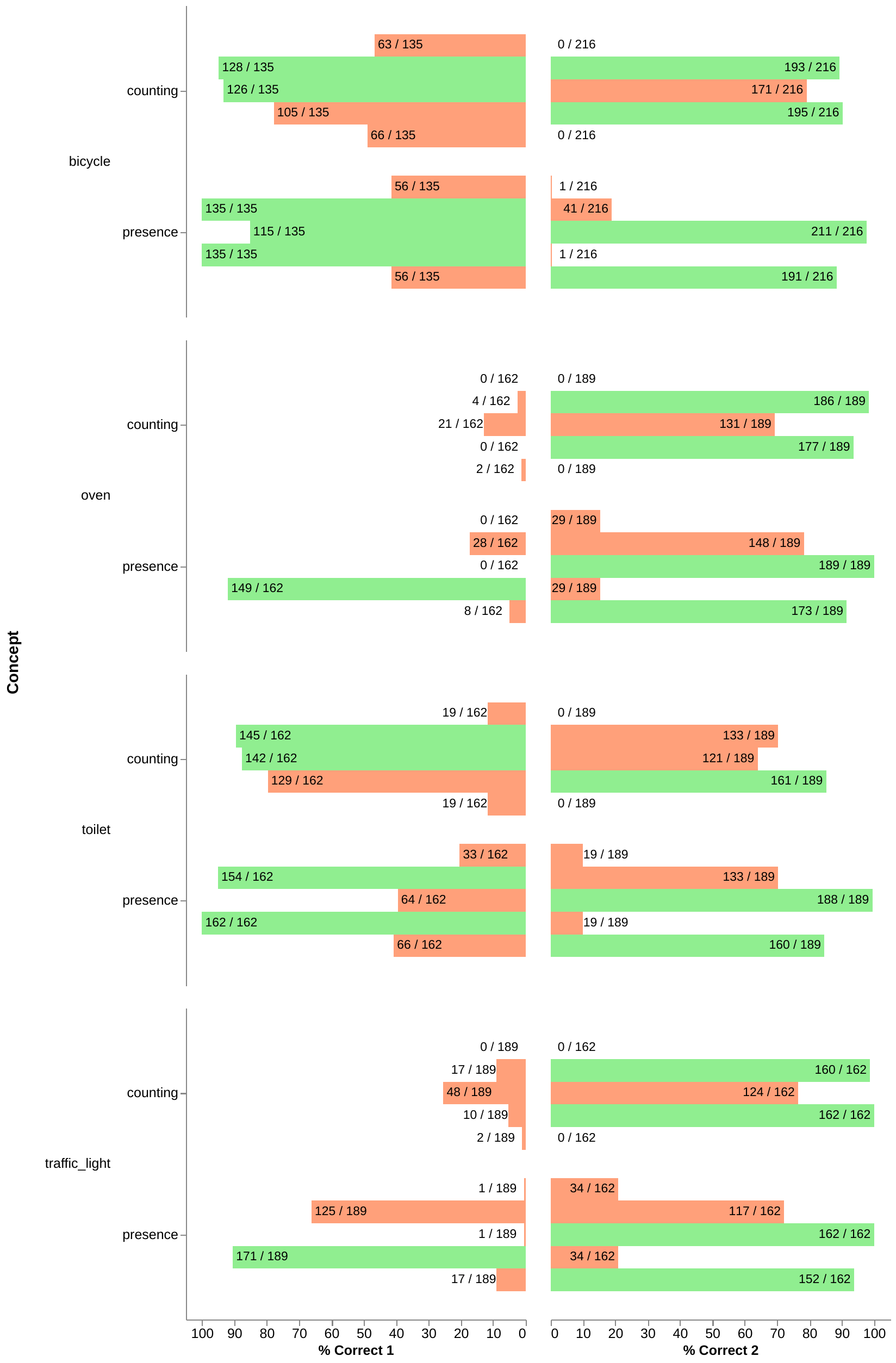}
    \caption{MobileVLM-V2 level 1 results on \cocoplainbgscrapbook.}
    \label{fig:mobilevlm-v2_coco_plain_bg_results}
\end{figure}

\begin{figure}[!htb]
    \centering
    \includegraphics[width=\linewidth]{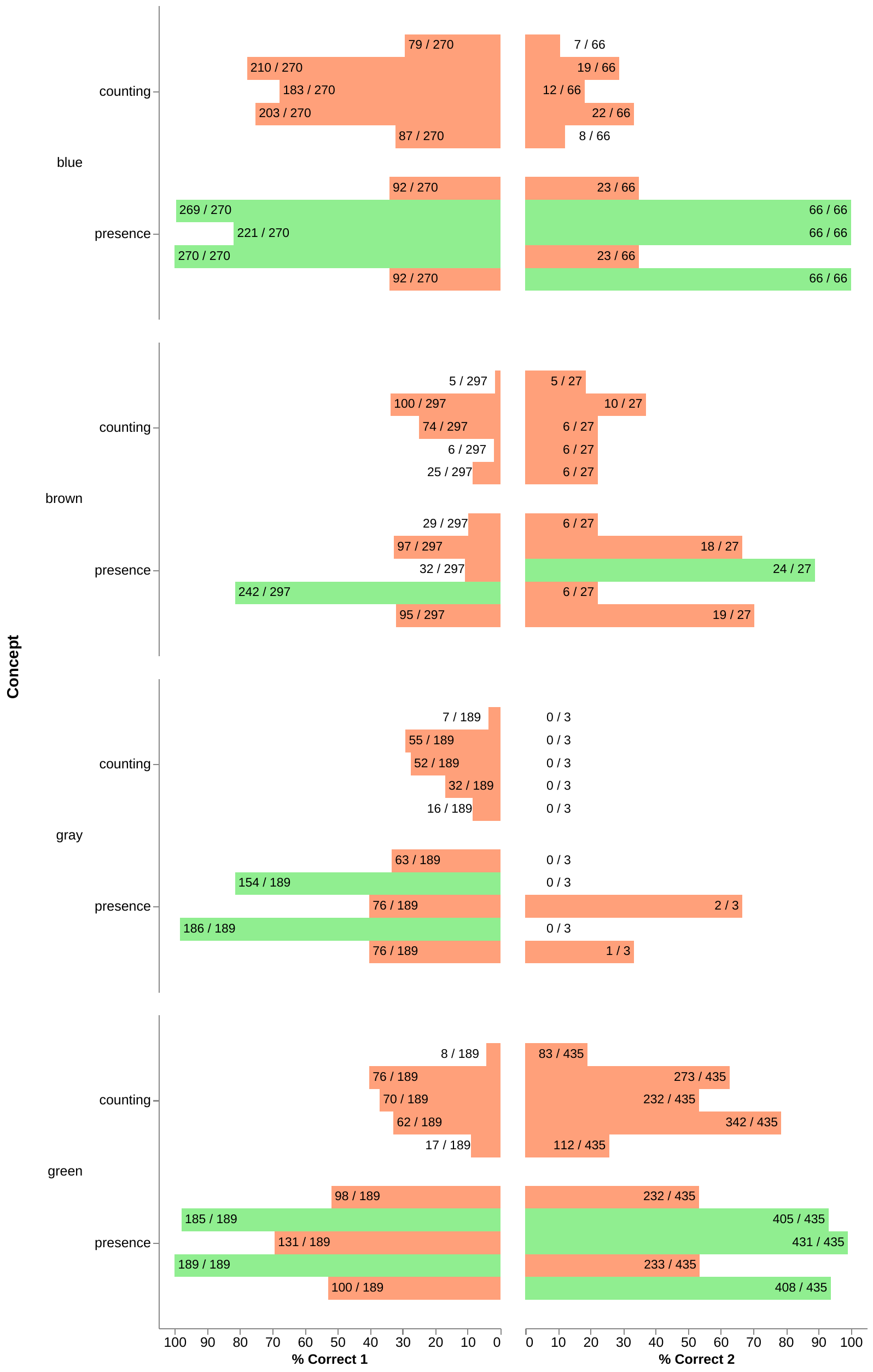}
    \caption{MobileVLM-V2 level 1 color results on \uniqueallscrapbook.}
    \label{fig:mobilevlm-v2_unique_all_deterministic_colors_results}
\end{figure}

\begin{figure}[!htb]
    \centering
    \includegraphics[width=\linewidth]{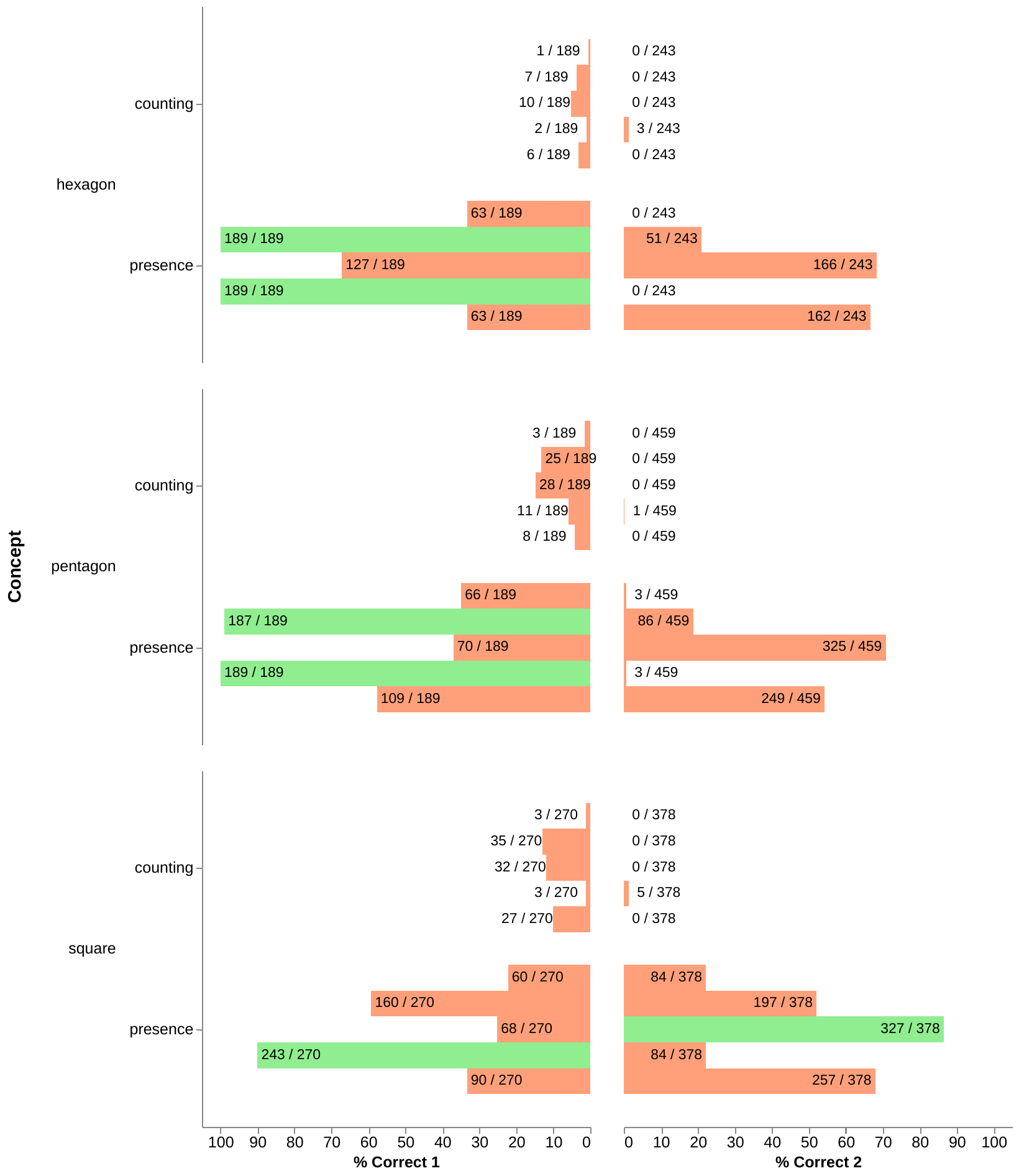}
    \caption{MobileVLM-V2 level 1 objects results on \uniqueallscrapbook.}
    \label{fig:mobilevlm-v2_unique_all_deterministic_objs_results}
\end{figure}

\begin{figure}[!htb]
    \centering
    \includegraphics[width=\linewidth]{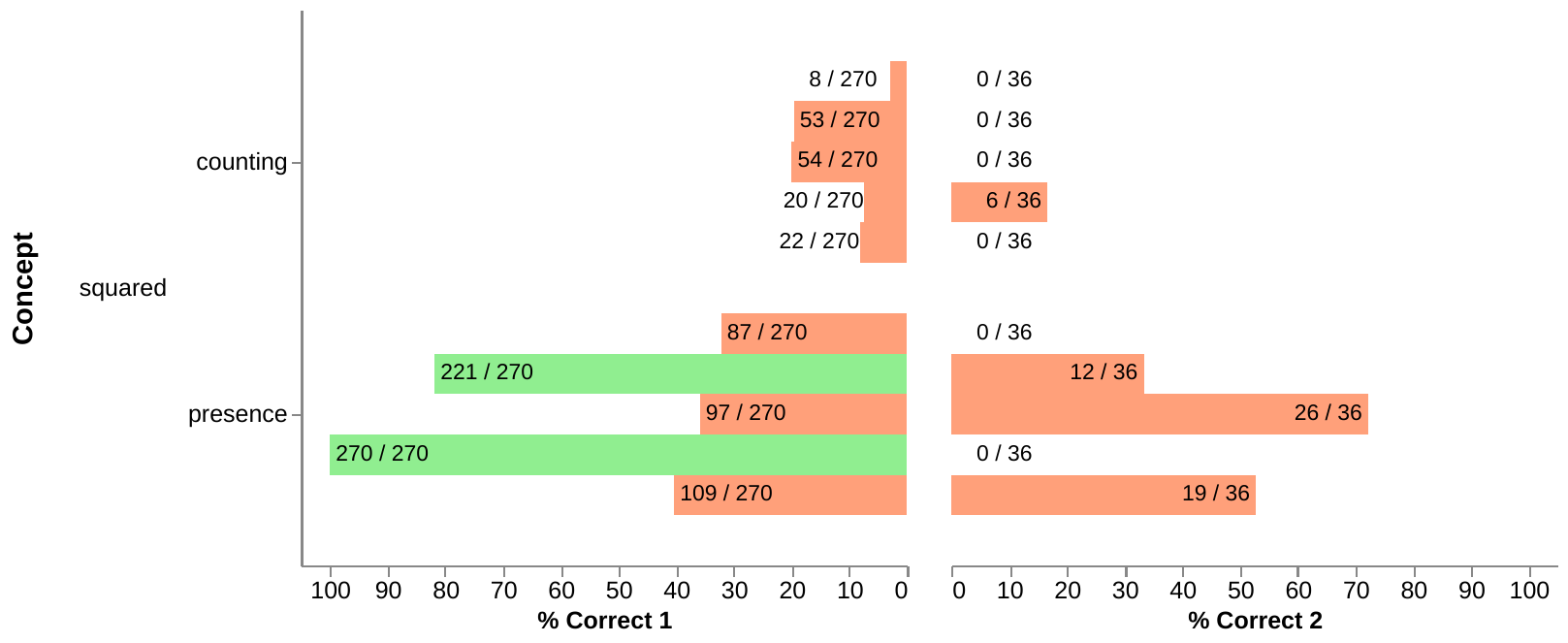}
    \caption{MobileVLM-V2 level 1 shapes results on \uniqueallscrapbook.}
    \label{fig:mobilevlm-v2_unique_all_deterministic_shapes_results}
\end{figure}

The MobileVLM-V2 model's performance for level 1 questions can be seen in Figure~\ref{fig:mobilevlm-v2_coco_results} for the \cocoscrapbook\ Dataset. The model demonstrated an inability to accurately answer counting questions for most of the concepts unless additional constraints were imposed, such as the introduction of a condition or enumeration of possible outcomes. The model's optimal performance was observed when a list of possible answers was provided, prompting responses such as ``choose one of the alternatives: yes, no, or not applicable''. A similar trend was observed when the model was evaluated on the \cocoplainbgscrapbook\ and \uniqueallscrapbook\ Datasets, as illustrated in Figures~\ref{fig:mobilevlm-v2_coco_plain_bg_results}, \ref{fig:mobilevlm-v2_unique_all_deterministic_colors_results}, \ref{fig:mobilevlm-v2_unique_all_deterministic_objs_results}, and \ref{fig:mobilevlm-v2_unique_all_deterministic_shapes_results}.

A comprehensive review of Figures~\ref{fig:mobilevlm-v2_unique_all_deterministic_objs_results} and \ref{fig:mobilevlm-v2_unique_all_deterministic_shapes_results} indicates that the model exhibited an inability to respond to count questions concerning geometric shapes across all question types. This finding suggests a potential inadequacy in the training data concerning shapes. This deficiency is indicative of the model's suboptimal understanding of the concept of shapes.


\subsection{MiniGPT-4}

The MiniGPT-4 model demonstrated the most optimal overall performance among all models evaluated, as evidenced by its accuracy rates across the \textit{Scrapbook} datasets. Table~\ref{tab:minigpt-4_errors} provides a detailed breakdown of the error types encountered by the model.

For the \cocoscrapbook\ dataset, MiniGPT-4 achieved an accuracy of 64.15\%. However, the introduction of absurd questions significantly impacted its performance, reducing the accuracy to 34.02\%. The primary source of errors in this dataset was plausible wrong answers, which accounted for 49.3\% of the errors in the non-absurd filter and 85.4\% in the full dataset. This phenomenon occurs when the model provides a plausible wrong answer for all variations of the question, such as responding with ``bicycle'' when the correct answer is ``refrigerator'', which indicates that while the model often provided incorrect answers, these answers were still contextually reasonable.

Answer disagreements were also notable, comprising 46.9\% of the errors in the non-absurd filter and 13.5\% in the full dataset, meaning the model ofter contradicts itself when answering different forms of the same question. Error disagreements were relatively low, at 3.3\% and 1.0\% respectively, suggesting a degree of consistency in the model's responses.



In the \cocoplainbgscrapbook\ and \uniqueallscrapbook\ datasets, the model's accuracy also decreased with the inclusion of absurd questions. Plausible wrong answers remained the dominant error type, making up most of the errors in the non-absurd filter, going up to 88.5\% in the full \cocoplainbgscrapbook\ dataset. For the \uniqueallscrapbook\ dataset, answer disagreements were 36.9\% and 18.6\% respectively, while error disagreements were 4.4\% and 3.0\%, showing a higher level of inconsistency compared to the other datasets.


\begin{table*}[!htb]
    \setlength{\tabcolsep}{2.8mm}
    \centering
    \caption{MiniGPT-4 error types for all \textit{Scrapbook} datasets.}
    \label{tab:minigpt-4_errors}
    \small
    \begin{tabular}{c|cr|cr|cr}
        \hline
        \multirow{2}{*}{Errors} & \multicolumn{2}{c|}{\cocoscrapbook} & \multicolumn{2}{c|}{\cocoplainbgscrapbook} & \multicolumn{2}{c}{\uniqueallscrapbook} \\ \cline{2-7}
        & \multicolumn{1}{c|}{Non-absurd} & All & \multicolumn{1}{c|}{Non-absurd} & All & \multicolumn{1}{c|}{Non-absurd} & All \\
        \hline
        Answer disagreement & \multicolumn{1}{c|}{12,572} & 12,572 & \multicolumn{1}{c|}{7,019} & 7,019 & \multicolumn{1}{c|}{98,338} & 98,338 \\
        Error disagreement & \multicolumn{1}{c|}{878} & 902 & \multicolumn{1}{c|}{389} & 389 & \multicolumn{1}{c|}{11,663} & 16,037 \\
        Plausible wrong answers & \multicolumn{1}{c|}{\textcolor{red}{13,213}} & 79,474 & \multicolumn{1}{c|}{\textcolor{red}{10,087}} & 58,027 & \multicolumn{1}{c|}{\textcolor{red}{153,016}} & 411,766 \\
        Unexpected answers & \multicolumn{1}{c|}{165} & 165 & \multicolumn{1}{c|}{102} & 102 & \multicolumn{1}{c|}{3,424} & 3,427 \\
        Invalidated by simpler image & \multicolumn{1}{c|}{3,426} & 3,426 & \multicolumn{1}{c|}{2,272} & 2,272 & \multicolumn{1}{c|}{6,813} & 6,813 \\
        Total & \multicolumn{1}{c|}{30,254} & 96,539 & \multicolumn{1}{c|}{19,869} & 67,809 & \multicolumn{1}{c|}{273,254} & 536,381 \\
        \hline
    \end{tabular}
\end{table*}

\begin{figure}[!htb]
    \centering
    \includegraphics[width=\linewidth]{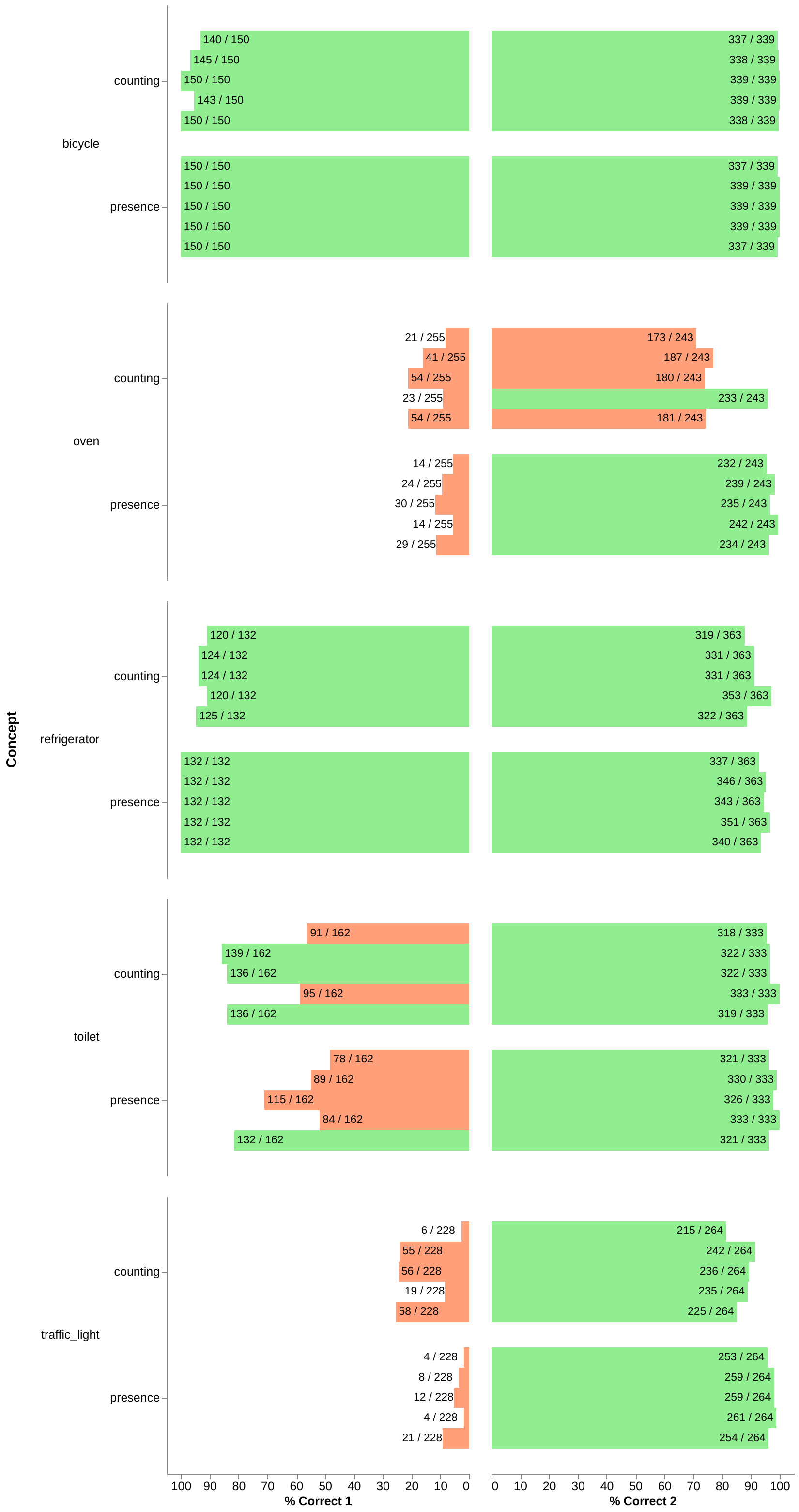}
    \caption{MiniGPT-4 level 1 results on \cocoscrapbook.}
    \label{fig:minigpt-4_coco_results}
\end{figure}

As illustrated in Figure~\ref{fig:minigpt-4_coco_results}, the MiniGPT-4 model demonstrates notable efficacy in addressing the presence and counting tasks for certain concepts, attaining over 90\% accuracy for the concepts of  ``bicycle'' and ``refrigerator''. When evaluating the model's performance on individual question forms, it was observed that the model demonstrated optimal performance when no additional information was provided or when a simple direction was given, such as ``Based on the scene, answer this question with a short sentence''. Conversely, when a condition was provided, such as ``Answer with the object class, when applicable'', or when all possible answers were listed, the performance of the model decreased.

\begin{figure}[!htb]
    \centering
    \includegraphics[width=\linewidth]{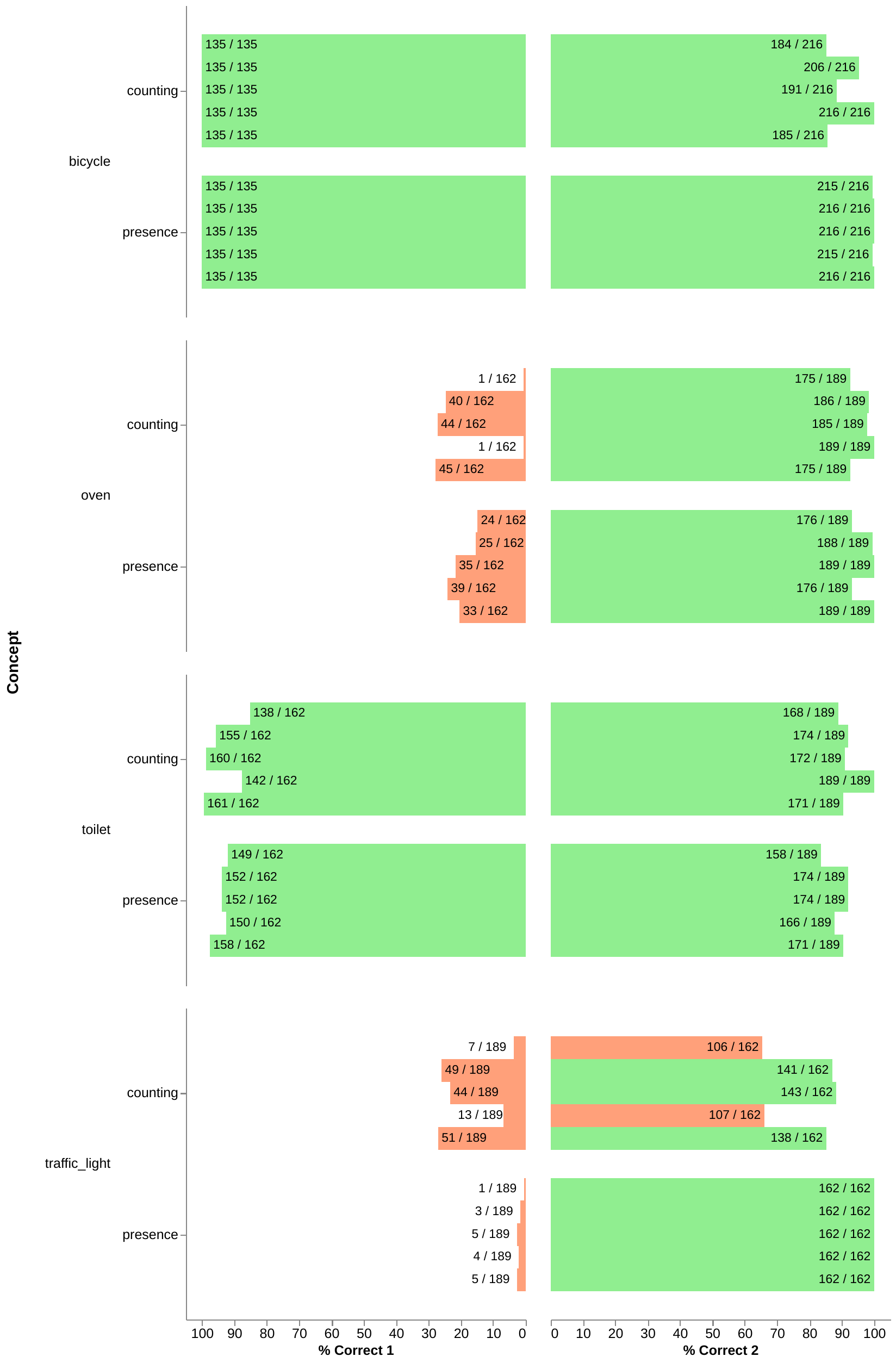}
    \caption{MiniGPT-4 level 1 results on \cocoplainbgscrapbook.}
    \label{fig:minigpt-4_coco_plain_bg_results}
\end{figure}

A similar behavior was observed when the model was tested on the \cocoplainbgscrapbook, as reported in Figure~\ref{fig:minigpt-4_coco_plain_bg_results}. For instance, for these same types of questions, two concepts passed on the first validation level for both datasets based on objects from COCO, being ``bicycle'' and ``refrigerator'' for the \cocoscrapbook\ and ``bicycle'' and ``toilet'' for the \cocoplainbgscrapbook. Given the validation of a few concepts at the initial level, an investigation was conducted to ascertain the model's capacity to address questions concerning absolute positions at the subsequent level. However, the model demonstrated an inability to provide accurate responses when the object was present in a different position within the image, which can be observed in subgroup 2 in Tables~\ref{tab:minigpt-4_coco_lvl_2_bicycle_results} and \ref{tab:minigpt-4_coco_lvl_2_refrigerator_results} for the \cocoscrapbook\ and Tables~\ref{tab:minigpt-4_coco_plain_bg_lvl_2_bicycle_results} and \ref{tab:minigpt-4_coco_plain_bg_lvl_2_toilet_results} for the \cocoplainbgscrapbook.

\begin{table*}[!htb]
    \setlength{\tabcolsep}{2.8mm}
    \centering
    \caption{Results of the MiniGPT-4 model on the \cocoscrapbook\ Dataset for the second validation level with the concept ``bicycle''.}
    \label{tab:minigpt-4_coco_lvl_2_bicycle_results}
    \small
    \begin{tabular}{c|c|cr|cr|cr}
        \hline
        \multirow{2}{*}{Position} & \multirow{2}{*}{Type} & \multicolumn{2}{c|}{Subgroup 1} & \multicolumn{2}{c|}{Subgroup 2} & \multicolumn{2}{c}{Subgroup 3} \\ \cline{3-8}
        & & \multicolumn{1}{c|}{Corrects} & \% & \multicolumn{1}{c|}{Corrects} & \% & \multicolumn{1}{c|}{Corrects} & \% \\
        \hline
        \multirow{2}{*}{absolute right} & Presence & \multicolumn{1}{r|}{39 / 39} & 100.0 & \multicolumn{1}{r|}{\textcolor{red}{0 / 111}} & \textcolor{red}{0.0} & \multicolumn{1}{r|}{339 / 354} & 95.8 \\
        & Counting & \multicolumn{1}{r|}{36 / 39} & 92.3 & \multicolumn{1}{r|}{\textcolor{red}{0 / 111}} & \textcolor{red}{0.0} & \multicolumn{1}{r|}{330 / 354} & 93.2 \\
        \hline
        \multirow{2}{*}{center} & Presence & \multicolumn{1}{r|}{21 / 21} & 100.0 & \multicolumn{1}{r|}{\textcolor{red}{0 / 129}} & \textcolor{red}{0.0} & \multicolumn{1}{r|}{338 / 342} & 98.8 \\
        & Counting & \multicolumn{1}{r|}{21 / 21} & 100.0 & \multicolumn{1}{r|}{\textcolor{red}{0 / 129}} & \textcolor{red}{0.0} & \multicolumn{1}{r|}{339 / 342} & 99.1 \\
        \hline
        \multirow{2}{*}{down left} & Presence & \multicolumn{1}{r|}{9 / 9} & 100.0 & \multicolumn{1}{r|}{\textcolor{red}{7 / 141}} & \textcolor{red}{4.9} & \multicolumn{1}{r|}{339 / 339} & 100.0 \\
        & Counting & \multicolumn{1}{r|}{6 / 9} & 66.7 & \multicolumn{1}{r|}{\textcolor{red}{0 / 141}} & \textcolor{red}{0.0} & \multicolumn{1}{r|}{333 / 339} & 98.2 \\
        \hline
        \multirow{2}{*}{up right} & Presence & \multicolumn{1}{r|}{11 / 12} & 91.7 & \multicolumn{1}{r|}{\textcolor{red}{3 / 69}} & \textcolor{red}{4.3} & \multicolumn{1}{r|}{338 / 345} & 98.0 \\
        & Counting & \multicolumn{1}{r|}{12 / 12} & 100.0 & \multicolumn{1}{r|}{\textcolor{red}{0 / 69}} & \textcolor{red}{0.0} & \multicolumn{1}{r|}{331 / 345} & 95.9 \\
        \hline
    \end{tabular}
\end{table*}

\begin{table*}[!htb]
    \setlength{\tabcolsep}{2.8mm}
    \centering
    \caption{Results of the MiniGPT-4 model on the \cocoscrapbook\ Dataset for the second validation level with the concept ``refrigerator''.}
    \label{tab:minigpt-4_coco_lvl_2_refrigerator_results}
    \small
    \begin{tabular}{c|c|cr|cr|cr}
        \hline
        \multirow{2}{*}{Position} & \multirow{2}{*}{Type} & \multicolumn{2}{c|}{Subgroup 1} & \multicolumn{2}{c|}{Subgroup 2} & \multicolumn{2}{c}{Subgroup 3} \\ \cline{3-8}
        & & \multicolumn{1}{c|}{Corrects} & \% & \multicolumn{1}{c|}{Corrects} & \% & \multicolumn{1}{c|}{Corrects} & \% \\
        \hline
        \multirow{2}{*}{absolute right} & Presence & \multicolumn{1}{r|}{54 / 54} & 100.0 & \multicolumn{1}{r|}{\textcolor{red}{0 / 78}} & \textcolor{red}{0.0} & \multicolumn{1}{r|}{348 / 375} & 92.8 \\
        & Counting & \multicolumn{1}{r|}{54 / 54} & 100.0 & \multicolumn{1}{r|}{\textcolor{red}{0 / 78}} & \textcolor{red}{0.0} & \multicolumn{1}{r|}{275 / 375} & 73.3 \\
        \hline
        \multirow{2}{*}{center} & Presence & \multicolumn{1}{r|}{30 / 30} & 100.0 & \multicolumn{1}{r|}{\textcolor{red}{0 / 102}} & \textcolor{red}{0.0} & \multicolumn{1}{r|}{343 / 366} & 93.7 \\
        & Counting & \multicolumn{1}{r|}{18 / 30} & 60.0 & \multicolumn{1}{r|}{\textcolor{red}{0 / 102}} & \textcolor{red}{0.0} & \multicolumn{1}{r|}{318 / 366} & 86.9 \\
        \hline
    \end{tabular}
\end{table*}

\begin{table*}[!htb]
    \setlength{\tabcolsep}{2.8mm}
    \centering
    \caption{Results of the MiniGPT-4 model on the \cocoplainbgscrapbook\ Dataset for the second validation level with the concept ``bicycle''.}
    \label{tab:minigpt-4_coco_plain_bg_lvl_2_bicycle_results}
    \small
    \begin{tabular}{c|c|cr|cr|cr}
        \hline
        \multirow{2}{*}{Position} & \multirow{2}{*}{Type} & \multicolumn{2}{c|}{Subgroup 1} & \multicolumn{2}{c|}{Subgroup 2} & \multicolumn{2}{c}{Subgroup 3} \\ \cline{3-8}
        & & \multicolumn{1}{c|}{Corrects} & \% & \multicolumn{1}{c|}{Corrects} & \% & \multicolumn{1}{c|}{Corrects} & \% \\
        \hline
        \multirow{2}{*}{absolute right} & Presence & \multicolumn{1}{r|}{45 / 45} & 100.0 & \multicolumn{1}{r|}{\textcolor{red}{0 / 45}} & \textcolor{red}{0.0} & \multicolumn{1}{r|}{209 / 234} & 89.3 \\
        & Counting & \multicolumn{1}{r|}{42 / 45} & 93.3 & \multicolumn{1}{r|}{\textcolor{red}{0 / 45}} & \textcolor{red}{0.0} & \multicolumn{1}{r|}{210 / 234} & 89.7 \\
        \hline
        \multirow{2}{*}{center} & Presence & \multicolumn{1}{r|}{45 / 45} & 100.0 & \multicolumn{1}{r|}{\textcolor{red}{0 / 90}} & \textcolor{red}{0.0} & \multicolumn{1}{r|}{214 / 234} & 91.5 \\
        & Counting & \multicolumn{1}{r|}{45 / 45} & 100.0 & \multicolumn{1}{r|}{\textcolor{red}{0 / 90}} & \textcolor{red}{0.0} & \multicolumn{1}{r|}{212 / 234} & 90.6 \\
        \hline
    \end{tabular}
\end{table*}

\begin{table*}[!htb]
    \setlength{\tabcolsep}{2.8mm}
    \centering
    \caption{Results of the MiniGPT-4 model on the \cocoplainbgscrapbook\ Dataset for the second validation level with the concept ``toilet''.}
    \label{tab:minigpt-4_coco_plain_bg_lvl_2_toilet_results}
    \small
    \begin{tabular}{c|c|cr|cr|cr}
        \hline
        \multirow{2}{*}{Position} & \multirow{2}{*}{Type} & \multicolumn{2}{c|}{Subgroup 1} & \multicolumn{2}{c|}{Subgroup 2} & \multicolumn{2}{c}{Subgroup 3} \\ \cline{3-8}
        & & \multicolumn{1}{c|}{Corrects} & \% & \multicolumn{1}{c|}{Corrects} & \% & \multicolumn{1}{c|}{Corrects} & \% \\
        \hline
        \multirow{2}{*}{absolute right} & Presence & \multicolumn{1}{r|}{30 / 30} & 100.0 & \multicolumn{1}{r|}{\textcolor{red}{0 / 36}} & \textcolor{red}{0.0} & \multicolumn{1}{r|}{143 / 210} & 68.1 \\
        & Counting & \multicolumn{1}{r|}{30 / 30} & 100.0 & \multicolumn{1}{r|}{\textcolor{red}{0 / 36}} & \textcolor{red}{0.0} & \multicolumn{1}{r|}{158 / 210} & 75.2 \\
        \hline
        \multirow{2}{*}{center} & Presence & \multicolumn{1}{r|}{24 / 24} & 100.0 & \multicolumn{1}{r|}{\textcolor{red}{2 / 138}} & \textcolor{red}{1.4} & \multicolumn{1}{r|}{143 / 195} & 73.3 \\
        & Counting & \multicolumn{1}{r|}{18 / 24} & 75.0 & \multicolumn{1}{r|}{\textcolor{red}{2 / 138}} & \textcolor{red}{1.4} & \multicolumn{1}{r|}{170 / 195} & 87.2 \\
        \hline
        \multirow{2}{*}{up left} & Presence & \multicolumn{1}{r|}{11 / 12} & 91.7 & \multicolumn{1}{r|}{\textcolor{red}{7 / 150}} & \textcolor{red}{4.7} & \multicolumn{1}{r|}{149 / 192} & 77.6 \\
        & Counting & \multicolumn{1}{r|}{12 / 12} & 100.0 & \multicolumn{1}{r|}{\textcolor{red}{6 / 150}} & \textcolor{red}{4.0} & \multicolumn{1}{r|}{175 / 192} & 91.1 \\
        \hline
    \end{tabular}
\end{table*}

\begin{figure}[!htb]
    \centering
    \includegraphics[width=\linewidth]{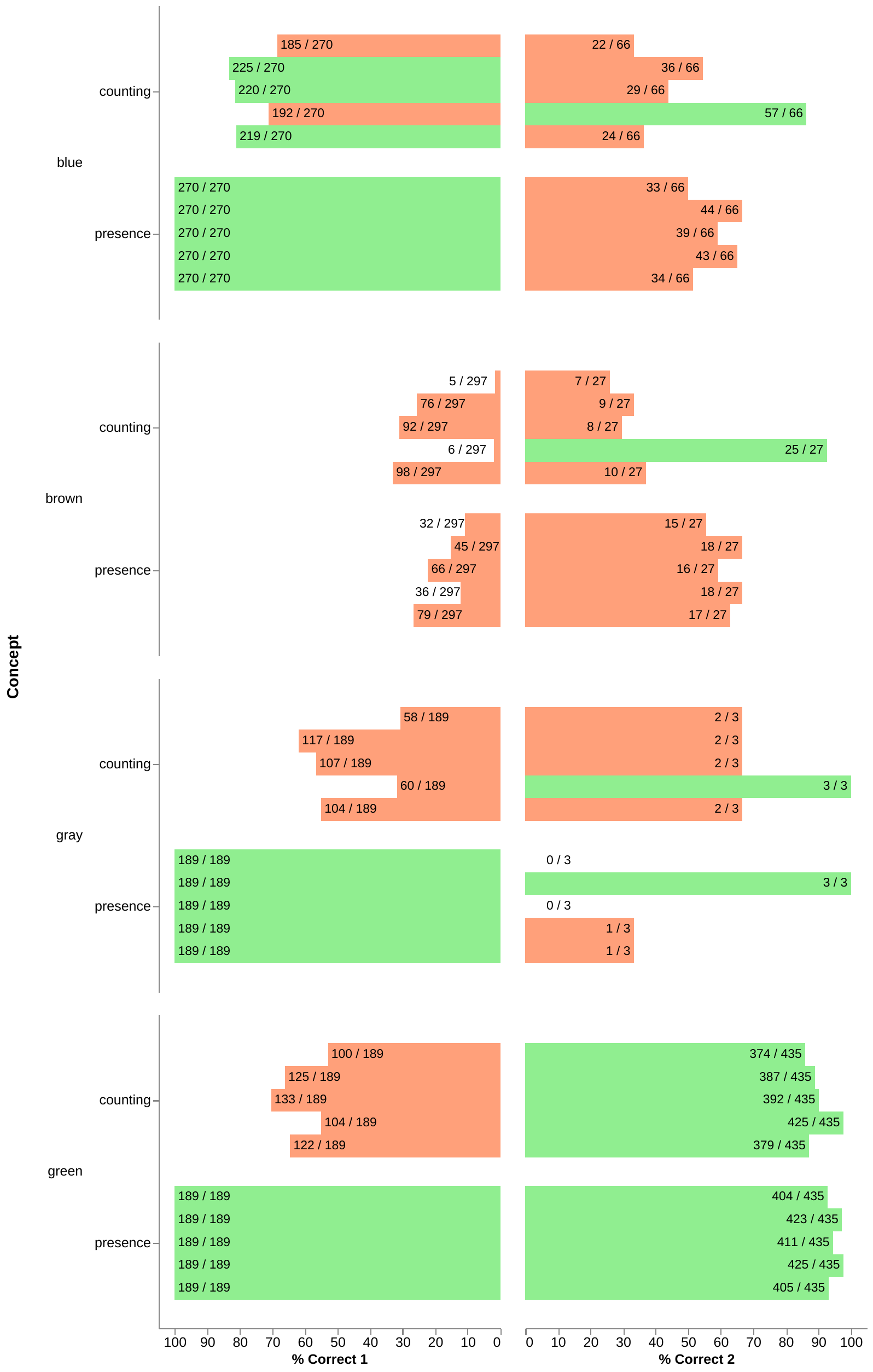}
    \caption{MiniGPT-4 level 1 color results on \uniqueallscrapbook.}
    \label{fig:minigpt-4_unique_all_deterministic_colors_results}
\end{figure}

\begin{figure}[!htb]
    \centering
    \includegraphics[width=\linewidth]{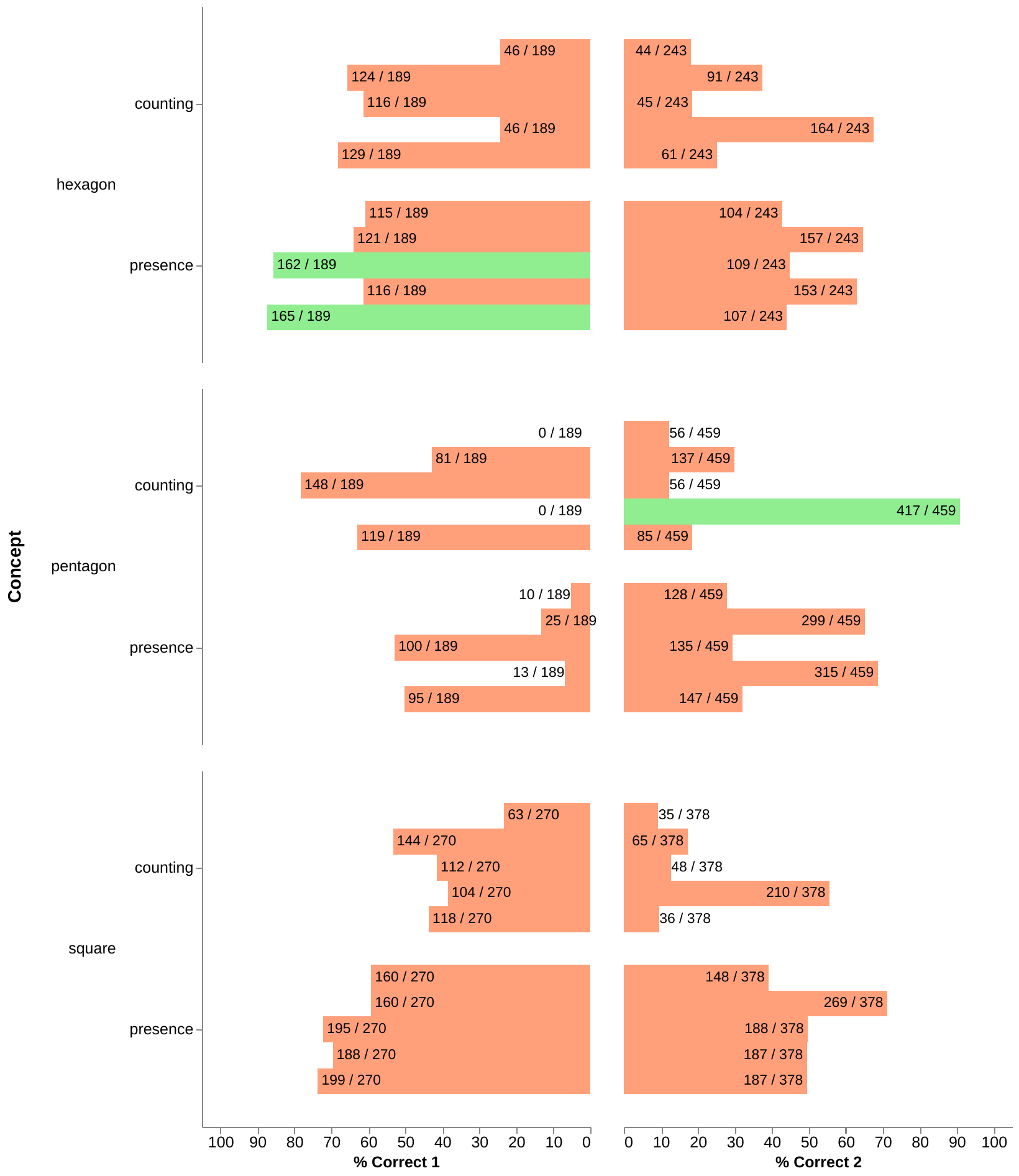}
    \caption{MiniGPT-4 level 1 objects results on \uniqueallscrapbook.}
    \label{fig:minigpt-4_unique_all_deterministic_objs_results}
\end{figure}

\begin{figure}[!htb]
    \centering
    \includegraphics[width=\linewidth]{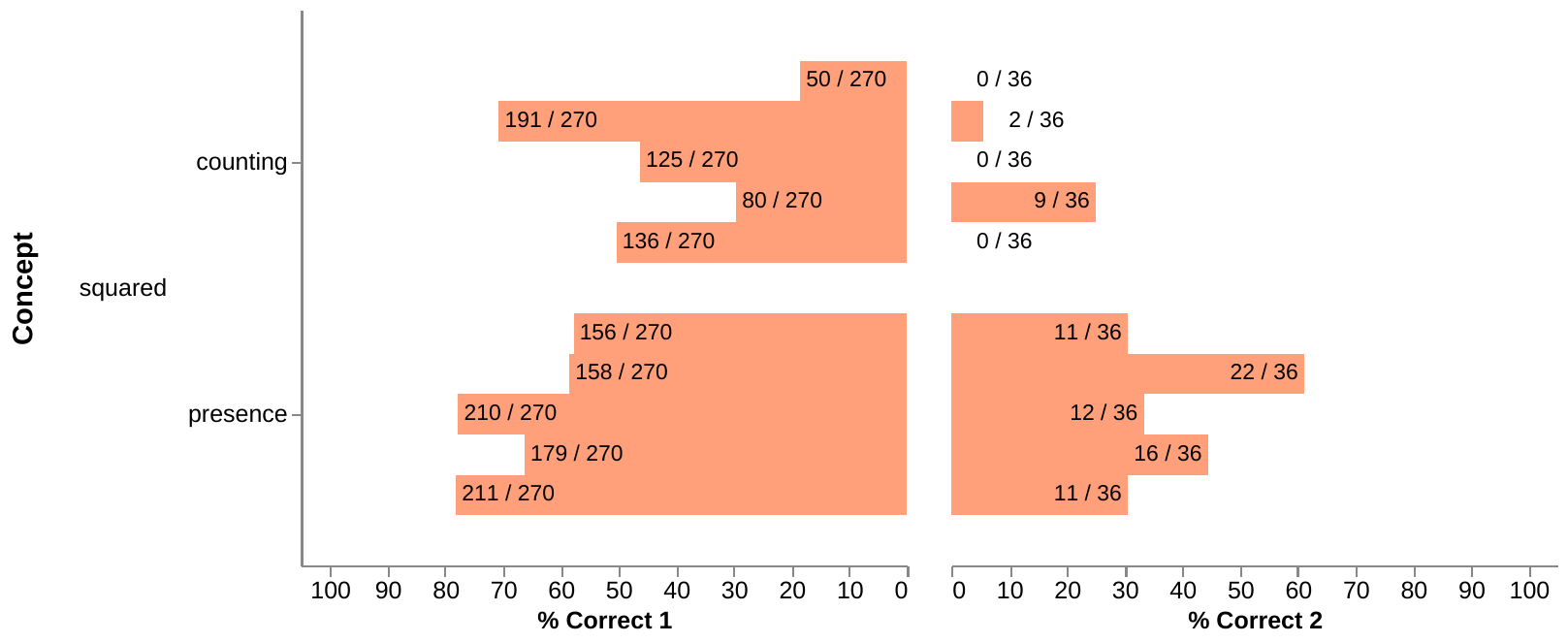}
    \caption{MiniGPT-4 level 1 shapes results on \uniqueallscrapbook.}
    \label{fig:minigpt-4_unique_all_deterministic_shapes_results}
\end{figure}

When evaluated with the \uniqueallscrapbook\ Dataset, as displayed in Figures~\ref{fig:minigpt-4_unique_all_deterministic_colors_results}, \ref{fig:minigpt-4_unique_all_deterministic_objs_results}, and \ref{fig:minigpt-4_unique_all_deterministic_shapes_results}, the results were not entirely consistent. The question forms that led to a decrease in performance on the \cocoscrapbook\ dataset resulted in an increase in performance on the \uniqueallscrapbook\ dataset, but only for subgroup 2. The results for subgroup 1 followed a similar trend as observed in the other datasets, with the model performing better when no additional information was provided or a simple direction was given. Notably, in this particular dataset version, the model failed to accurately identify any of the concepts in the initial level, underscoring its deficiencies in comprehending fundamental concepts such as color and shape.

\subsection{TinyGPT-V}


The TinyGPT-V model's performance across the \textit{Scrapbook} datasets reveals several key insights, as detailed in Table~\ref{tab:tinygpt-v_errors}.

For the \cocoscrapbook\ dataset, the model exhibited a significant proportion of answer disagreements, accounting for approximately 40.8\% of the errors in the non-absurd filter and 20.5\% in the full dataset. This indicates that the model frequently provided answers that contradicts itself on zat least one of the variations. Additionally, the model showed a high percentage of plausible wrong answers, with 42.5\% in the non-absurd filter and 71.1\% in the full dataset. This suggests that while the model's answers were often incorrect, they were still reasonable within the context of the questions posed. The error disagreement rates were also notable, with 16.1\% in the non-absurd filter and 8.2\% in the full dataset, indicating some inconsistency in the model's responses.



In the \cocoplainbgscrapbook\ dataset, the model's performance was similar, with answer disagreements making up 19.1\% of the errors in the non-absurd filter and 9.5\% in the full dataset. The plausible wrong answers remained high, at 69.3\% and 84.1\% respectively, reinforcing the trend observed in the \cocoscrapbook\ dataset. For the \uniqueallscrapbook\ dataset, the model's performance showed a different pattern. The answer disagreements were relatively low, at 6.5\% in the non-absurd filter and 3.4\% in the full dataset. However, the plausible wrong answers were significantly higher, at 70.9\% and 82.0\% respectively. This indicates that while the model's answers were often incorrect, they were still contextually plausible. The error disagreement rates were 18.9\% in the non-absurd filter and 12.6\% in the full dataset, suggesting a higher level of inconsistency in this dataset compared to the others.

\begin{figure}[!htb]
    \centering
    \includegraphics[width=\linewidth]{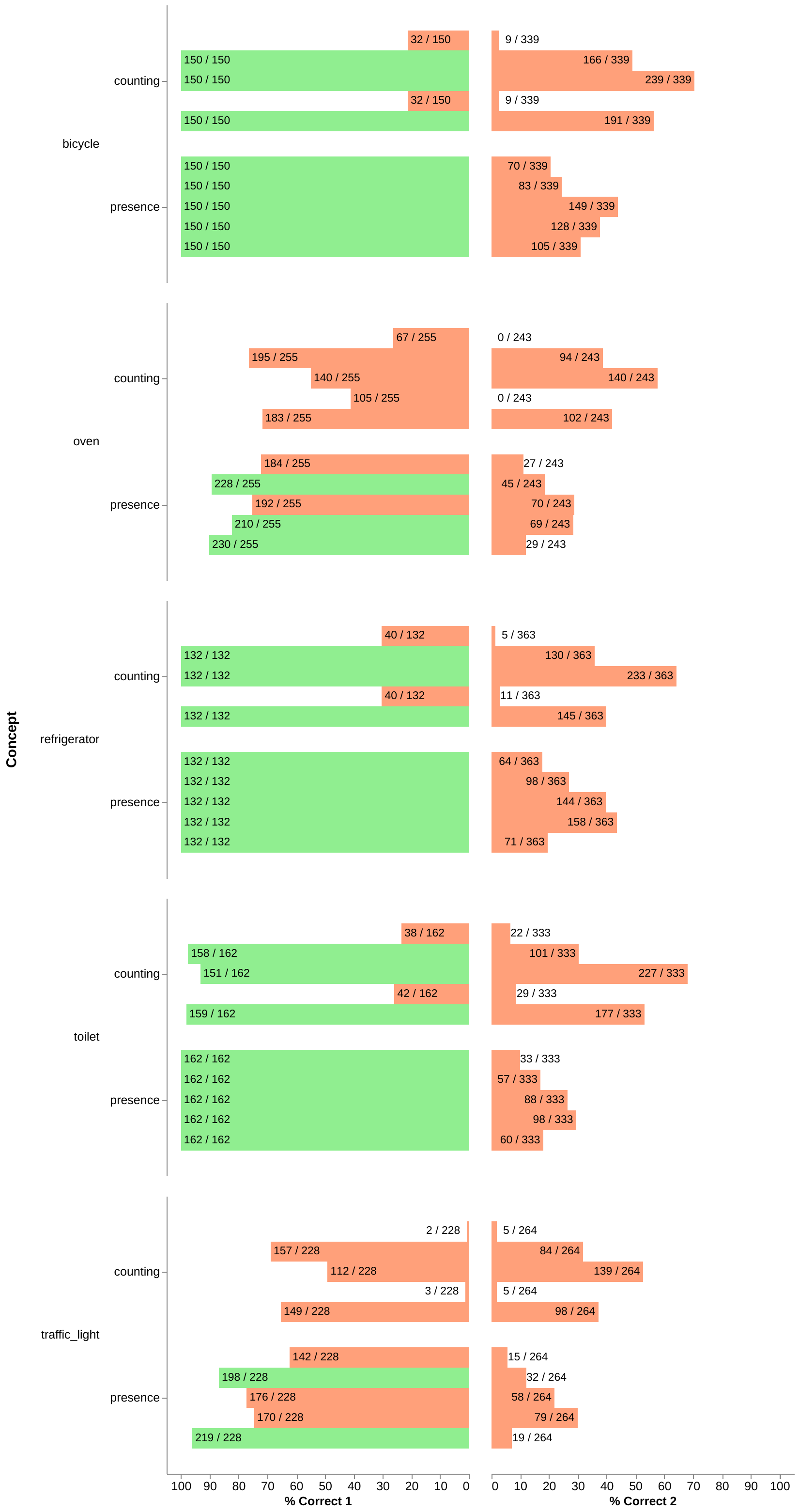}
    \caption{TinyGPT-V level 1 results on \cocoscrapbook.}
    \label{fig:tinygpt-v_coco_results}
\end{figure}

The performance of the TinyGPT-V model on the \cocoscrapbook\ Dataset at the initial level of validation is presented in Table~\ref{fig:tinygpt-v_coco_results}. The model apparently demonstrates a high degree of proficiency in recognizing the presence of objects, achieving over 75\% accuracy across all concepts. Additionally, the model exhibits somewhat efficacy in quantifying the number of objects present in images when a condition or direction is given, in the form of ``answer with a number, when applicable'' or ``based on the image, respond to this question with a short answer''.

\begin{table*}[!htb]
    \setlength{\tabcolsep}{2.8mm}
    \centering
    \caption{TinyGPT-V error types for all \textit{Scrapbook} datasets.}
    \label{tab:tinygpt-v_errors}
    \small
    \begin{tabular}{c|cr|cr|cr}
        \hline
        \multirow{2}{*}{Errors} & \multicolumn{2}{c|}{\cocoscrapbook} & \multicolumn{2}{c|}{\cocoplainbgscrapbook} & \multicolumn{2}{c}{\uniqueallscrapbook} \\ \cline{2-7}
        & \multicolumn{1}{c|}{Non-absurd} & All & \multicolumn{1}{c|}{Non-absurd} & All & \multicolumn{1}{c|}{Non-absurd} & All \\
        \hline
        Answer disagreement & \multicolumn{1}{c|}{27187} & 27187 & \multicolumn{1}{c|}{9092} & 9092 & \multicolumn{1}{c|}{18298} & 18325 \\
        Error disagreement & \multicolumn{1}{c|}{10714} & 10845 & \multicolumn{1}{c|}{5242} & 5835 & \multicolumn{1}{c|}{53242} & 68690 \\
        Plausible wrong answers & \multicolumn{1}{c|}{\textcolor{red}{28330}} & 94484 & \multicolumn{1}{c|}{\textcolor{red}{33056}} & 80403 & \multicolumn{1}{c|}{\textcolor{red}{199770}} & 446475 \\
        Unexpected answers & \multicolumn{1}{c|}{396} & 396 & \multicolumn{1}{c|}{320} & 320 & \multicolumn{1}{c|}{10614} & 11561 \\
        Invalidated by simpler image & \multicolumn{1}{c|}{1152} & 1152 & \multicolumn{1}{c|}{299} & 299 & \multicolumn{1}{c|}{3433} & 3433 \\
        Total & \multicolumn{1}{c|}{67779} & 134064 & \multicolumn{1}{c|}{48009} & 95949 & \multicolumn{1}{c|}{285357} & 548484 \\
        \hline
    \end{tabular}
\end{table*}

However, the model demonstrates poor performance in the absence of objects in the scene, with results that are lower than chance level. This suggests that the model may have been inadequately trained with negative examples, potentially exhibiting a bias towards answering ``yes'' to questions about the presence of objects and a tendency to answer more than one when questioned about the number of objects of any particular type.

\begin{figure}[!htb]
    \centering
    \includegraphics[width=\linewidth]{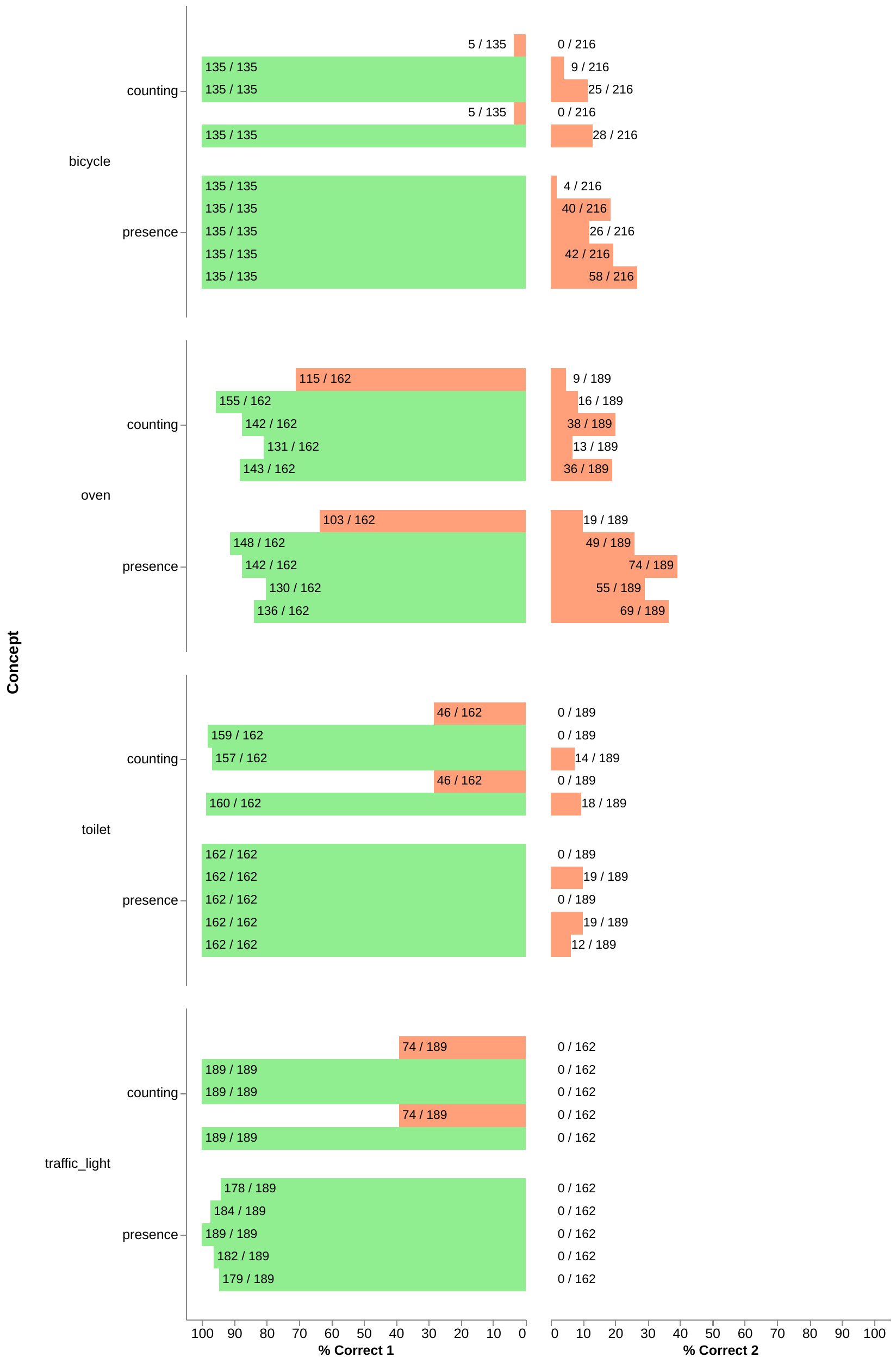}
    \caption{TinyGPT-V level 1 results on \cocoplainbgscrapbook.}
    \label{fig:tinygpt-v_coco_plain_bg_results}
\end{figure}

\begin{figure}[!htb]
    \centering
    \includegraphics[width=\linewidth]{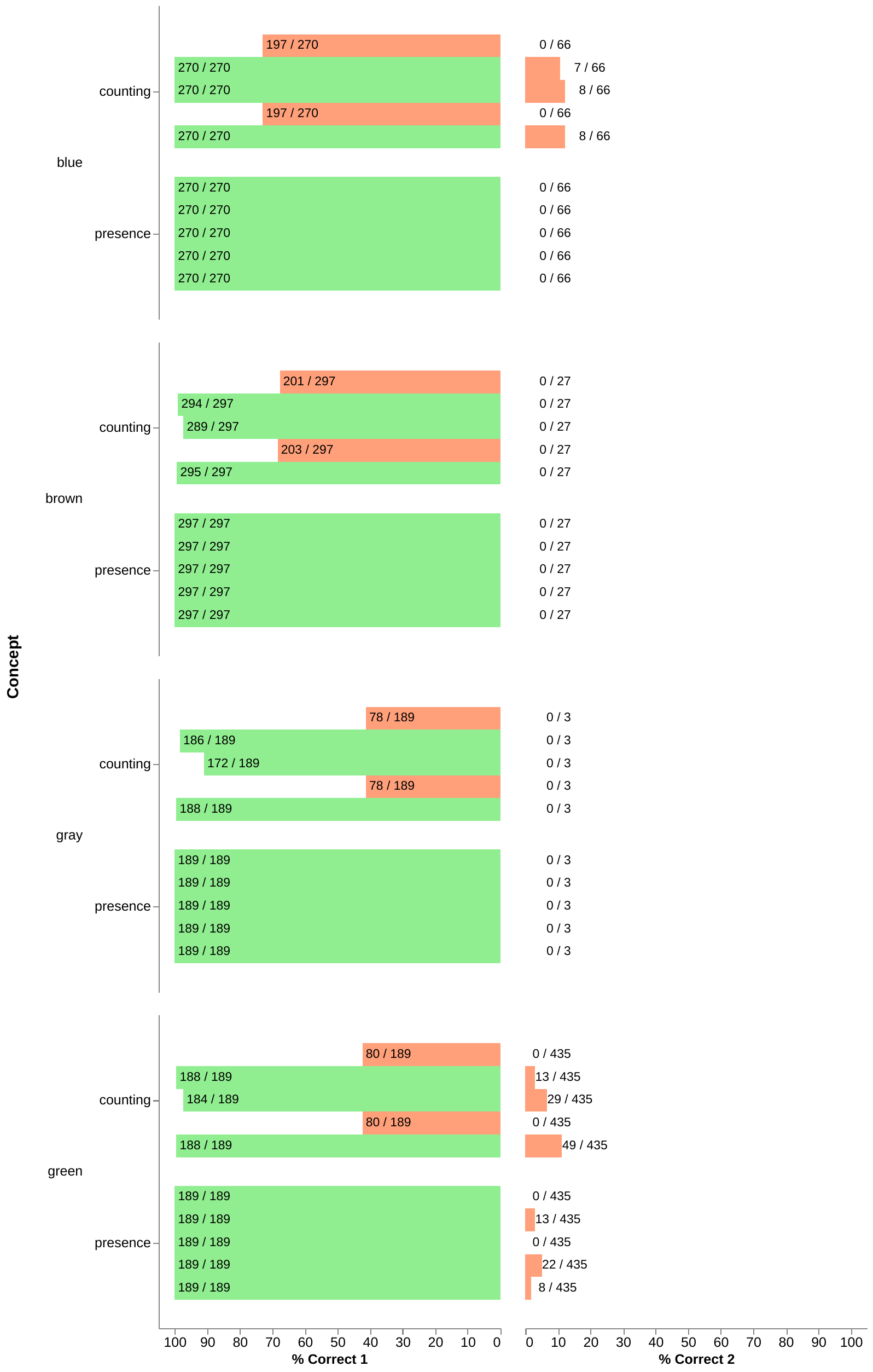}
    \caption{TinyGPT-V level 1 color results on \uniqueallscrapbook.}
    \label{fig:tinygpt-v_unique_all_deterministic_colors_results}
\end{figure}

\begin{figure}[!htb]
    \centering
    \includegraphics[width=\linewidth]{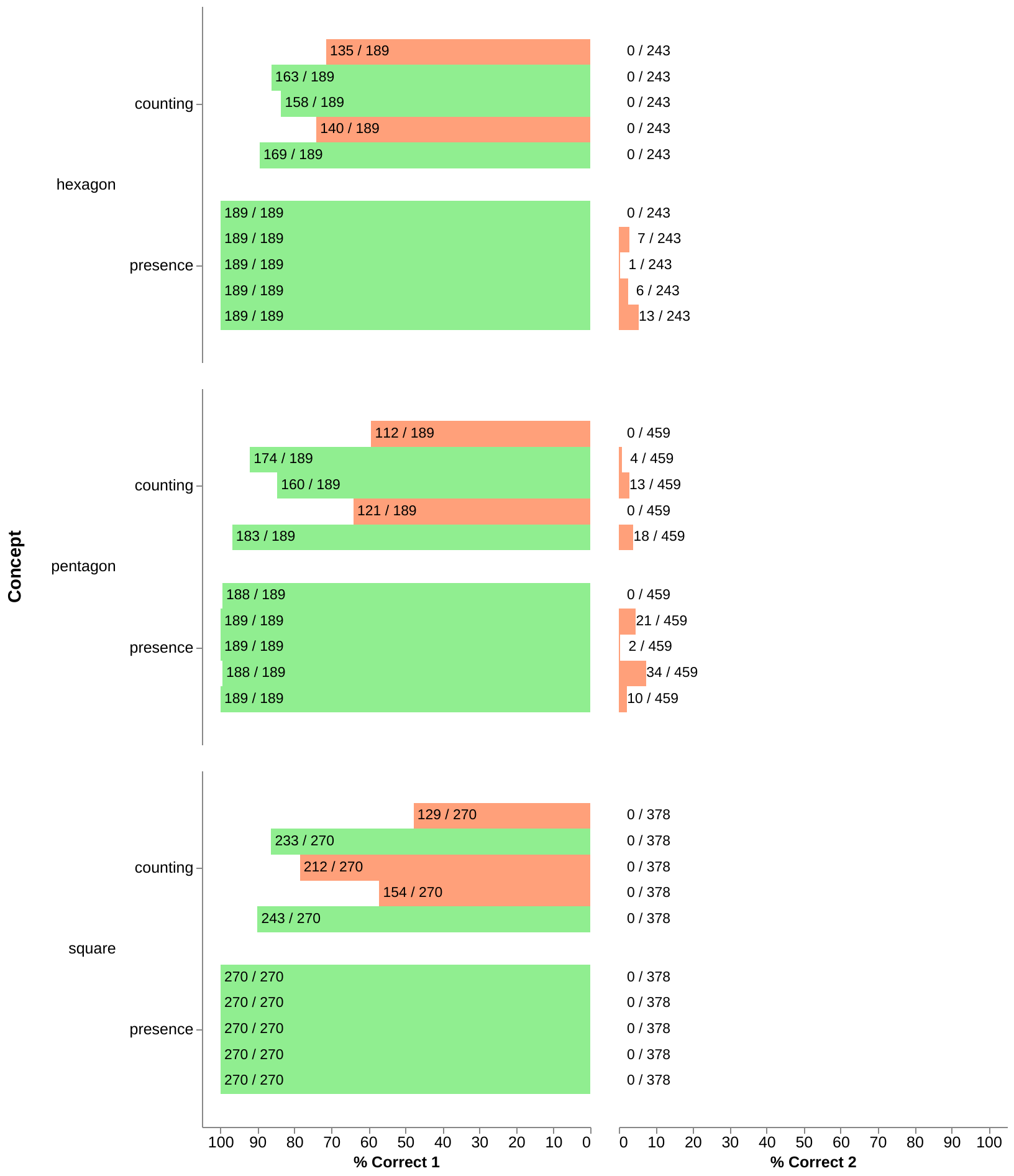}
    \caption{TinyGPT-V level 1 objects results on \uniqueallscrapbook.}
    \label{fig:tinygpt-v_unique_all_deterministic_objs_results}
\end{figure}

\begin{figure}[!htb]
    \centering
    \includegraphics[width=\linewidth]{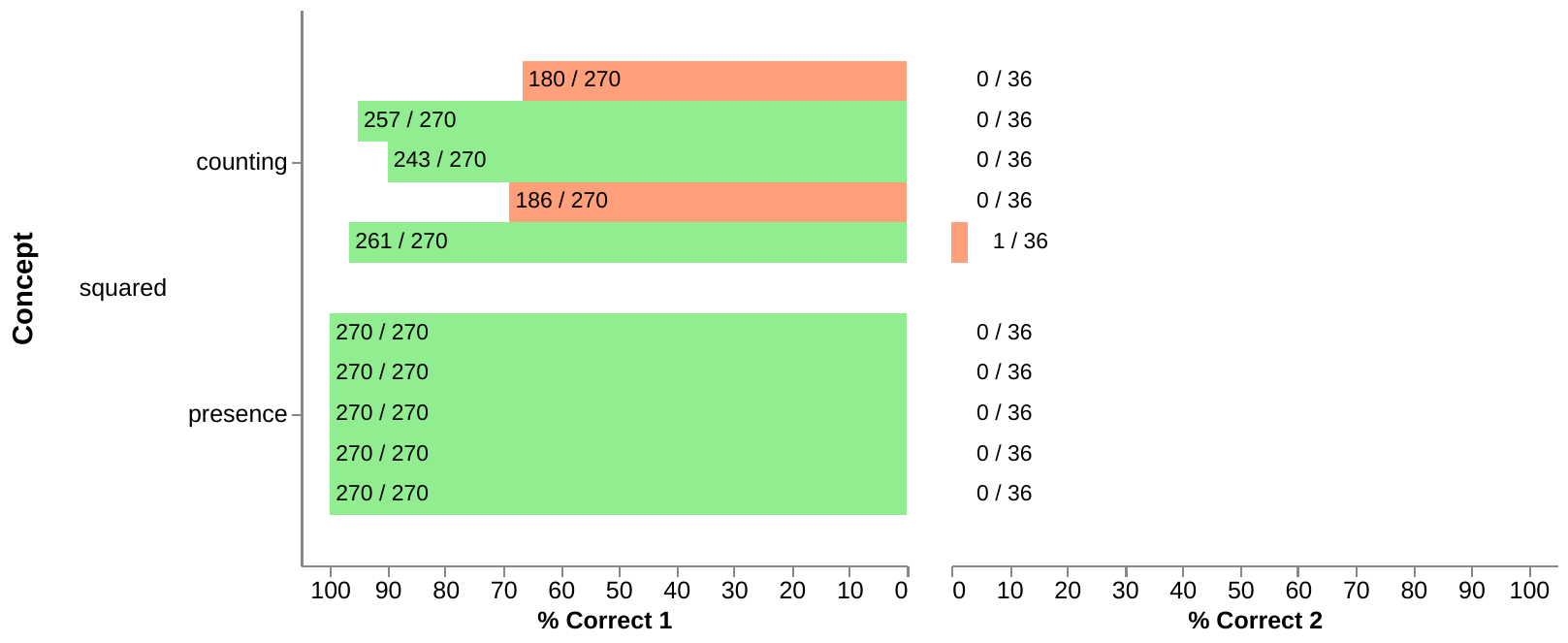}
    \caption{TinyGPT-V level 1 shapes results on \uniqueallscrapbook.}
    \label{fig:tinygpt-v_unique_all_deterministic_shapes_results}
\end{figure}

Furthermore, an analysis reveals a decline in model performance when the potential answers to a question are enumerated. This suggests that the model is unable to effectively process the additional information provided. A similar behavior is observed when the model is tested on the \cocoplainbgscrapbook\ and \uniqueallscrapbook\ datasets, as illustrated in Figures~\ref{fig:tinygpt-v_coco_plain_bg_results}, \ref{fig:tinygpt-v_unique_all_deterministic_colors_results}, \ref{fig:tinygpt-v_unique_all_deterministic_objs_results}, and \ref{fig:tinygpt-v_unique_all_deterministic_shapes_results}.

\subsection{Discussion}

The findings concerning the baseline models' performance on the various versions of the \scrapbook\ dataset indicate significant variations in their accuracy, depending on the supplementary information provided with each question. This suggests a failure to adequately handle diverse linguistic aspects. Notably, MiniGPT-4 exhibited the least variability in accuracy across different question types when the answers pertained to objects from the COCO dataset. However, no model has yet demonstrated satisfactory performance at the initial level in terms of color and geometric shapes, indicating a lack of overall ability to accurately extract the fundamental characteristics of an object necessary to respond to the posed questions.

TinyGPT-V exhibited a bias toward affirmative responses to questions concerning the existence of objects, while MobileVLM-V2 demonstrated an inability to address counting questions pertaining to geometric shapes across all question types. Both models failed at level 1 questions, and as a result, could not be evaluated at subsequent levels. When evaluated at the second level, MiniGPT-4 failed to accurately answer questions when the object was present in the image, but in a different position, suggesting a lack of understanding regarding image positioning. This finding indicates that the models are only capable of perceiving the presence of an object, rather than interpreting its specific position within the image.

To ensure the effective deployment of these models, further research is necessary to enhance their understanding of the relationships between objects and their positions within images, as well as the linguistic nuances that can influence their responses. As discussed and illustrated in this study, a model's correct response does not necessarily equate to an accurate interpretation of the image, which is a crucial consideration for the broader utilization of these models.

Future research should prioritize enhancing the models' capacity to discern the context of images while addressing the intricacies of queries and providing precise answers based on the information provided. While a substantial dataset may offer the models a general understanding of the objects, it is imperative to probe them with questions that demand a more profound comprehension of the objects' fundamental properties and their interrelationships. It is unreasonable to expect a model to respond correctly and with a comprehensive explanation to a complex inquiry if it is unable to answer a basic one.

\section{Conclusions}

In this work, we introduced a novel framework, \scrapbook, designed to generate extensive datasets for probing the learned concepts of AI models. By focusing on fundamental concepts such as object recognition, absolute and relative positions, and attribute identification, we aimed to ensure that models can reliably understand and process essential elements of visual information before tackling more complex tasks.

Our experiments demonstrated that while current models exhibited proficiency in recognizing the presence of objects and enumerating them, they encountered challenges in comprehending positional information and handling questions with additional conditions. This finding indicates a need for further research and development to enhance the models' comprehension of spatial relationships and their ability to process more complex queries.

TinyGPT-V and MobileVLM-V2 failed to reach the minimum threshold required for level 1 questions, thus indicating an incapacity to answer fundamental questions concerning the concepts. Conversely, MiniGPT-4 was capable of reaching level 2 questions for at least one concept. However, it encountered difficulties with more complex questions, such as those involving absurdity. Additionally, it faltered in questions pertaining to the placement of objects in images, indicating an inability to comprehend absolute position concepts.

The \scrapbook\ framework provides a valuable tool for generating diverse and comprehensive datasets, which can be used to rigorously evaluate and improve the performance of AI models. The detailed analysis of the models' errors revealed that current AI systems often provide contextually plausible but incorrect answers, highlighting the need for more robust training data and algorithms to handle complex queries and positional information accurately.

Future work will focus on expanding the framework to include more complex concepts and question templates, while exploring new methods to enhance model interpretability and explainability. Additionally, improving the models' ability to handle diverse linguistic aspects and accurately extract fundamental characteristics of objects will be crucial for enhancing their overall performance and reliability.

Overall, our approach highlights the importance of thoroughly validating AI models' understanding of basic concepts, paving the way for more robust and reliable AI systems capable of performing a wide range of tasks with greater accuracy and consistency. By addressing the identified shortcomings, we can develop AI models that not only recognize objects but also understand their spatial relationships and contextual nuances, leading to more advanced and dependable AI applications.

\section*{Acknowledgments}
\label{acknowledgment}

The authors would like to thank National Council for Scientific and Technological Development (CNPq \# 309330/2018-1) for its financial support and NVIDIA for the donation of a GPU as part of the GPU Grant Program.

\bibliographystyle{model3-num-names}
\bibliography{paper}

\section{Appendix}

\subsection{MSCOCO Objects Classes}
\label{app:mscoco_objects}

\begin{myitemize}
    \item accessory: backpack, handbag, suitcase, tie, umbrella;
    \item animal: bear, cat, cow, dog, elephant, giraffe, horse, sheep, zebra;
    \item appliance: microwave, oven, refrigerator, sink, toaster;
    \item electronic: cell phone, keyboard, laptop, mouse, remote;
    \item food: banana, broccoli, cake, donut, hot dog, orange, pizza, sandwich;
    \item furniture: bed, couch, toilet;
    \item indoor: book, clock, scissors, teddy bear, toothbrush;
    \item kitchen: bottle, fork, knife, spoon;
    \item outdoor: bench, fire hydrant, stop sign, traffic light;
    \item sports: baseball bat, skateboard, tennis racket;
    \item vehicle: airplane, bicycle, boat, bus, car, motorcycle, train, truck.
\end{myitemize}

\subsection{Models Results on Scrapbook Datasets}

\begin{figure*}[!htb]
	\centering
	\includegraphics[width=\linewidth]{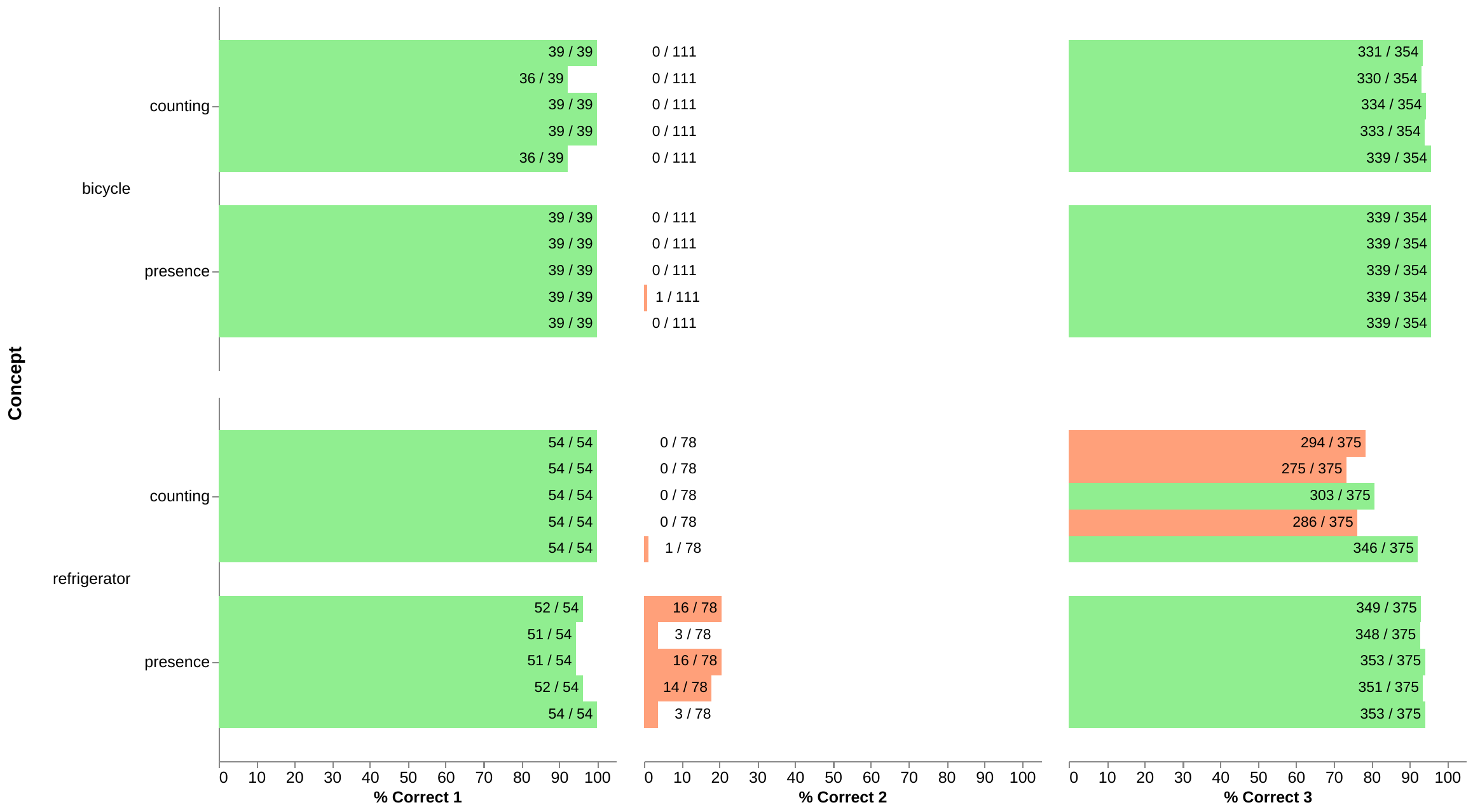}
	\caption{MiniGPT-4 level 2 results on COCO for absolute right position}
	\label{fig:minigpt-4_coco_absolute_right_results}
\end{figure*}

\begin{figure*}[!htb]
	\centering
	\includegraphics[width=\linewidth]{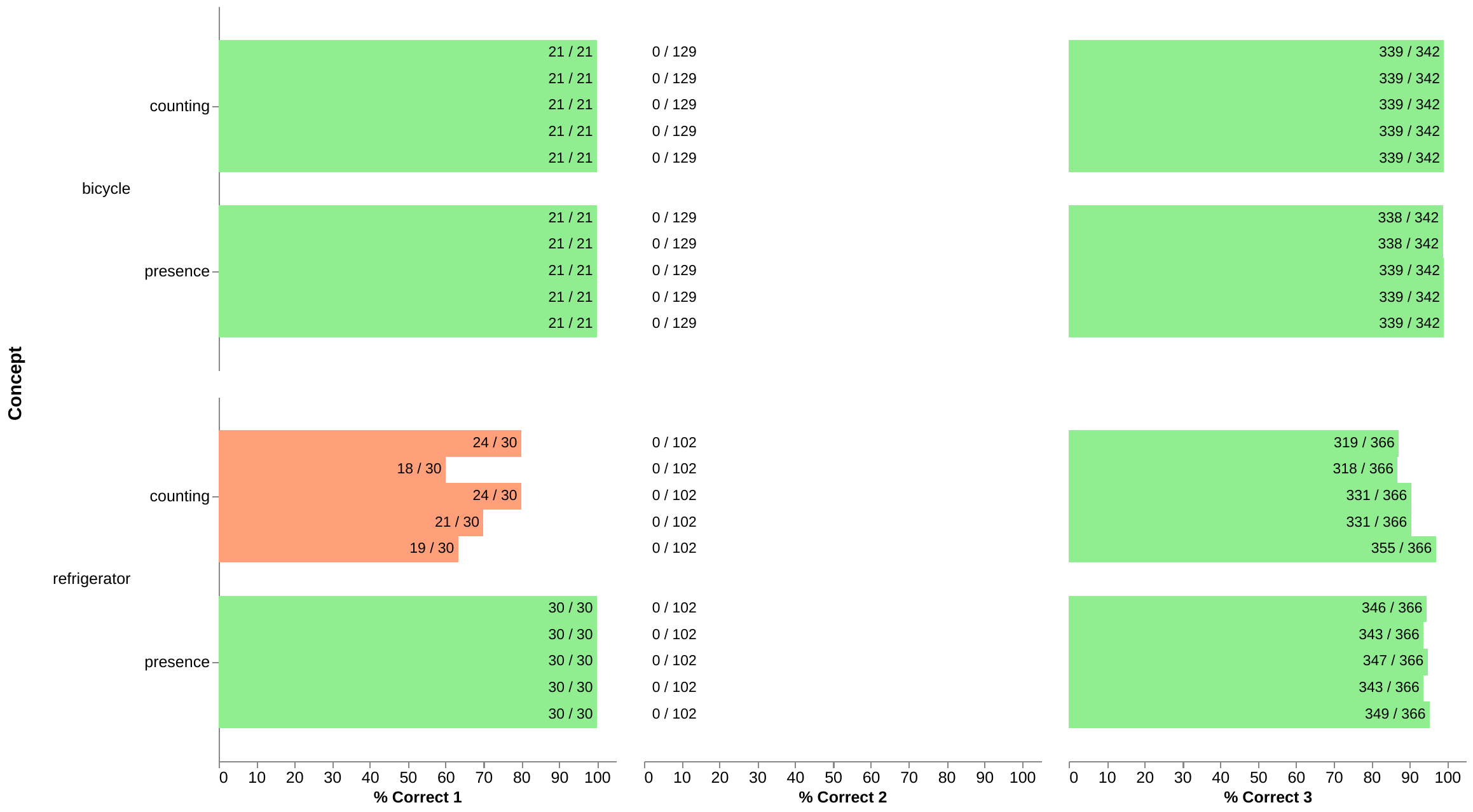}
	\caption{MiniGPT-4 level 2 results on COCO for center position}
	\label{fig:minigpt-4_coco_center_results}
\end{figure*}

\begin{figure*}[!htb]
	\centering
	\includegraphics[width=\linewidth]{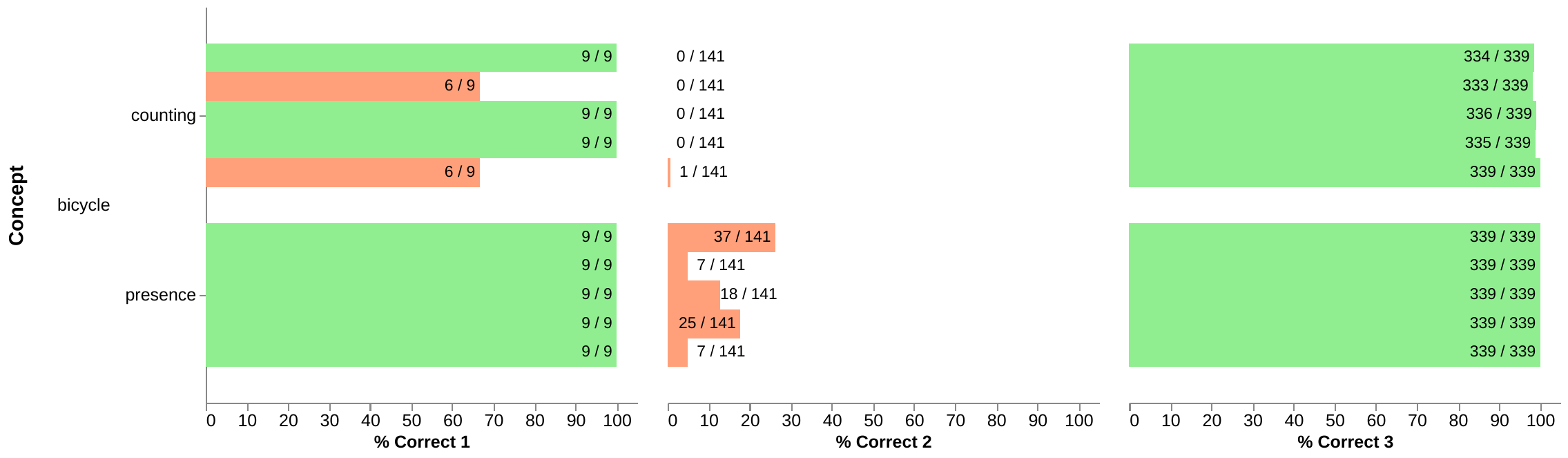}
	\caption{MiniGPT-4 level 2 results on COCO for down left position}
	\label{fig:minigpt-4_coco_down_left_results}
\end{figure*}

\begin{figure*}[!htb]
	\centering
	\includegraphics[width=\linewidth]{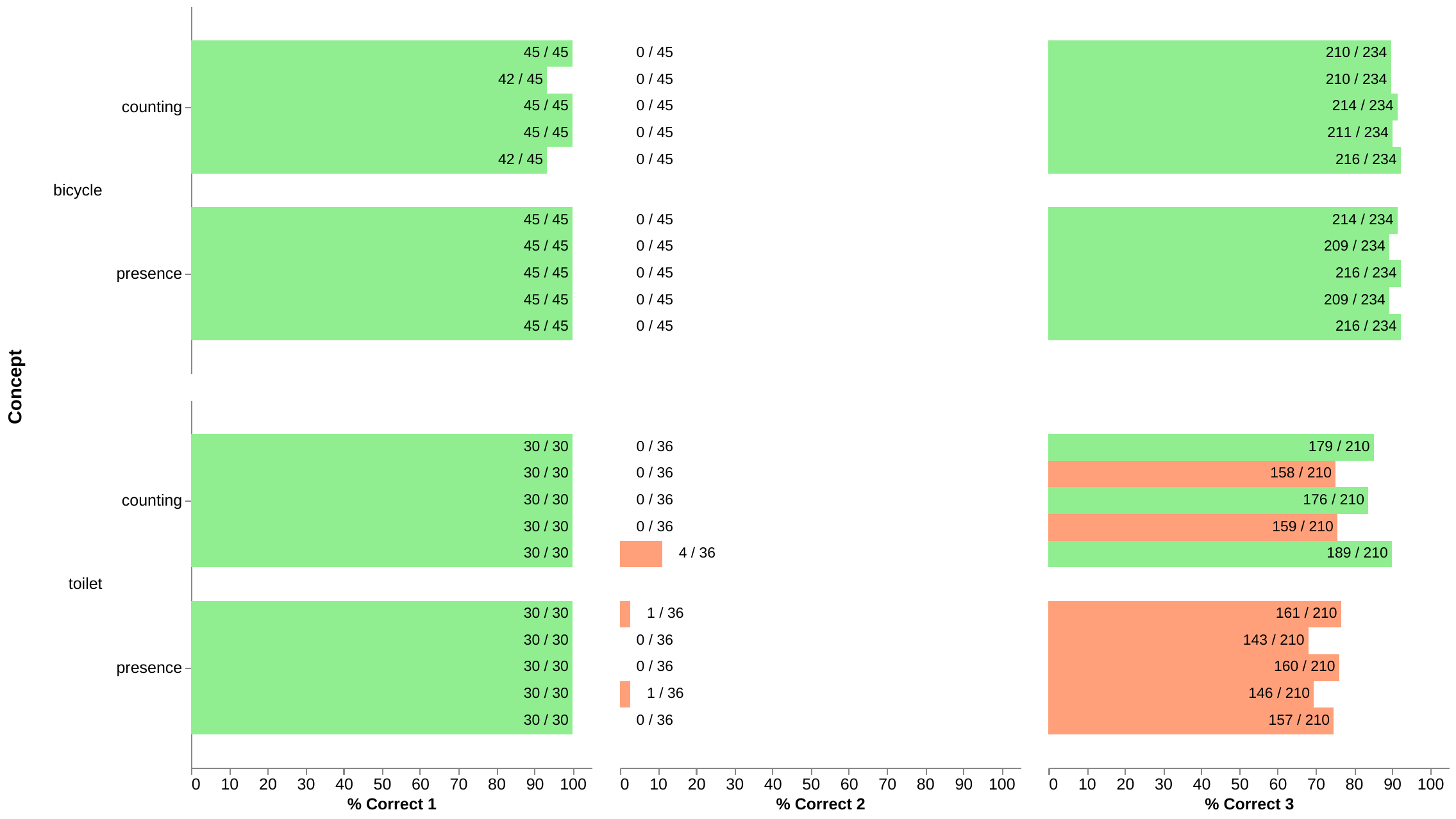}
	\caption{MiniGPT-4 level 2 results on COCO with plain background for absolute right position}
	\label{fig:minigpt-4_coco_plain_bg_absolute_right_results}
\end{figure*}

\begin{figure*}[!htb]
	\centering
	\includegraphics[width=\linewidth]{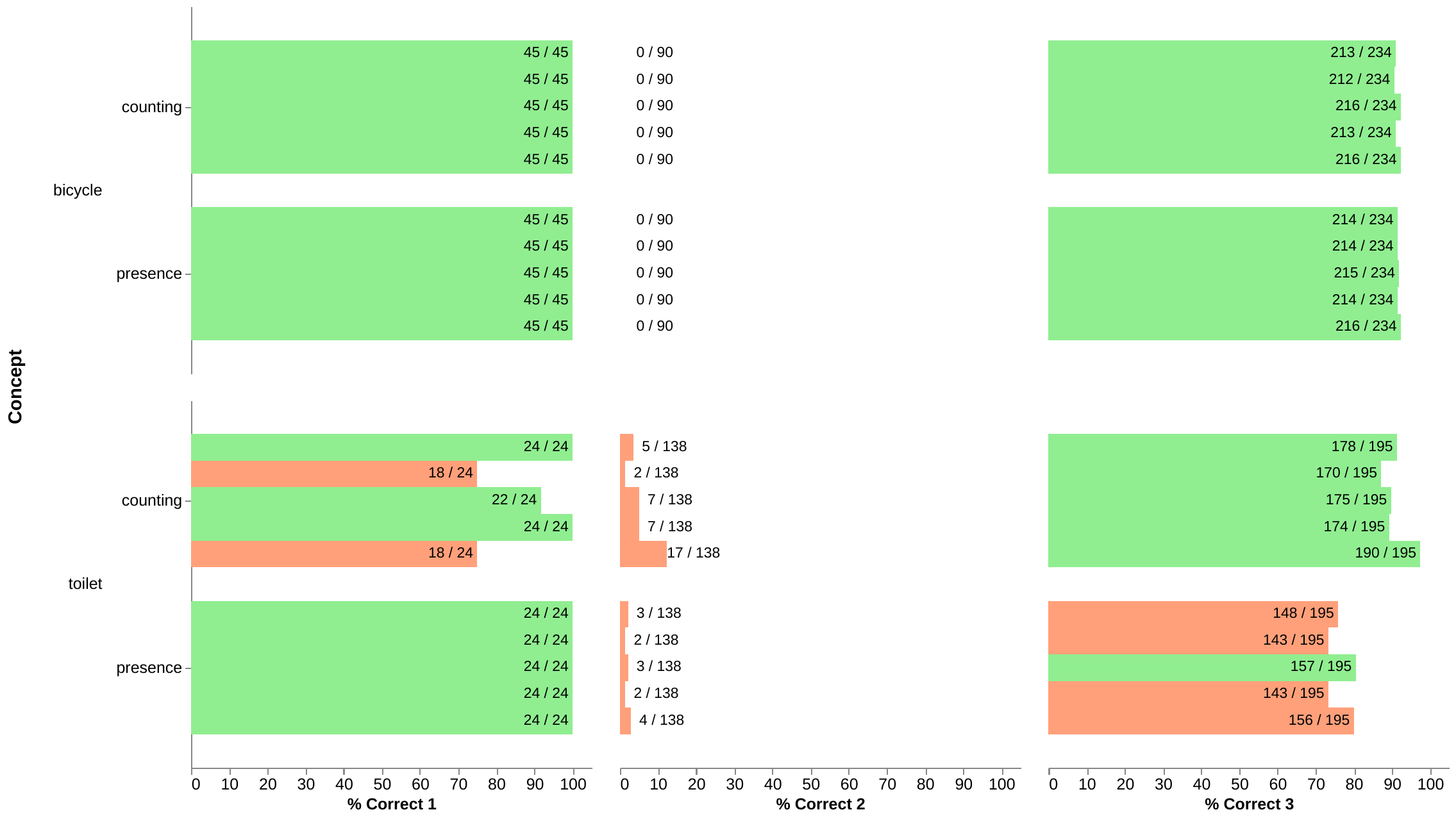}
	\caption{MiniGPT-4 level 2 results on COCO with plain background for center position}
	\label{fig:minigpt-4_coco_plain_bg_center_results}
\end{figure*}

\begin{figure*}[!htb]
	\centering
	\includegraphics[width=\linewidth]{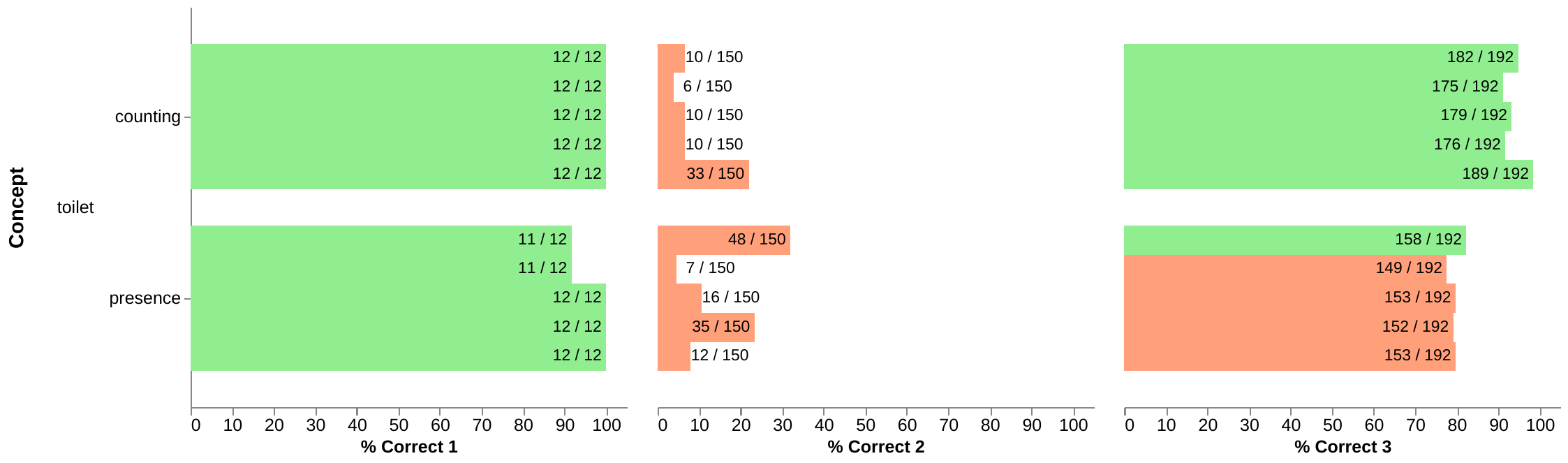}
	\caption{MiniGPT-4 level 2 results on COCO with plain background for up left position}
	\label{fig:minigpt-4_coco_plain_bg_up_left_results}
\end{figure*}

\begin{figure*}[!htb]
	\centering
	\includegraphics[width=\linewidth]{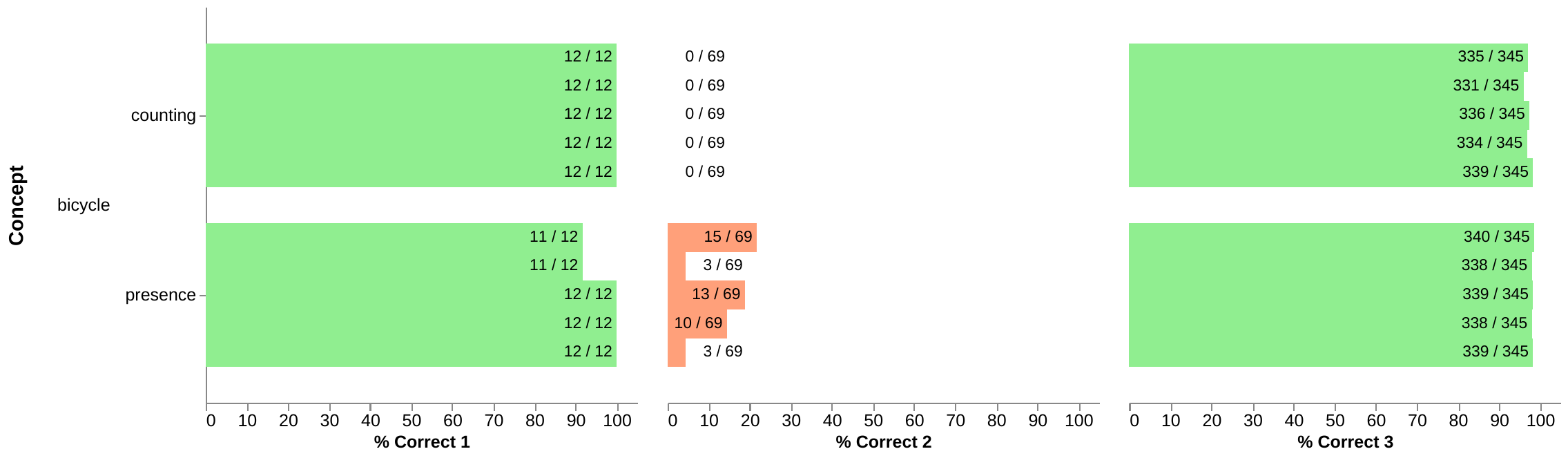}
	\caption{MiniGPT-4 level 2 results on COCO for up right position}
	\label{fig:minigpt-4_coco_up_right_results}
\end{figure*}

\end{document}